%% file: main.tex
\documentclass[10pt,twocolumn,letterpaper]{article}

\usepackage{cvpr}              %
\makeatletter
\@namedef{ver@everyshi.sty}{}
\makeatother
\usepackage{graphicx}
\usepackage{amsmath}
\usepackage{amssymb}
\usepackage{booktabs}
\usepackage{adjustbox}
\usepackage{tikz}
\usepackage{tikz-3dplot}
\usepackage{pgfplots}
\usepackage{graphicx}
\usepackage{color}
\usepackage{units}
\usepackage{tabularx}
\usepackage{xfrac}
\usepackage{enumitem}

\usepackage{makecell}
\usepackage{array}
\newcolumntype{P}[1]{>{\centering\arraybackslash}p{#1}}
\newcolumntype{M}[1]{>{\centering\arraybackslash}m{#1}}
\newcolumntype{C}{>{\centering\arraybackslash}X}
\usetikzlibrary{calc,math, positioning, intersections, decorations, external, patterns, arrows,perspective}

\pgfplotsset{compat=1.16}

\newcommand{\Sec}{Section~}
\newcommand{\Fig}{Fig.~}
\newcommand{\Tab}{Table~}

\newcommand{\Eq}{Equation.~}
\newcommand{\aeone}{AE${<}11.25$}
\newcommand{\aetwo}{AE${<}22.5$}
\newcommand{\aethree}{AE${<}30$}
\graphicspath{ {./images/} }

\newcommand{\mae}[4]
{   \begin{tikzpicture}[]
        \node at (0,0) (pic) [anchor = north, inner sep = 0pt, outer sep = 0pt, draw = black] { \includegraphics[width=#2]{#1} };
        \node at (pic.north west) [below right, inner sep = 1pt, outer sep = 1pt, fill = #4, draw opacity = 1.0, fill opacity = 0.7] {\tiny #3};
    \end{tikzpicture}
}

\usepackage[pagebackref,breaklinks,colorlinks]{hyperref}

\newcommand{\rotateRPY}[3]%
{   \pgfmathsetmacro{\rollangle}{#1}
    \pgfmathsetmacro{\pitchangle}{#2}
    \pgfmathsetmacro{\yawangle}{#3}

    \pgfmathsetmacro{\newxx}{cos(\yawangle)*cos(\pitchangle)}
    \pgfmathsetmacro{\newxy}{sin(\yawangle)*cos(\pitchangle)}
    \pgfmathsetmacro{\newxz}{-sin(\pitchangle)}
    \path (\newxx,\newxy,\newxz);
    \pgfgetlastxy{\nxx}{\nxy};

    \pgfmathsetmacro{\newyx}{cos(\yawangle)*sin(\pitchangle)*sin(\rollangle)-sin(\yawangle)*cos(\rollangle)}
    \pgfmathsetmacro{\newyy}{sin(\yawangle)*sin(\pitchangle)*sin(\rollangle)+ cos(\yawangle)*cos(\rollangle)}
    \pgfmathsetmacro{\newyz}{cos(\pitchangle)*sin(\rollangle)}
    \path (\newyx,\newyy,\newyz);
    \pgfgetlastxy{\nyx}{\nyy};

    \pgfmathsetmacro{\newzx}{cos(\yawangle)*sin(\pitchangle)*cos(\rollangle)+ sin(\yawangle)*sin(\rollangle)}
    \pgfmathsetmacro{\newzy}{sin(\yawangle)*sin(\pitchangle)*cos(\rollangle)-cos(\yawangle)*sin(\rollangle)}
    \pgfmathsetmacro{\newzz}{cos(\pitchangle)*cos(\rollangle)}
    \path (\newzx,\newzy,\newzz);
    \pgfgetlastxy{\nzx}{\nzy};
}

\usepackage[capitalize]{cleveref}
\crefname{section}{Sec.}{Secs.}
\Crefname{section}{Section}{Sections}
\Crefname{table}{Table}{Tables}
\crefname{table}{Tab.}{Tabs.}

\definecolor{somegray}{rgb}{0.5, 0.5, 0.5}
\newcommand{\darkgrayed}[1]{\textcolor{somegray}{#1}}
\makeatletter
\newcommand*\titleheader[1]{\gdef\@titleheader{#1}}
\AtBeginDocument{%
  \let\st@red@title\@title
  \def\@title{%
    \vskip-3em
    \bgroup\normalfont\large\centering\@titleheader\par\egroup
    \vskip1.5em\st@red@title}
}
\makeatother

\titleheader{\darkgrayed{This paper has been accepted for publication at the\\
IEEE Conference on Computer Vision and Pattern Recognition (CVPR), Vancouver, 2023.
\copyright IEEE}}

\title{Event-based Shape from Polarization}

\begin{document}

\author{Manasi Muglikar \textsuperscript{1}
\qquad
Leonard Bauersfeld \textsuperscript{1}
\qquad 
Diederik Paul Moeys \textsuperscript{2}
\qquad 
Davide Scaramuzza \textsuperscript{1} \\[6pt]
\,\textsuperscript{1}Robotics and Perception Group, University of Zurich, Switzerland \\
\textsuperscript{2}Advanced Sensors and Modelling  Group, SONY R\&D Center Europe, SL1
}
\newcommand\nomarkerfootnote[1]{%
  \begingroup
  \renewcommand\thefootnote{}\footnote{#1}%
  \addtocounter{footnote}{-1}%
  \endgroup
}

\maketitle

 \input{sections/00_abstract}

 \input{sections/01_introduction}

 \input{sections/02_related_works}

 \input{sections/03_methodology}
 \input{sections/04_experiments}

 \input{sections/05_conclusion}
 \input{sections/06_acknoledgement}

\input{sections/supplementary_material}
{\small
\bibliographystyle{ieee_fullname}
\bibliography{main}
}

\end{document}

%% file: sections/00_abstract.tex
\begin{abstract}
    State-of-the-art solutions for Shape-from-Polarization (SfP) suffer from a speed-resolution tradeoff: they either sacrifice the number of polarization angles measured or necessitate lengthy acquisition times due to framerate constraints, thus compromising either accuracy or latency.
    We tackle this tradeoff using event cameras. Event cameras operate at microseconds resolution with negligible motion blur, and output a continuous stream of events that precisely measures how light changes over time asynchronously.
    We propose a setup that consists of a linear polarizer rotating at high speeds in front of an event camera.
    Our method uses the continuous event stream caused by the rotation to reconstruct relative intensities at multiple polarizer angles. 
    Experiments demonstrate that our method outperforms physics-based baselines using frames, reducing the MAE by $25\%$ in synthetic and real-world datasets. 
    In the real world, we observe, however, that the challenging conditions (i.e., when few events are generated) harm the performance of physics-based solutions. 
    To overcome this, we propose a learning-based approach that learns to estimate surface normals even at low event-rates, improving the physics-based approach by $52\%$ on the real world dataset.
    The proposed system achieves an acquisition speed equivalent to $50$ fps ($>$twice the framerate of the commercial polarization sensor) while retaining the spatial resolution of 1MP.
    Our evaluation is based on the first large-scale dataset for event-based SfP.

    \vspace*{-12pt}
\end{abstract}

%% file: sections/01_introduction.tex
\vspace*{3pt}
\noindent\textbf{Code, dataset and video are available under: } \\
\url{https://rpg.ifi.uzh.ch/esfp.html} \\
    \url{https://youtu.be/sF3Ue2Zkpec}

\section{Introduction}
\begin{figure}
    \centering
    \includegraphics[width=0.92\linewidth]{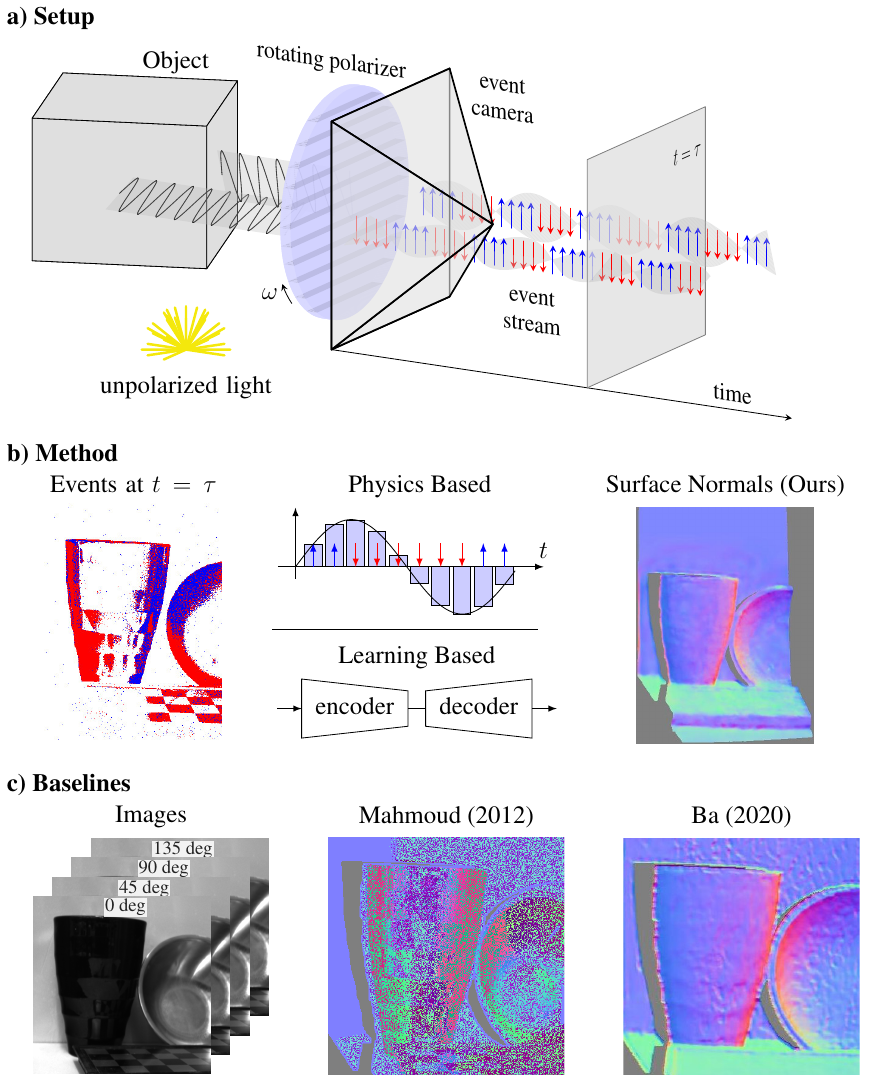}
    \vspace*{-9pt}
    \caption{Surface normal estimation using event-based SfP.
    (a) Rotating a polarizer in front of an event camera creates sinosoidal changes in intensities, triggering events.
    (b) The proposed event-based method uses the continuous event stream to reconstruct relative intensities at multiple polarizer angles which is used to estimate surface normals using physics-based and learning-based method.
    (c) Our approach outperforms image-based baselines \cite{Mahmoud12ICIP, Ba20ECCV}.
    }
    \label{fig:method_overview}
    \vspace*{-18pt}
\end{figure}

Polarization cues have been used in many applications across computer vision, including image dehazing \cite{Schechner01CVPR}, panorama stitching and mosaicing \cite{Schechner05PAMI}, reflection removal \cite{Lei20CVPR}, image segmentation \cite{Mei22CVPR}, optical flow gyroscope, \cite{Tzabari20ECCV} and material classification \cite{baek2018siggraph}.
Among these, Shape-from-Polarization (SfP) methods exploit changes in polarization information to infer geometric properties of an object \cite{Wolff97IVC, Atkinson18MVA, Kadambi15ICCV, Ba20ECCV, Lei22CVPR}. 
It uses variations in radiance under different polarizer angles to estimate the 3D surface of a given object. 
In particular, when unpolarized light is reflected from a surface, it becomes partially polarized depending on the geometry and material of the surface. 
Surface normals, and thus 3D shape, can then be estimated by orienting a polarizing filter in front of a camera sensor and studying the relationship between the polarizer angle and the magnitude of light transmission.
SfP has a number of advantages over both active and passive depth sensing methods. Unlike active depth sensors that use structured light (SL) \cite{KinectV1, IntelRealSense} or time-of-flight (ToF), SfP is not limited by material type and can be applied to non-Lambertian surfaces like transparent glass and reflective, metallic surfaces.

Despite these advantages, however, estimating high-quality surface normals from polarization images is still an open challenge. %
\emph{Division of Focal Plane (DoFP)} methods \cite{PolarM17,Lucid18,Lei22CVPR, Ba20ECCV} trade-off spatial resolution for latency and allow for the capture of four polarizations in the same image. This is achieved through a complex manufacturing process that requires precisely placing a micro-array of four polarization filters on the image sensor \cite{Lucid18, PolarM17}, as shown in \Fig \ref{fig:dofp_dot:a} 
Despite the reduced latency, this system constrains the maximum number of polarization angles that can be captured, potentially impacting the accuracy of the estimates as we show in our results.%
Additionally, the spatial resolution of the sensor is also reduced, requiring further mosaicing-based algorithms for high-resolution reconstruction \cite{Wu21Optics}.
On the other hand, \emph{Division of Time  (DoT)} methods \cite{Wolff97IVC, Atkinson18MVA, Kadambi15ICCV} provide full-resolution images and are not limited in the number of polarization angles they can capture thanks to a rotating polarizing filter put in front of the image sensor. 
The frame rate of the sensing camera, however, effectively limits the rate at which the filter can rotate, increasing the acquisition time significantly (acquisition time $= N/f$, where $N$ is the number of polarizer angles and $f$ is the framerate of the camera). 
For this reason, commercial solutions, such as the Lucid Polarisens \cite{Lucid18}, favor DoFP, despite the lower resolution of both polarization angles and image pixels.
To overcome this shortcoming, recently, significant progress has been made with data-driven priors \cite{Ba20ECCV, Lei22CVPR}. However, these solutions still fall short in terms of computational complexity when compared to DoT methods. A solution able to bridge the accuracy of DoT with the speed of DoFP is thus still lacking in the field.

\input{images/fig_image_catagories}

In this paper, we tackle the speed-resolution trade-off using event cameras. Event cameras are efficient high-speed vision sensors that asynchronously measure changes in brightness intensity with microsecond resolution. We exploit these characteristics to design a DoT approach able to operate at high acquisition speeds (up to $5,000$ fps vs. $22$ fps of standard frame-based devices) and full-resolution ($1280 \times 720$) s shown in \Fig \ref{fig:method_overview}.
Thanks to the working principles of event-cameras, our sensing device provides a continuous stream of information for estimating the surface normal as compared to the discrete intensities captured at fixed polarization angles of traditional approaches. 
We present two algorithms to estimate surface normals from  events, one using geometry and the other based on a learning-based approach.
Our geometry-based method takes advantage of the continuous event stream to reconstruct relative intensities at multiple polarizer angles, which are then used to estimate the surface normal using traditional methods.
Since events provide a temporally rich information, this results in better reconstruction of intermediate intensities.
This leads to an improvement of upto $25\%$ in surface normal estimation, both on the synthetic dataset and on the real-world dataset.%
On the real dataset, however, the non-idealities of the event camera introduce a lower fill-rate (percentage of pixels triggering events) of $3.6\%$  in average (refer \Sec \ref{sec:psfp:noise}).
To overcome this, we propose a deep learning framework which uses a simple U-Net network to predict the dense surface normals from events. 
Our data-driven approach improves the accuracy over the geometry-based method by $52\%$.
\input{floats/tab_related_works}
Our contributions can be summarized as follows:
\vspace{-4pt}
\begin{itemize}[leftmargin=12pt]
    \setlength\itemsep{-2pt}
    \item A novel approach for shape-from-polarization using an event camera.
    Our approach utilizes the rich temporal information of events to reconstruct event intensities at multiple polarization angles.
    These event intensities are then used to estimate the surface normal.
    Our method outperforms previous state-of-the-art physics-based approaches using images by $25\%$ in terms of accuracy.
    \item A learning-based framework which predicts surface normals using events to solve the issue of low fill-rate common in the real-world.
    This  framework improves the estimation over physics-based approach by $52\%$ in terms of angular error.
    \item Lastly, we present the \textit{first} large scale dataset containing over 90 challenging scenes for SfP with events and images.
    Our dataset consists of events captured by rotating a polarizer in front of an event camera, as well as images captured using the Lucid Polarisens \cite{Lucid18}. %
\end{itemize}

%% file: images/fig_image_catagories.tex
\global\long\def\figHeight{2.0cm}
\begin{figure}[t!]
    \centering
    \subfloat[Division of focal plane]{\label{fig:dofp_dot:a}\includegraphics[trim={0cm 3cm 25cm 1cm},clip,height=\figHeight]{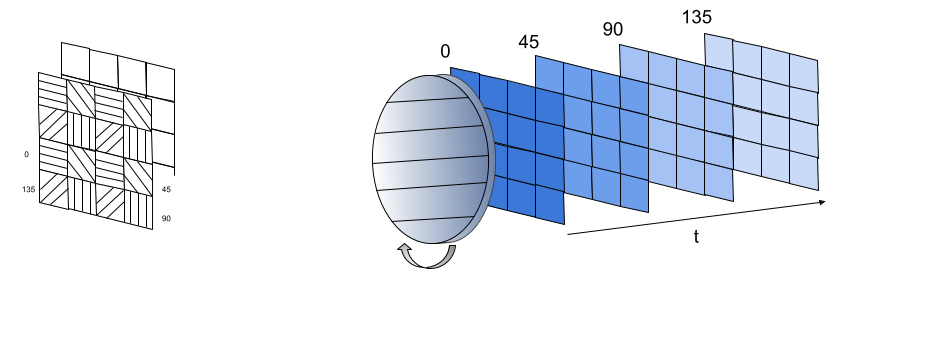}}\;
    \subfloat[Division of time]{\label{fig:dofp_dot:b}\includegraphics[trim={10cm 3cm 0cm 1cm},clip,height=\figHeight]{images/division_focal_plane_division_time.png}}\; \\
    
    \vspace*{-8pt}
    \caption{Illustration of SfP methods.}
    \label{fig:dofp_dot}
    \vspace*{-15pt}
\end{figure}

%% file: floats/tab_related_works.tex
\begin{table}[t!]
    \centering
    \begin{adjustbox}{max width=\columnwidth}
    \setlength{\tabcolsep}{3pt}
    \begin{tabular}{llll}
    \toprule
    \textbf{Dataset}  &  \textbf{Modality} & \textbf{(Resolution)} &  \textbf{Size}\\ \midrule
    Polar3D \cite{Kadambi15ICCV} & 6 Images(DoT) &18 MP  & 3 \\
    DeepSfP \cite{Ba20ECCV} & 4 Images(DoFP) & $1224 \times 1024$ & 236\\
    SPW \cite{Lei22CVPR} & 4 Images(DoFP) & $1224 \times 1024$ & 522\\ \midrule
    \textbf{ESfP- Synthetic (Ours)} & Events (DoT) + 12 Images(DoT) & $512 \times 512$ & $104$\\
    \textbf{ESfP- Real (Ours)} & Events (DoT) + 4 Images (DoFP) & $1280 \times 720$ & $90$\\
    \bottomrule
    \end{tabular}
    \end{adjustbox}
    \vspace{-1ex}
    \caption{\label{tab:related-work}Summary of publicly available datasets for SfP.}
    \vspace*{-9pt}
\end{table}

%% file: sections/02_related_works.tex
\section{Related work}
\vspace*{-3pt}

\paragraph{Frame-based SfP methods}
Shape-from-polarization estimates the normal of each point on an object's surface through Fresnel equations \cite{Collett05SPIE} by measuring the azimuthal and zenithal angles at each pixel. However, since a linear polarizer can only distinguish polarized light modulus $2\pi$, shape-from-polarization is often regarded as an under-determined problem and additional constraints are typically required to solve ambiguities in the measurements. 

Prior SfP systems that addressed the problem with traditional cameras are thoroughly discussed in \cite{durou20advances}. %
Early methods relied on assumptions about object surface and environment lighting to constrain the problem, such as pure specular \cite{rahmann2001reconstruction} or pure diffusion reflections \cite{atkinson2017polarisation} and surface convexity \cite{atkinson2006recovery,huynh2013shape}, but they typically fail when the theoretical model is violated. 
In order to ensure uniqueness in the solution, several methods exploit depth and geometric information to guide the surface reconstruction operation. A line of research \cite{atkinson2005multi,Miyazaki03ICCV,Miyazaki04PAMI,miyazaki2016surface} measures polarization information from multiple view points, while other authors exploit coarse depth maps from low-cost depth sensors, such as Kinect \cite{kadambi2017depth} or RGBD cameras \cite{zhu2019depth}, to obtain additional geometric cues. In single-view, spectral information measured by multi-band cameras can also be used to solve the $\pi$-ambiguity, as well as estimating refractive properties of the material useful for 3D reconstruction \cite{huynh2013shape,huynh2013shape,stolz2012shape}. Another class of methods considers photometric information to disambiguate SfP normal estimates by combining photometric stereo \cite{Drbohlav01ICCV, Ngo15CVPR, atkinson2017polarisation} and photometric constraints \cite{Mahmoud12ICIP} to impose shading constraints from multiple light directions.

More recently, \cite{Ba20ECCV, Lei22CVPR} presented a data-driven approach to solve the ambiguity, resulting in state-of-the-art performance.
Both these approaches, however, use a Polarisens sensor \cite{Lucid18} which compromises spatial resolution for faster acquisition time, thus limiting the potential of the learning-based approach.
We propose to address the shortcoming arsing form the spatial-temporal resolution trade-off using an event camera and take advantage of deep learning methods to tackle non-idealities of the event camera.

\vspace{-14pt}
\paragraph{Event-based SfP methods}
Since event cameras are novel sensors, there are very few papers that combine polarization information with events.
Recently, \cite{Gruev22SPIE} combined the DoFP approach with an event camera by manufacturing a micro array of polarizers and placing it on top of the event camera sensor.
This requires a very precise manufacturing process in which the polarizers are manually aligned with the pixels by inducing polarized motion for the pixels to see. Due to the low yield resulting from this process and to low-spatial resolution issues analogous to standard cameras, such as mosaicing, commercial solution are currently not available. 
Moreover, this approach also requires relative motion between the camera and the scene to capture event information, which further limits its application. 
In contrast, our approach, making use of a rotating polarizing filter, exploits the full spatial resolution of the event camera and does not require camera motion.
Since event cameras have a high temporal resolution, this approach enables us to rotate the filter at very high speeds (upto $1500$ RPM), thus enabling high-speed and high spatial resolution SfP.

\vspace*{-10pt}
\paragraph{SfP Datasets}
Prior SfP datasets are summarized in \Tab \ref{tab:related-work}.
The datasets were collected using standard cameras with either \textit{DoT} \cite{Kadambi15ICCV} or \textit{DoFP} \cite{Ba20ECCV, Lei22CVPR}.
Kadambi \etal \cite{Kadambi15ICCV} presented a small dataset which was collected using Canon Rebel T3i DSLR camera and a linear polarizer and images were collected at $6$ discrete polarizer angles.
The advent of the Lucid Polarsens \cite{Lucid18} camera enabled the generation of large scale dataset proposed  by Ba \etal \cite{Ba20ECCV} and  Lei \etal \cite{Lei22CVPR}.
These datasets capture only $4$ polarization angles as is the case with \textit{DoFP}-based methods, and do not contain events.
To push the limits of event-based SfP, in this paper, we propose two datasets, synthetic and real-world, which contain events, images and accurate groundtruth in challenging  scenarios.

%% file: sections/03_methodology.tex
\section{Event-based Shape from Polarization}
\vspace*{-4pt}

Event-cameras are novel, bio-inspired sensors that asynchronously measure \emph{changes} (i.e., temporal contrast) in illumination at every pixel, at the time they occur \cite{Lichtsteiner08ssc,Suh20iscas,Finateu20isscc, Posch11ssc}.
In particular, an event camera generates an event $e_k = (\mathbf{x}_k,t_k,p_k)$ at time $t_k$ when the difference of logarithmic brightness at the same pixel $\mathbf{x}_k=(x_k,y_k)^\top$  reaches a predefined threshold $C$:
\begin{equation}
\label{eq:egm}
    L(\mathbf{x_k},t_k) - L(\mathbf{x_k},t_k-\Delta t_k) = p_k C,
\end{equation}
where $p_k \in \{-1,+1\}$ is the sign (or polarity) of the brightness change, 
and $\Delta t_k$ is the time since the last event at the pixel $\mathbf{x}_k$. The result is a sparse sequence of events which are asynchronously triggered by illumination changes.

We consider the task of surface normal estimation using a polarizer and an event camera, as illustrated in \Fig \ref{fig:method_overview}.
The polarizer, rotating in front of the event camera at speed $\omega$ using a motor, changes the illumination of the incoming light. Given the orientation $\phi_\text{pol}$ of the polarizer, we can express the light intensity $I(\phi_\text{pol})$ passing through it as:
\begin{equation}
    I(\phi_\text{pol}) = I_\text{un}(1 + \rho\cdot \cos (2(\phi_\text{pol} - \phi)),
\end{equation}
where $I_\text{un}$ is the unpolarized intensity, $\rho$ is the degree of polarization and $\phi$ is the angle of polarization of the surface.
When rotating the polarizer, the sinusoidal intensity changes trigger events in the event camera.
For instance, when observing two surfaces which are perpendicular to each other with our setup, the corresponding events have  opposite polarity, as shown in \Fig \ref{fig:method_overview}. This is because the underlying sinusoids are phase shifted by $90 ^\circ$.

We take advantage of the close relationship between surface orientation and event generation to develop a principled procedure for estimating surface normals. By continuously capturing how light is affected by the polarizer as it rotates, events offer increased information than what is provided by regular frame-based devices. %

Based on these considerations, we start in \Sec \ref{sec:psfp} by describing how events can directly be used to estimate the normals in a physics-based fashion. This approach, however, assumes events are generated following an ideal model, which might limit its performance when edge-cases are met in the real world due to nonidealities. 
Leveraging these findings, we also propose a deep learning method that improves predictions in \Sec \ref{sec:lsfp}.

\subsection{Physics-based ESfP}
\label{sec:psfp}
Over one rotation of the polarizer, the intensity at each pixel $\mathbf{x}_p$ follows a sine wave, as depicted in \Fig \ref{fig:method_overview}. By combining this behavior with the event generation model in \cref{eq:egm}, we can estimate the intensity $I_e(\mathbf{x}_p)$ at any time $\tau$ as:
\vspace{-9pt}
\begin{align}
     L(\tau) &= L(\tau-\delta t)+ p_{t} C =  L(0) + \sum_{t=0}^{\tau} p_{t} C 
\label{eq:ev_intensity} \\[2pt]
     I_e(\tau) &= \exp\left(L(0)\right) \cdot \exp\Big(\sum_{t=0}^{\tau} p_{t} C\Big),
\end{align}
where we dropped the pixel location $x_p$ from the logarithmic brightness for readability.

Since events only capture relative intensity changes and not absolute ones, there is no way to recover the constant $ e^{L(0)}$ without direct intensity measurements.
However, we show below that for the computation of the surface normal, the knowledge of absolute intensity is not required, and we can thus avoid estimating $ e^{L(0)}$.

Since $\phi_{pol} = \omega \cdot t$, we can use the above formulation to reconstruct ``event-intensities" ($I_e$) \emph{at multiple angles} $\phi_{pol}$, rather than time instants. 
These ``event-intensities" are then used to estimate the surface normal using traditional algorithms.
Below, we show the estimation of $\rho$ and $\phi$ using the event intensities computed at $4$ polarizer angles as an example, where we change the notation from from $I_e(\cdot)$ to $I_e[\cdot]$ to highlight the change in variable from time to angles. Notice that, in practice, we estimate event intensities at $12$ polarizer angles. Nonetheless, in both cases, we can demonstrate that estimating surface normal does not require the knowledge of absolute intensity.
The SfP equations\cite{Kadambi15ICCV, Ba20ECCV, Lei22CVPR} can be substituted with our event intensities as follows: 
\begin{align}
    \phi &= 0.5 \cdot \arctan \frac{I_e[\sfrac{\pi}{4}]-I_e[\sfrac{3\pi}{4}]}{I_e[0]-I_e[\sfrac{\pi}{2}]} \\
         &= 0.5 \cdot \arctan \frac{e^{\sum_{t=0}^{\pi/4\omega} p_{t} C - \sum_{t=0}^{3\pi/4\omega} p_{t} C}}{-\sum_{t=0}^{\pi/2\omega} p_{t} C}
\end{align}
Similarly, we can estimate $\rho$ as,
\begin{equation}
    \rho = 0.5 \cdot \arctan \frac{\sqrt{(I_e[\sfrac{\pi}{2}]^2 + (I_e[\sfrac{\pi}{4}]^2 - I_e[\sfrac{3\pi}{4}]^2)}}{I_e[\sfrac{\pi}{2}]+I_e[\sfrac{\pi}{4}]+I_e[\sfrac{3\pi}{4}]},\\
\end{equation}
\begin{equation}
    \nonumber = 0.5 \cdot \arctan \frac{\sqrt{e^{2\sum\limits_{t=0}^{\pi/2\omega} p_{t} C} + \raisebox{2mm}{\Big(}e^{\sum\limits_{t=0}^{\pi/4\omega} p_{t} C} - e^{\sum\limits_{t=0}^{3\pi/4\omega} p_{t} C}\raisebox{2mm}{\Big)}^2}}{ e^{\sum\limits_{t=0}^{\pi/2\omega} p_{t} C} +  e^{\sum\limits_{t=0}^{\pi/4\omega} p_{t} C} + e^{\sum\limits_{t=0}^{3\pi/4\omega} p_{t} C}}
\end{equation}

Using the above $\rho$ and $\phi$, the azimuth ($\alpha$) and zenith angle ($\theta$) of the surface normal can be estimated. We refer the reader to the supplementary materials for a complete derivation, which takes into account all the $12$ angles we consider in our approach and the estimation of surface normal from $\rho$ and $\theta$. 
Since we reconstruct event intensities, any traditional algorithm developed for frame-based approaches can be utilized along with our approach.
This opens the door to build upon the past research and develop better phyiscs-based algorithms from events.

\vspace*{-9pt}
\paragraph{Event non-idealities}
\label{sec:psfp:noise}
An event camera, like any physical sensor, is affected by noise caused by nonidealities of its individual electric circuits and the manufacturing process.
The most important parameter that affects the noise generation model is the contrast sensitivity threshold $C$.
If $C$ is too high, this results in less number events triggered, thus increasing the signal to noise ratio; however, if it $C$ is set too low, a higher number of noisy events are generated.
In our application, it is favourable to have a low contrast threshold, in the order of 1-5\% as the polarization signal is often weak (especially in low light conditions).
Commercial available event sensors \cite{propheseeevk} however, are manufactured for a general purpose use of event cameras, and therefore have a very high threshold, typically ranging between $30$\% and $50$\%\footnote{The contrast threshold can be quantified as the percentage of intensity change with respect to the previous intensity value.}.
This directly affects the number of events that are triggered (or the fill-rate), resulting in a sparse estimation.
Since the polarization signal is especially weak for front-parallel surfaces in real world scenes, we find that these surfaces rarely trigger any event.
To address this sensor issue, we propose to tackle the problem using a data-driven approach.

\subsection{Learning-based ESfP}
\label{sec:lsfp}
We now describe our approach to solve the problem of shape-from-polarization using deep learning.
As described above, events can be used to reconstruct event intensities at multiple polarization angles.
Inspired by the frame-based solution of Ba \etal \cite{Ba20ECCV,Lei22CVPR}, we propose to tackle the problem of estimating the surface normals using a U-Net \cite{Ronneberger15icmicci} architecture.
The network predicts the surface normal $\mathbf{N}$ as a 3 channel tensor $\mathbf{N} = (\sin \theta \, \cos \alpha, \sin \theta \,\sin \alpha, \cos \theta)$, 
where $\theta$ and $\alpha$ are the zenith and azimuth angle respectively.

To adopt this network to event data, we need to convert the sparse event stream into a frame-like event representation that is compatible with the CNN layers. A straightforward approach, which demonstrated remarkable performance on several tasks \cite{Rebecq19pami, Hidalgo203dv, Gehrig213dv, Muglikar213dv2}, is to use the voxel grid \cite{Zhu18eccvw} representation as input to the encoder network.
This representation splits the temporal domain into bins, and aggregates events in each pixel as the sum of their polarity.
However, doing so, empirically results in a higher mean absolute error, most likely due to events with opposite polarity in the same bin nullifying each other, and thus resulting in an empty feature. Please refer the supplementary material for more details.
We therefore design a novel event representation specifically suited for the SfP task.
In \Eq \ref{eq:ev_intensity}, we observe that the accumulated sum of event polarities is directly related to the image intensity at a given time instant.
Based on this consideration, we propose the \emph{Cumulative Voxel Grid Representation} (CVGR), a variation of the voxel grid \cite{Zhu18eccvw} that preserves polarity information even across long time intervals.
Similar to previous works on learning with events \cite{Rebecq19pami, Muglikar213dv2, Tulyakov21cvpr}, we first build a voxel grid \cite{Zhu18eccvw} by partitioning the time domain into $b=0,...,B-1$ equally-sized temporal windows, each aggregating the sequence of events $\mathcal{E}_b = \left\{e_k \mid t_k \in \big[\frac{b\cdot T}{B}, \frac{(b+1) \cdot T}{B}\big] \right\}$.
We then compute the cumulative sum over the bins and multiply this quantity with the contrast threshold.
\vspace*{-9pt}
\begin{equation}
    E(x, y, b) = \sum_{i=0}^{i=b} C \cdot V(x, y, i) = \sum_{i=0}^{i=b} C \Big(\sum_{\!\!\!\!\!\substack{e_k \in \mathcal{E}_i :\\ x_k=x, y_k=y }}{\!\!\!\!p_k}\Big).
\end{equation}

This final representation $E$ is used as an input to the network.
In our experiments, we use a voxel grid with $8$ bins, thus our CVGR has the dimension $ H \times W \times 8$.

This novel event-representations is then fed to a $8$-level U-Net encoder-decoder architecture which densely regresses the normals. We apply cosine similarity loss function on the unit normalized predictions for training. We show this simple architecture is enough to recover high-quality normal vectors, even in locations where no event has been triggered, thus directly addressing the weakness of purely model-based solutions. 

%% file: sections/04_experiments.tex
\vspace*{-6pt}
\section{Datasets}
To the best of our knowledge, there exists no dataset on which the proposed approach can be evaluated.
We, therefore, first evaluate our approach on a synthetic dataset, rendered using an accurate polarimetric renderer \cite{jakob22mitsuba3} to perform controlled experiments. 
To evaluate our approach in the real world, we also build a prototype of the proposed setup using an event camera and a polarizing filter. 
We provide more details in the next sections.

\subsection{ESfP-Synthetic Dataset}
We use the Mitsuba renderer \cite{jakob22mitsuba3} to render a scene consisting of a textured mesh illuminated with a point light source.
A polarizer lens was rotated in front of the camera between $0$ and $180$ degrees with 15 degrees intervals, resulting in 12 rendered images.
Images thus collected were then used to simulate events using ESIM \cite{Rebecq18corl} with a contrast sensitivity threshold of 5\%.
The dataset consists of $89$ training and $15$ test sequences, each containing the images captured with the polarizer, together with the events generated from these, as well as the groundtruth surface normal provided by the renderer.
Meshes used from data generation were obtained from the \textit{Google Scanned Objects} dataset \cite{Downs22arxiv}  and the textures derived from \cite{Baek20siggraph}.
More details are provided in the supplementary material.
To increase data variability, the position of the light source was randomized and the mesh  rotated for each scene.

\subsection{ESfP-Real Dataset}
\input{images/fig_hardware}
To evaluate our approach in the real world, we built a prototype using a Prophesee Gen 4 event camera \cite{Finateu20isscc} and a \emph{Breakthrough Photography X4 CPL} \cite{BreakthroughX4} linear polarizer with a quarter-wave plate, commonly referred to as a circular polarizer (CPL).
The prototype is shown in \Fig \ref{fig:hwsetup}.
The polarizer can rotate at speeds of up to \unit[1500]{rpm} using a brushless DC motor.
The groundtruth is generated using Event-based Structured Light (ESL) \cite{Muglikar213DV}, which consists of a laser point projector combined with an event camera. 
It provides accurate groundtruth thanks to laser scanner, while also having a small acquisition time (\unit[16]{ms}). 
Another advantage of using ESL is that the groundtruth is always in the frame of the event camera and therefore no additional alignment is required.

We use this setup to collect the \textit{first} large scale dataset consisting of several objects with different textures and shapes, and featuring multiple illumination and scene depths, for a total of $90$ scenes.
Additionally, we use a Lucid Polarisens camera \cite{Lucid18} to collect polarization images of the same scene.
The images are obtained at $4$ polarization angles \{$0$, $45$, $90$, $135$\} with a resolution of $1224 \times 1220$. 
We rectify the images and aligned them with the event camera  to enable fair comparison with available image-based only methods.
The dataset was recorded with the rotation speeds of \unit{150}{RPM} with illumination of \unit{200}{lux}.
\vspace*{-2pt}
\section{Experiments}
This section evaluates the performance of our event-based SfP system for surface normal estimation. 
We begin by introducing the baselines and performance metrics which will be used for evaluation on our dataset. 
Then, we perform experiments on the ESfP-Synthetic dataset to quantify the accuracy of our proposed physics-based and learning-based methods.
Finally, we evaluate these approach on our ESfP-Real dataset and study the proposed approach on real scenes under challenging conditions.
\vspace*{-9pt}
\paragraph{Implementation details}
We implement our physics-based approach in Python.
As noted in \Sec \ref{sec:psfp}, we reconstruct event intensities at $12$ polarizer angles.
Our physics-based approach does not require any hyperparameter tuning.
The proposed learning-based approach was implemented in PyTorch.
We train our network for 1000 epochs with a batch size of 4 on NVIDIA Tesla V100.
We used a learning rate of $1e-4$ and Adam \cite{Kingma15iclr} as optimizer. 

\input{floats/fig_qual_mitsuba}
\input{floats/tab_mitsuba}
\vspace*{-9pt}
\paragraph{Evaluation Metrics}We use $4$ metrics to quantify the predicted surface normals, namely \textit{Mean Angular Error} (MAE), \textit{$\%$ Angular Error under $11.25 ^\circ$} (\aeone), \textit{$\%$ Angular Error under $22.5 ^\circ$} (\aetwo) and \textit{$\%$ Angular Error under $30 ^\circ$} (\aethree). 
The first is a widely used metric for computing the angular error of the predicted surface normal (lower is better) \cite{Ba20ECCV, Lei22CVPR}, while the last three, which we also refer to as angular accuracy, measure the percentage of pixels under  $11.25 ^\circ$,  $22.5 ^\circ$ and $30 ^\circ$ of angular error (higher is better).
We introduce another metric called \textit{fill-rate} to compare events and images.
The fill-rate measures the percentage of pixels that are triggered by SfP, i.e., by the polarizing filter's rotation.
For image-based approaches, this always equals to $1$, since images capture intensity for every pixel. However, for events this becomes an important parameter, since the number of triggered pixels depends on the contrast sensitivity of the camera as explained in \Sec \ref{sec:psfp:noise}.

\vspace*{-9pt}
\paragraph{Baselines} %
We evaluate the proposed methods against frame-based SfP solutions. 
As we propose to tackle the problem both in a model-based and a data-driven manner, we include state-of-the-art methods from both categories as baselines in our evaluation. 
Smith \etal \cite{Smith19PAMI} and Mahmoud \etal \cite{Mahmoud12ICIP} recover the surface normal using physics-based SfP methods\footnote{https://github.com/waps101/depth-from-polarisation}, whereas Ba \etal \cite{Ba20ECCV} uses a learning-based SfP method.
We re-implemented the last approach and retrained it on our datasets using only the images as input.
\input{floats/fig_qual_plane}
Note, while Lei \etal \cite{Lei22CVPR} improved upon the works of Ba \etal, by introducing view-encoding as an input to the network to generalize better to natural scenes, the goal of this paper is to show the performance of our learning-based framework on object-level scenes used in Ba \etal.
We therefore, do not compare our approach against Lei \etal.

The event-based SfP solution proposed in \cite{Gruev22SPIE} is not commercially available and requires specific manufacturing in order to be replicated.
Moreover, this sensor requires relative motion between the scene and the camera and would not result in any surface normal estimation in our experiments.
We, therefore, do not include comparison to this method in our experiments.
In the following sections, we will first evaluate our methods in a simulated setting where precise ground truth allows for an accurate evaluation, and then proceed by studying their performance on the real world dataset we collected.

\subsection{Results on ESfP-Synthetic}
\Tab \ref{tab:comparison} shows the performance of our event-based approaches (both geometry-based and learning-based SfP) on the synthetic dataset.
As it can be observed, frames-based solutions leveraging a physics-based approach are limited by the number of polarization images (and thus angles) which can be used for inference. When the same approach is used, i.e., in the case of Smith \etal \cite{Smith19PAMI}, simply increasing the number of images from $4$ to $12$ decreases the angular error by nearly $8\%$. %
This gain in performance, however, comes at the cost of either spatial resolution or acquisition speed. In the case of a DoT approach, for instance, because the acquisition time is linearly proportional to the number of frames acquired, increasing the angles from $4$ to $12$ triples the acquisition time at a given frame-rate.

On the other hand, our approach is not limited by this tradeoff thanks to the high-temporal resolution of the event camera. Exploiting the continuous stream of events acquired when observing the light being polarized, our geometry-based approach outperforms other physics-based  methods using $4$ images by $14\%$ in angular error. %
In \Fig \ref{fig:qual_mitsuba}, we show qualitative results of our approach against the baselines.
While our learning-based approach boosts the performance compared to physics-based methods significantly, it falls short of the image-based counterpart.
The reason is because the learning-based method tries to hallucinate the surface normal prediction in the absence of events, whereas the image counterpart still has a dense spatial information to guide the network to interpolate the surface normals.
In future, it would be worth investigating a combination of events and images to improve the performance.

\subsection{Results on ESfP-Real}

\input{floats/fig_qual_real}
\input{floats/tab_realdata}
\input{floats/fig_qual_learning_real}
We also compare these methods on the real dataset in \Tab \ref{tab:realdataset} and report qualitative results in \Fig \ref{fig:qual_real}. We observe similar conclusions as with the synthetic dataset. The proposed geometry-based approach outperforms previous physics-based solutions by $25.8\%$ in angular error.%
As mentioned in \Sec \ref{sec:psfp:noise}, a real event camera introduces several non-idealities, one of them being the contrast sensitivity. With a typical event camera the contrast sensitivity cannot be usually set lower than $30\%$, resulting in the inability to capture small intensity changes. An example of this phenomenon is given in \Fig \ref{fig:qual_learn_real}, where the presence of fronto-parallel geometries causes a polarization signal and therefore a small fill-rate metric.
On average, for this dataset, the fill-rate is around $3.6\%$.
This, however, is exactly the condition where the proposed learning based approach excel. 
By leveraging learned data-priors, it can recover accurate normals even in the presence of very few events. 
The improvements in both angular error and accuracy, show the proposed method outperforms all baselines, despite falling short by a small margin to Ba \etal \cite{Ba20ECCV}. 
We emphasize, however, that these results were obtained under ideal lighting conditions, where data-priors of learning-based approaches are enough to compensate for the limited polarization information that standard sensors can provide. We now show the performance of our methods when these perfect conditions are not satisfied against baselines.

\vspace{-15pt}
\paragraph{Effect of illumination}
\input{floats/fig_illum_example}
\Fig \ref{fig:illum_real} highlights the high dynamic range (HDR) advantage of using an event camera for SfP under different illumination conditions.
We increased the illumination from \unit[528]{Lux} to \unit[1442]{Lux} and evaluated the performance our proposed approach and compared it to the image-baselines.
The corresponding images and events are shown.
As it can be noticed, the performance of our methods remains consistent across different illumination conditions. This is case for both the model-based and learning-based variants, highlighting that events provide more information than images, and are thus more suited for these challenging conditions.

\vspace*{-9pt}
\paragraph{Effect of speed}
In this section, we assess the performance of our method under different rotation speeds of the motor.
In \Tab \ref{tab:realdataset_speed}, we find the performance of both our methods improves slightly with an increase in the speed.
This can be explained as at higher rotation speeds, the rate of intensity change is higher, which makes non-linearities and non-idealities in the event-camera's less prominent, resulting in a better quality of events.
Please refer to the suppl. material.

\subsection{Generalization to outdoor scenes}
In \Fig \ref{fig:qual_outdoor}, we show that our method generalizes to outdoor scenes as well, despite being trained on single-object and near-depth scenes.
Fine tuning on this dataset is not possible due to the unavailability of any groundtruth.
In the future, the proposed work would benefit by adopting a view-encoding proposed in \cite{Lei22CVPR} here.

\input{floats/tab_speed_effect}

\input{floats/fig_outdoor}

%% file: images/fig_hardware.tex
\begin{figure}
    \centering
   \includegraphics[width=1\linewidth]{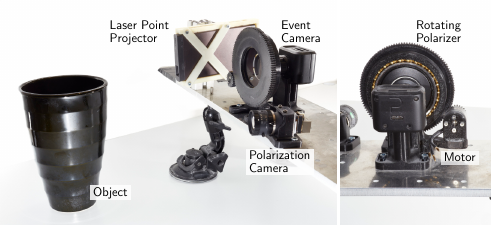} 
   \vspace*{-15pt}
    \caption{Hardware setup of event camera coupled with a rotating linear polarizer.}
   \vspace*{-15pt}
    \label{fig:hwsetup}
\end{figure}

%% file: floats/fig_qual_mitsuba.tex
\begin{figure}[!t]
	\centering
    \setlength{\tabcolsep}{2pt}
	\begin{tabularx}{1\linewidth}{CCCCCC}
		Image & Events & \cite{Mahmoud12ICIP} & \cite{Smith19PAMI} & Ours & GT
		\\
        \mae{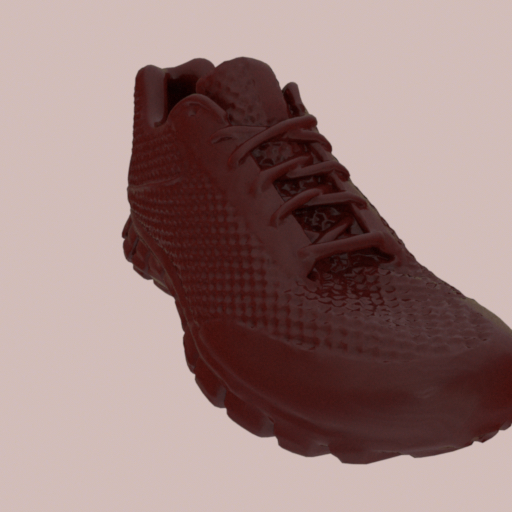}{1\linewidth}{}{none}
		&\mae{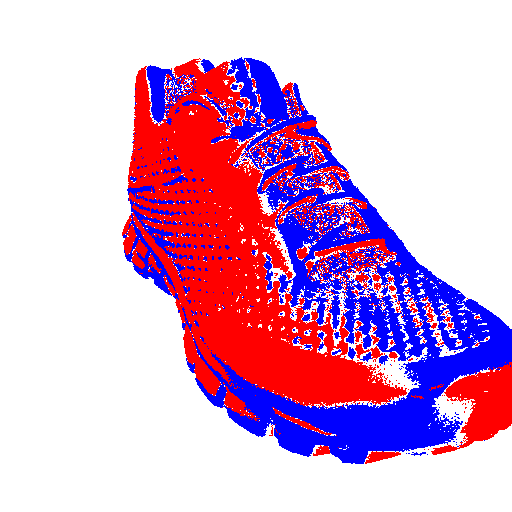}{1\linewidth}{}{none}
		&\mae{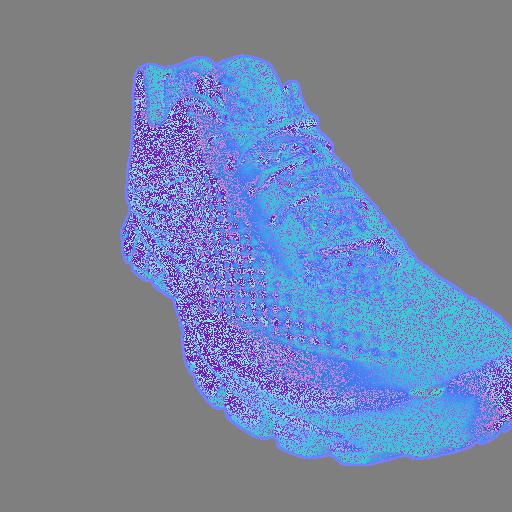}{1\linewidth}{71.9}{white}
		&\mae{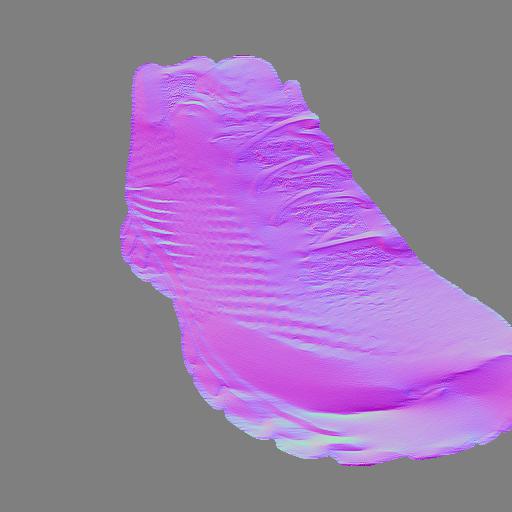}{1\linewidth}{64.0}{white}
		&\mae{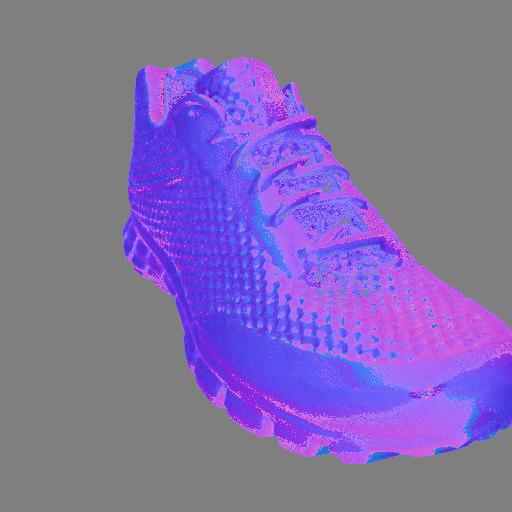}{1\linewidth}{39.4}{white}
		&\mae{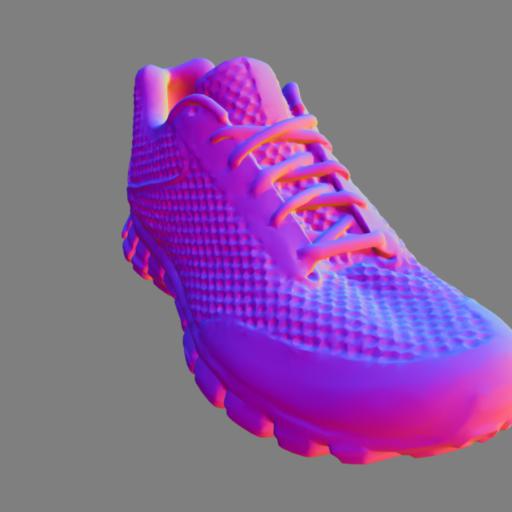}{1\linewidth}{}{none}
		\\
		\mae{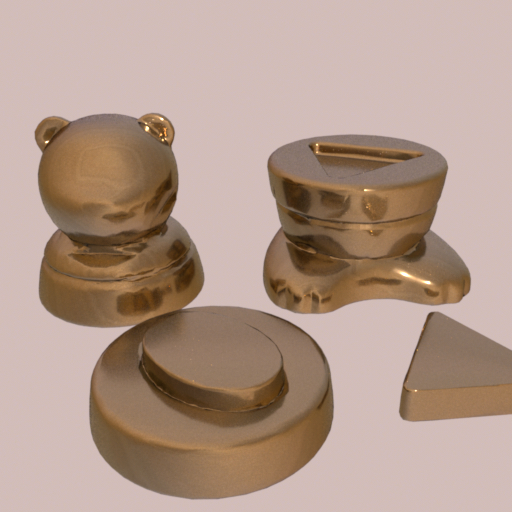}{1\linewidth}{}{none}
		&\mae{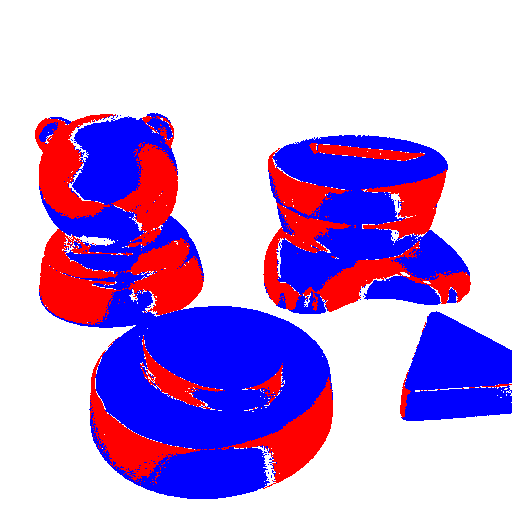}{1\linewidth}{}{none}
		&\mae{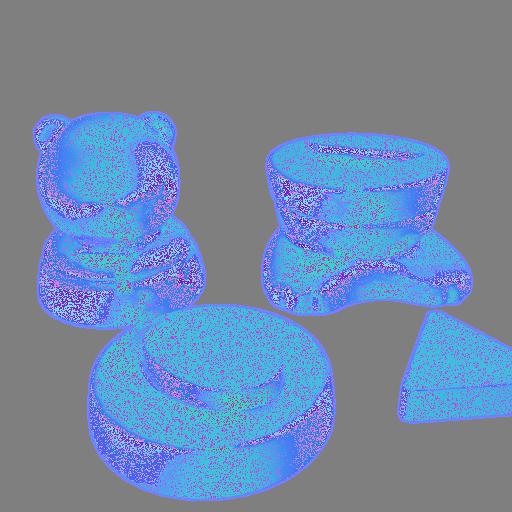}{1\linewidth}{74.7}{white}
		&\mae{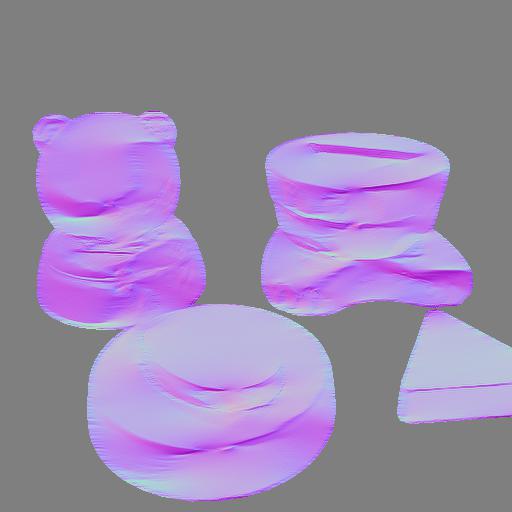}{1\linewidth}{70.8}{white}
		&\mae{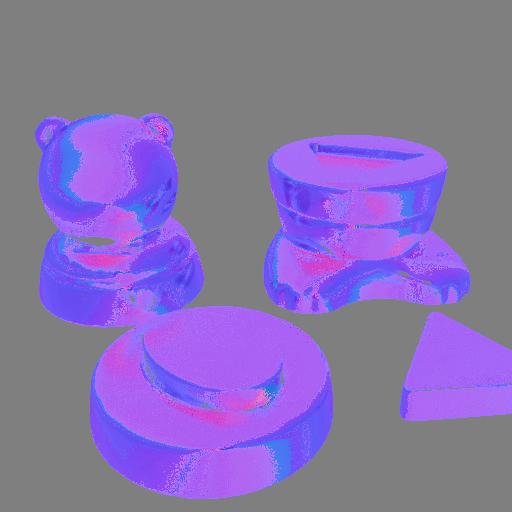}{1\linewidth}{60.0}{white}
		&\mae{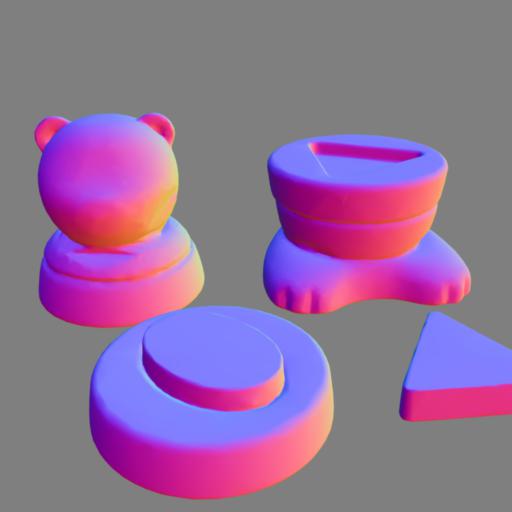}{1\linewidth}{}{none}
		\\
		\mae{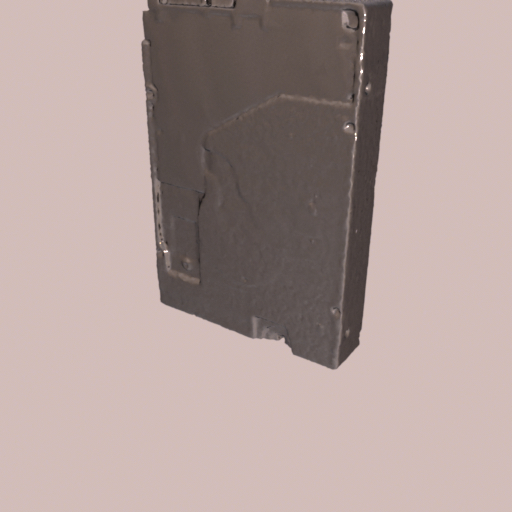}{1\linewidth}{}{none}
		&\mae{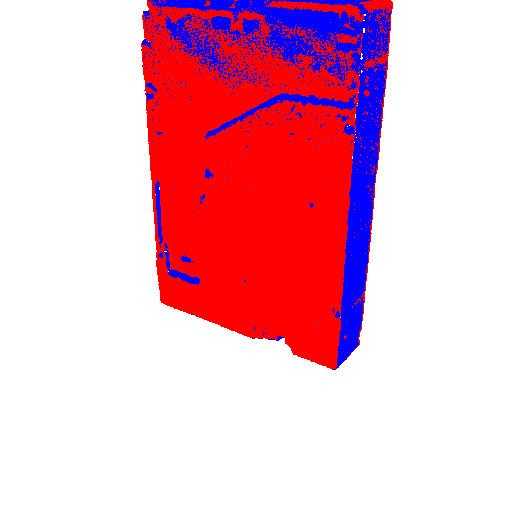}{1\linewidth}{}{none}
		&\mae{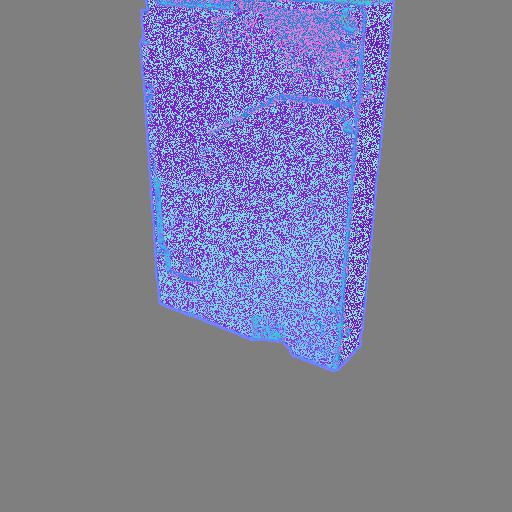}{1\linewidth}{89.9}{white}
		&\mae{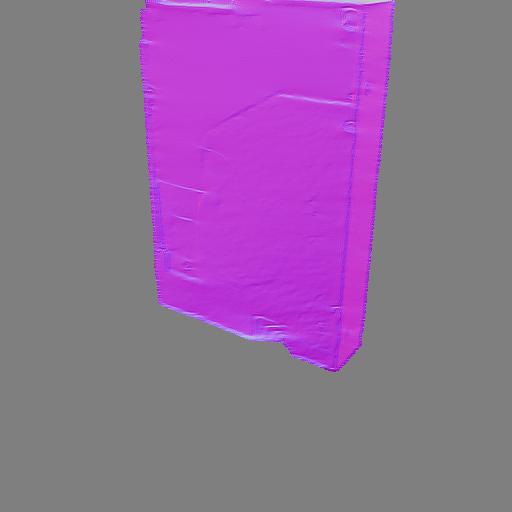}{1\linewidth}{78.3}{white}
		&\mae{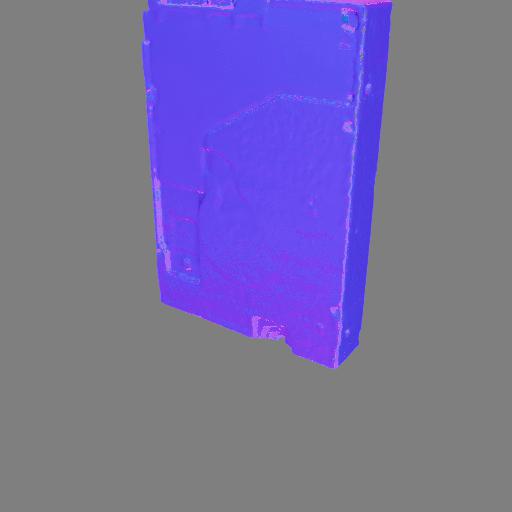}{1\linewidth}{28.3}{white}
		&\mae{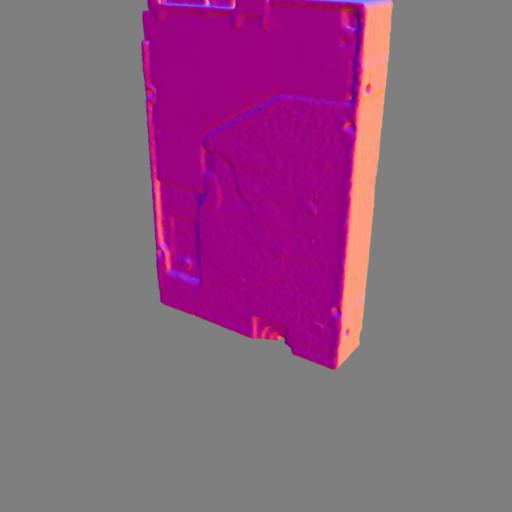}{1\linewidth}{}{none}
		\\

	\end{tabularx}
    \vspace*{-12pt}
	\caption{Comparison of physics-based methods on ESfP-Synthetic. We report the angular error on the top-left corner of each picture. 
	}
	\label{fig:qual_mitsuba}
	\vspace{-2ex}
\end{figure}

%% file: floats/tab_mitsuba.tex
\begin{table}[!t]
    \centering
    \begin{adjustbox}{max width=\linewidth}
    \setlength{\tabcolsep}{4pt}
    {\small
    \begin{tabular}{|l|c|c|c rrr|}
        \toprule
         Method & Input & Task & Angular Error $\downarrow$  &  \multicolumn{3}{c|}{Accuracy $\uparrow$} \\
         & & & Mean & \aeone & \aetwo & \aethree \\
        \midrule
        Mahmoud \etal \cite{Mahmoud12ICIP} & 4 I & Physics-based & 80.923 & 0.034 & 0.065 & 0.085 \\
        Smith \etal \cite{Smith19PAMI} & 4 I & Physics-based & 67.684  & 0.010 & 0.047 & 0.106\\
        Smith \etal \cite{Smith19PAMI} & 12 I& Physics-based  & 62.476 & 0.007 & 0.043 & 0.097\\
        \textbf{Ours (P)} & E &  Physics-based & \underline{58.196} & \underline{0.007} & \underline{0.046} & \underline{0.095} \\ 
        \midrule
        Ba \etal \cite{Ba20ECCV} & 4 I & Learning-based & \textbf{24.509} &\textbf{ 0.357} & \textbf{0.623} & \textbf{0.718}  \\ 
        \textbf{Ours (L) } & E &  Learning-based& 27.953 & 0.263 & 0.527 & 0.655\\ 
        \bottomrule
    \end{tabular}}
    \end{adjustbox}
    \vspace*{-6pt}
    \caption{State-of-the-art comparison on ESfP-Synthetic in terms of accuracy and angular error. The \emph{Input} column report whether the method uses events (E) or images (I), in which case the number of frames is also reported. We underline best results among the physics-based methods, and use bold text for learning-based ones.
    }
    \label{tab:comparison}
    \vspace*{-16pt}
\end{table}

%% file: floats/fig_qual_plane.tex
\begin{figure}[t]
	\centering
    \setlength{\tabcolsep}{2pt}
	\begin{tabularx}{1\linewidth}{CCCC}
		Image & Ba \etal \cite{Ba20ECCV} & Ours (L) & GT
		\\
		 \mae{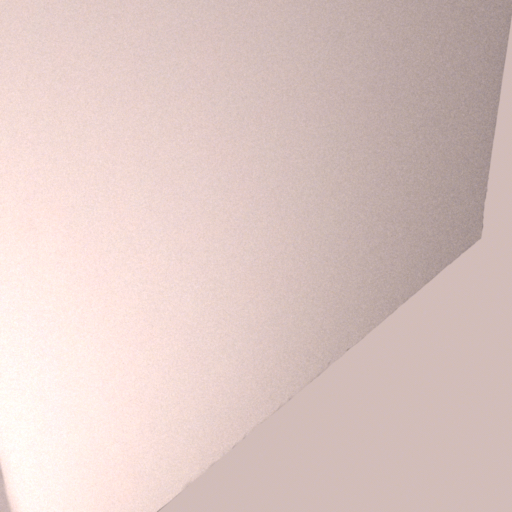}{1\linewidth}{}{none}
		&\mae{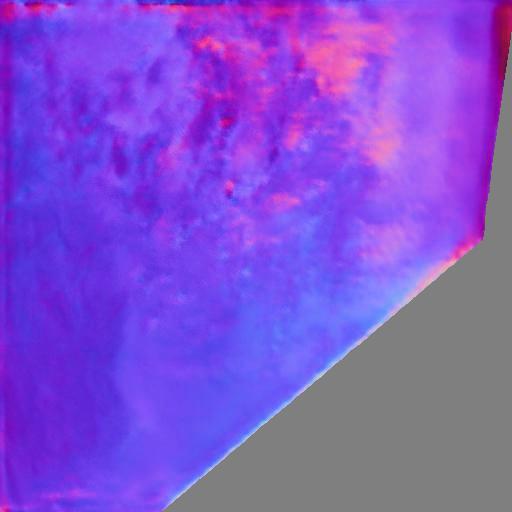}{1\linewidth}{}{none}
		&\mae{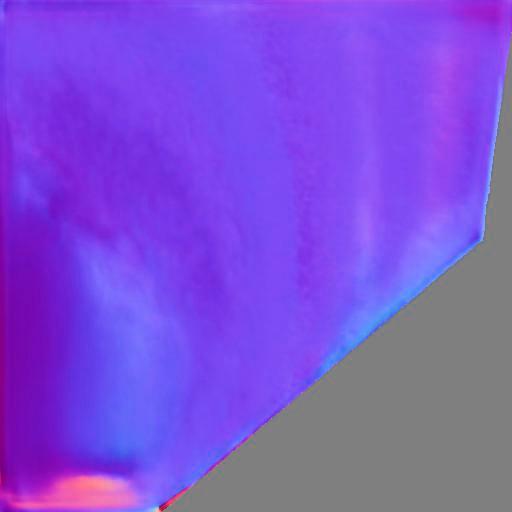}{1\linewidth}{}{none}
		&\mae{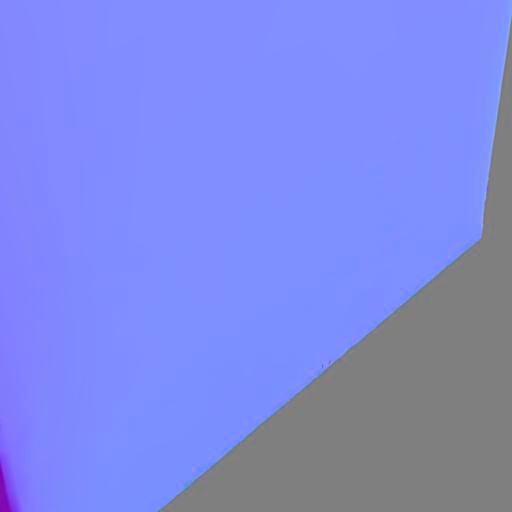}{1\linewidth}{}{none}
		\\

	\end{tabularx}
    \vspace{-1ex}
	\caption{Comparison of our learning-based method against image-counterpart on ESfP-Synthetic scene with low-contrast.}
	\label{fig:qual_mitsuba_plane}
	\vspace{-12pt}
\end{figure}

%% file: floats/fig_qual_real.tex
\begin{figure}[!t]
	\centering
    \setlength{\tabcolsep}{2pt}
	\begin{tabularx}{1\linewidth}{CCCCCC}
		Image & Events & \cite{Mahmoud12ICIP} & \cite{Smith19PAMI} & Ours (P) & GT
		\\
		 \mae{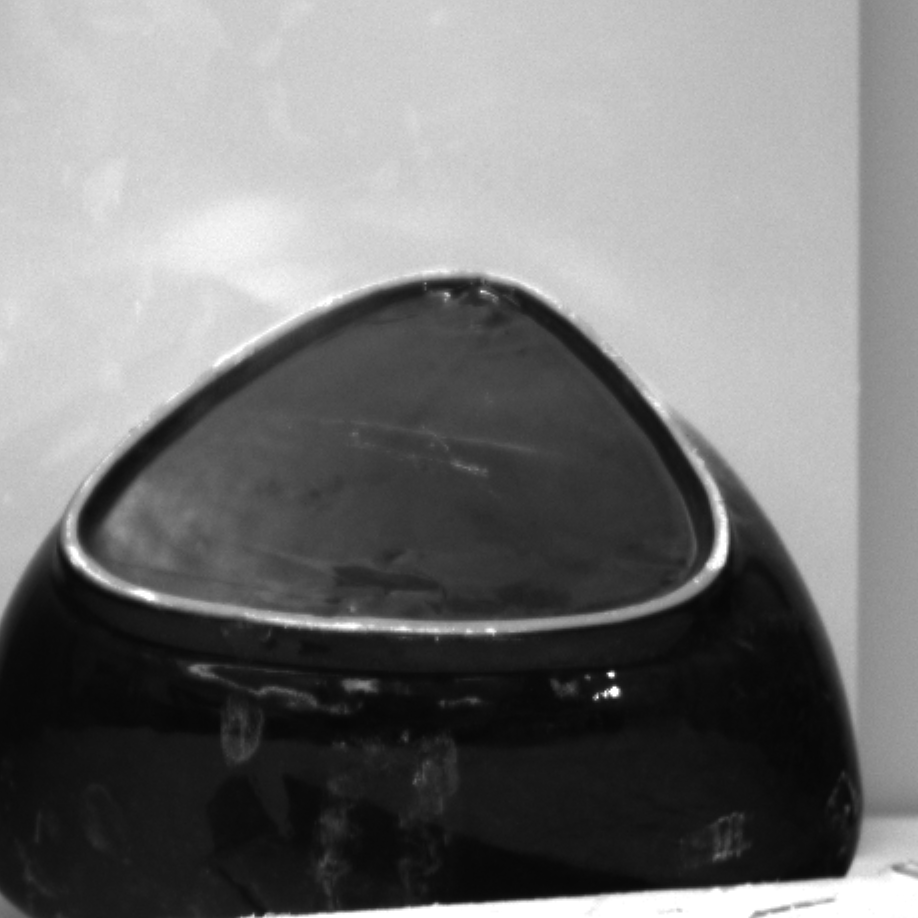}{1\linewidth}{}{none}
		&\mae{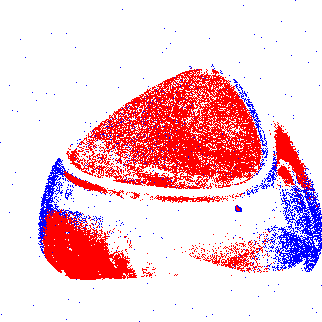}{1\linewidth}{}{none}
		&\mae{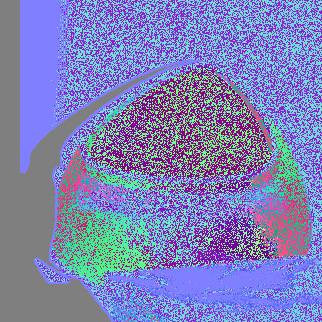}{1\linewidth}{55.96}{white}
		&\mae{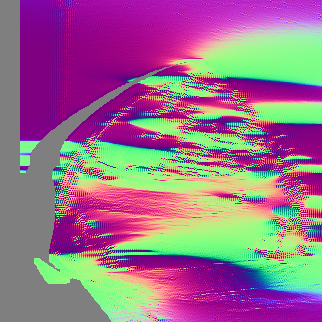}{1\linewidth}{73.85}{white}
		&\mae{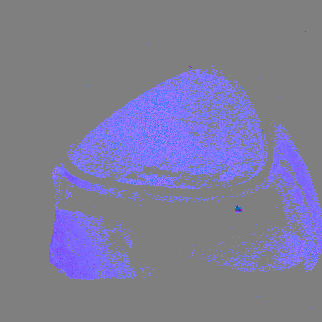}{1\linewidth}{37.38}{white}
		&\mae{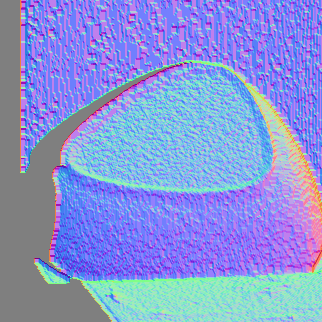}{1\linewidth}{}{none}
		\\
		 \mae{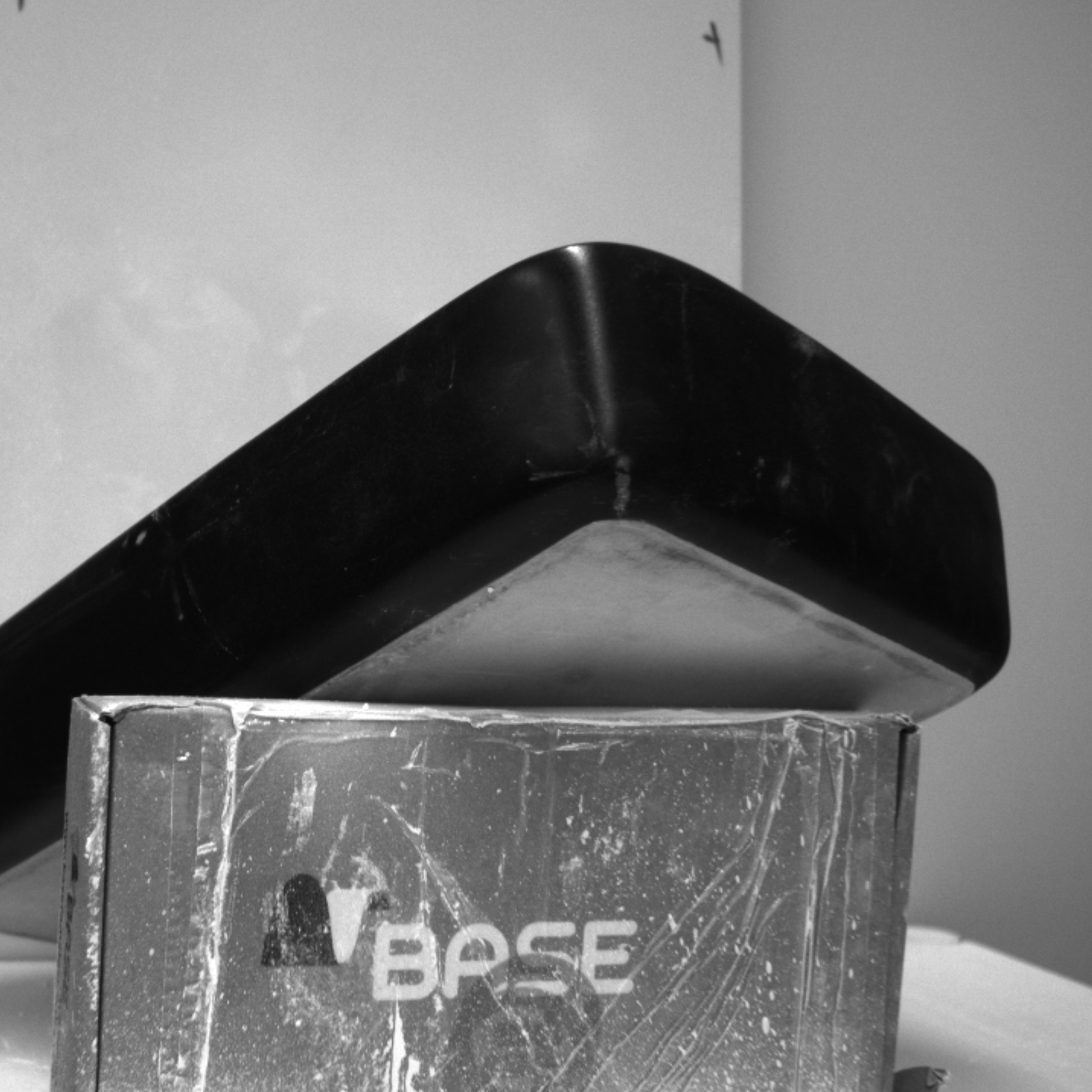}{1\linewidth}{}{none}
		&\mae{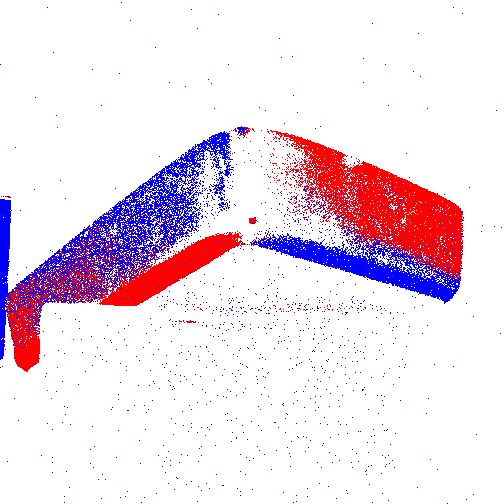}{1\linewidth}{}{none}
		&\mae{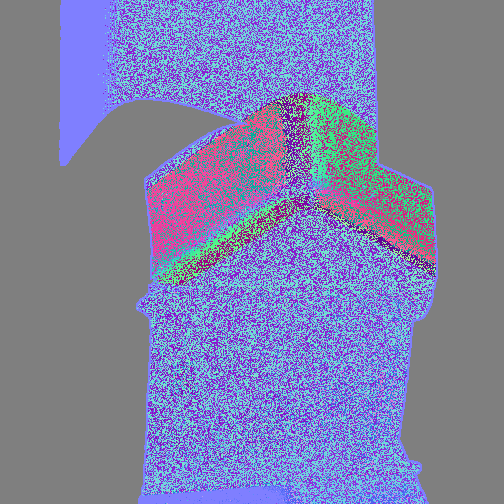}{1\linewidth}{55.62}{white}
		&\mae{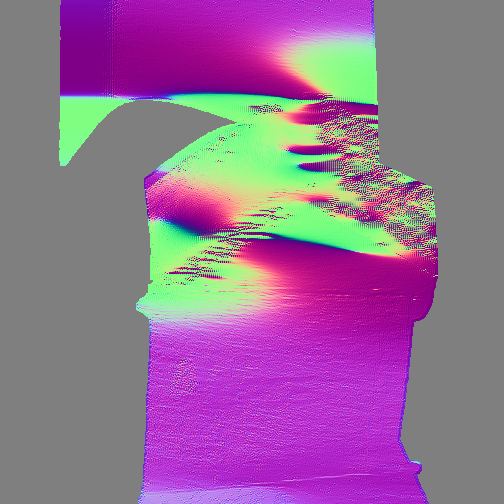}{1\linewidth}{66.68}{white}
		&\mae{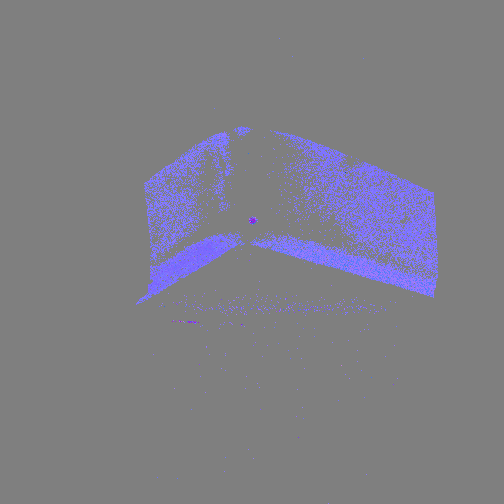}{1\linewidth}{25.94}{white}
		&\mae{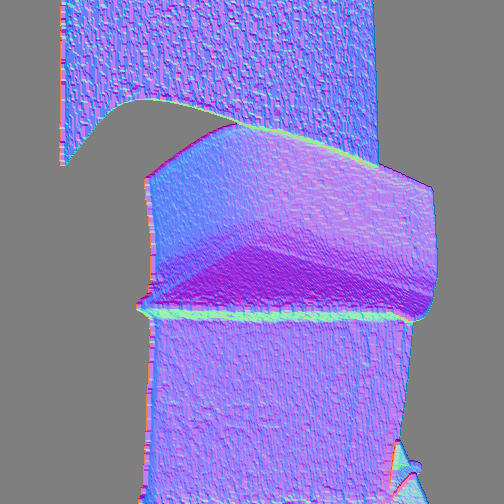}{1\linewidth}{}{none}
		\\
		 \mae{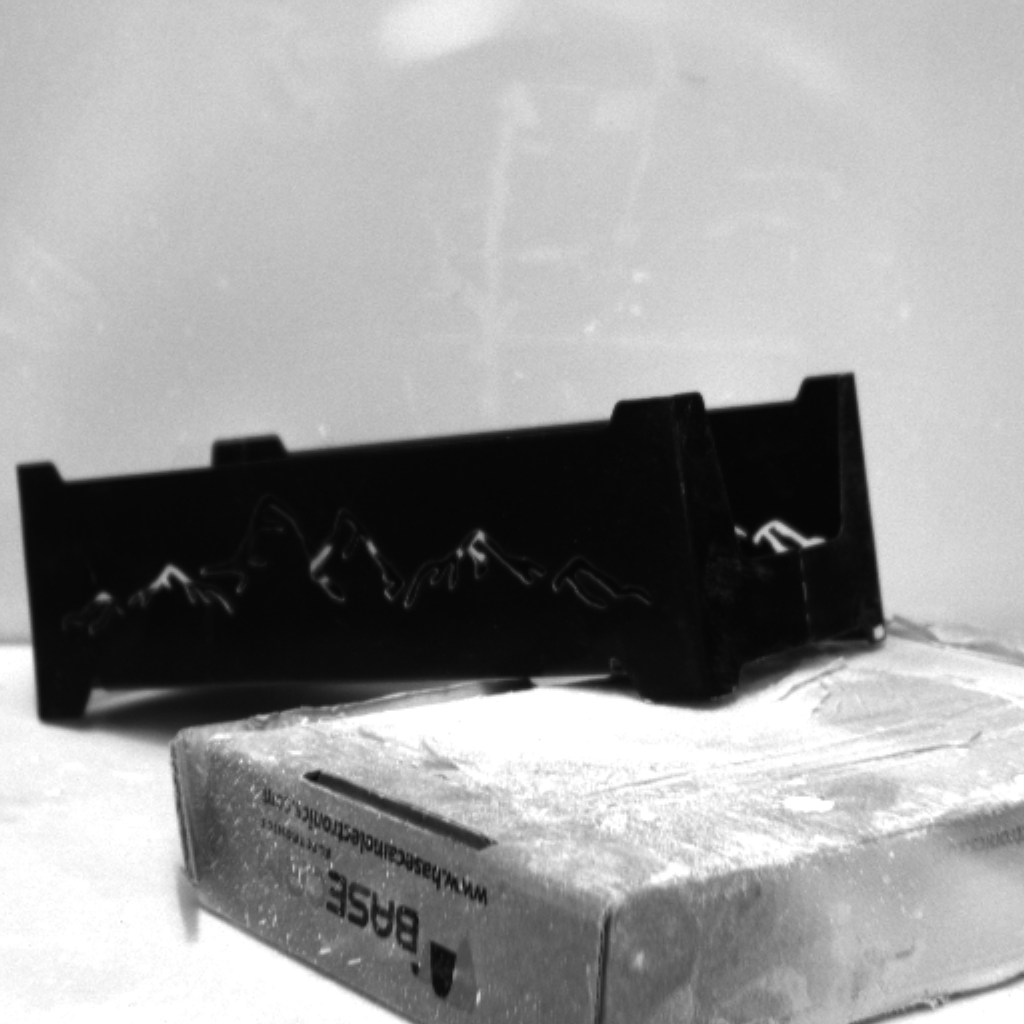}{1\linewidth}{}{none}
		&\mae{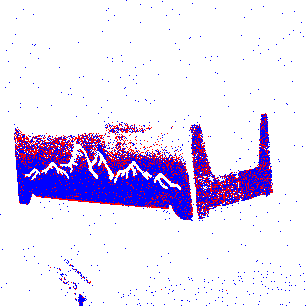}{1\linewidth}{}{none}
		&\mae{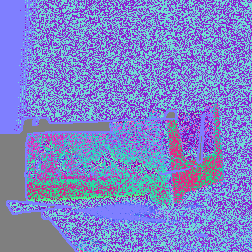}{1\linewidth}{50.54}{white}
		&\mae{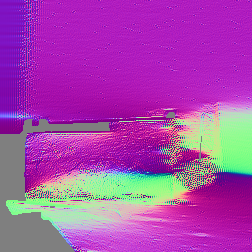}{1\linewidth}{66.73}{white}
		&\mae{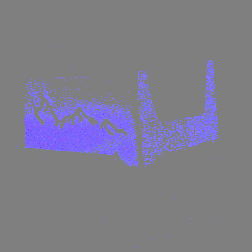}{1\linewidth}{30.61}{white}
		&\mae{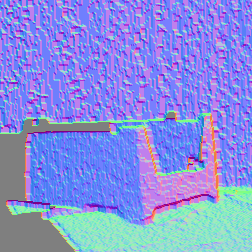}{1\linewidth}{}{none}
		\\

	\end{tabularx}
    \vspace{-1ex}
	\caption{Comparison of physics-based methods using images \cite{Mahmoud12ICIP} and \cite{Smith19PAMI} on the ESfP-Real. 
	Our physics-based method results in sparse estimation as explained in \Sec \ref{sec:psfp:noise}}
	\label{fig:qual_real}
	\vspace{-2ex}
\end{figure}

%% file: floats/tab_realdata.tex
\begin{table}[t]
    \centering
    \begin{adjustbox}{max width=\linewidth}
    \setlength{\tabcolsep}{4pt}
    {\small
    \begin{tabular}{|l|c|c rrr|}
        \toprule
        Method & Input & Angular Error $\downarrow$  &  \multicolumn{3}{c|}{Accuracy $\uparrow$} \\
        & & Mean & \aeone & \aetwo & \aethree \\
        \midrule
        Smith \etal \cite{Smith19PAMI} & I & 72.525 & 0.009 & 0.034 & 0.058\\
        Mahmoud \etal \cite{Mahmoud12ICIP} & I & 56.278 & 0.032 & 0.091 & 0.163 \\
       \textbf{Ours (P)}  & E & \underline{38.786} & \underline{0.087} & \underline{0.22} &  \underline{0.452} \\  \midrule
        Ba \etal \cite{Ba20ECCV}  & I &  \textbf{26.157} &\textbf{ 0.103 }&\textbf{ 0.467} &\textbf{ 0.710} \\
        \textbf{Ours (L)}  & E & 26.677 & 0.105 & 0.452 & 0.691 \\ \bottomrule
    \end{tabular}}
    \end{adjustbox}
    \vspace{-9pt}
    \caption{State-of-the-art comparison on ESfP-Real in terms of angular error and accuracy. We use underlined text to mark the best results within the physics-based category, and use bold text for best results among learning-based ones.}
    \label{tab:realdataset}
    \vspace{-12pt}
\end{table}

%% file: floats/fig_qual_learning_real.tex
\begin{figure}[t]
	\centering
    \setlength{\tabcolsep}{2pt}
	\begin{tabularx}{1\linewidth}{CCCCC}
		Scene & Ba \cite{Ba20ECCV} & Ours P & Ours L  & GT
		\\
		\mae{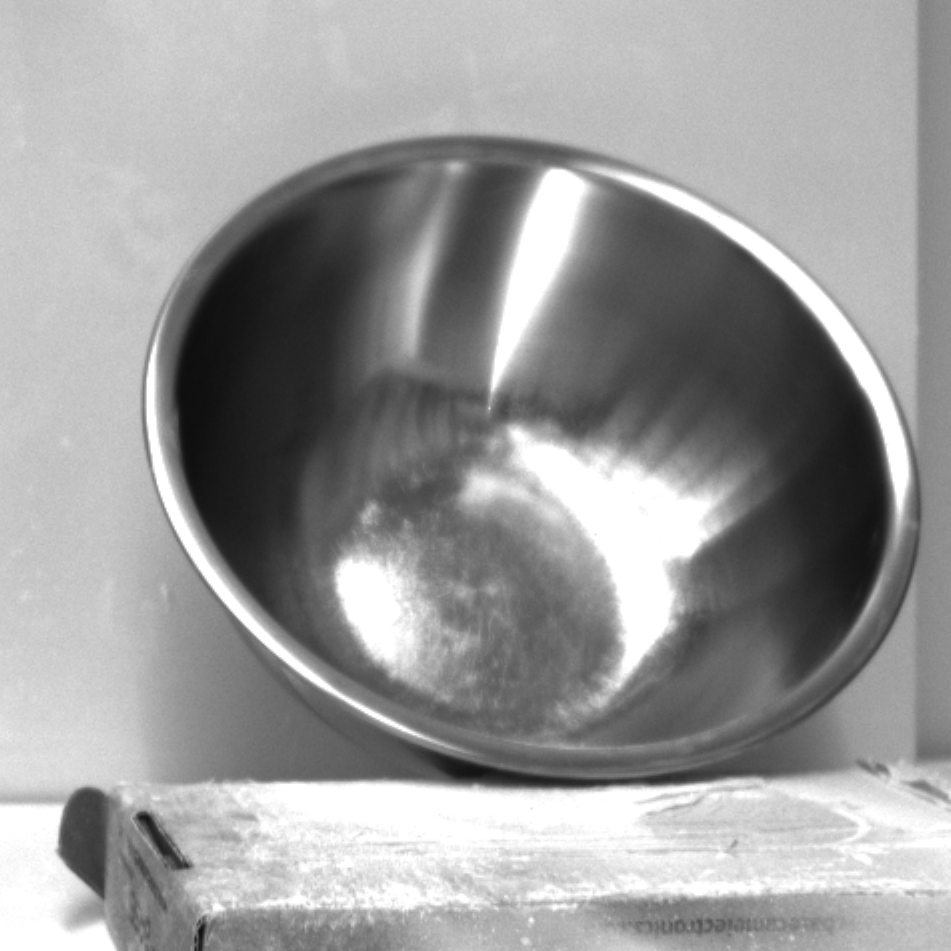}{1\linewidth}{}{none}
		&\mae{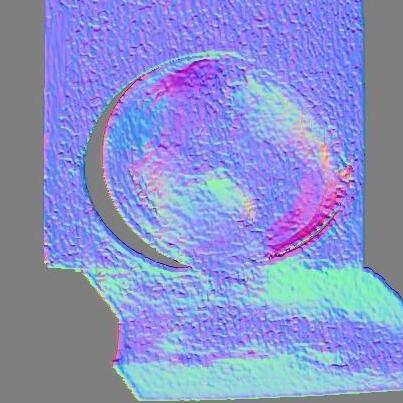}{1\linewidth}{28.47}{white}
		&\mae{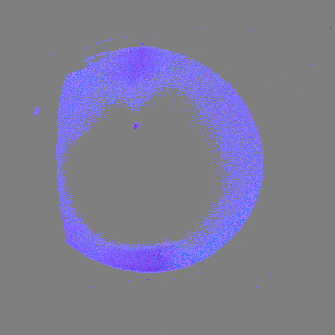}{1\linewidth}{42.55}{white}
		&\mae{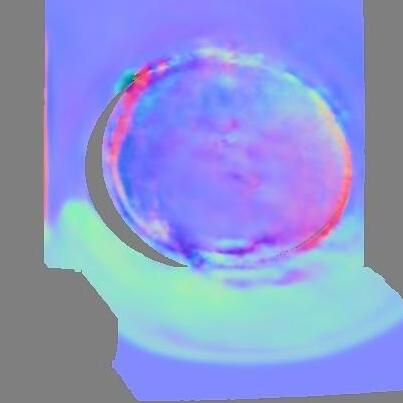}{1\linewidth}{33.39}{white}
		&\mae{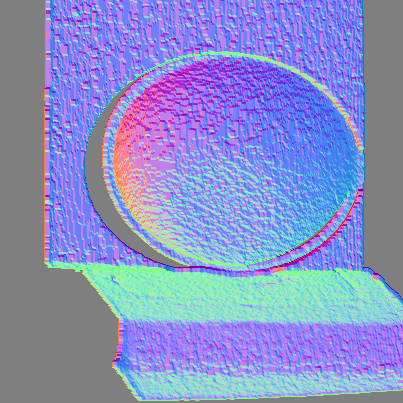}{1\linewidth}{}{none}
		\\
		\mae{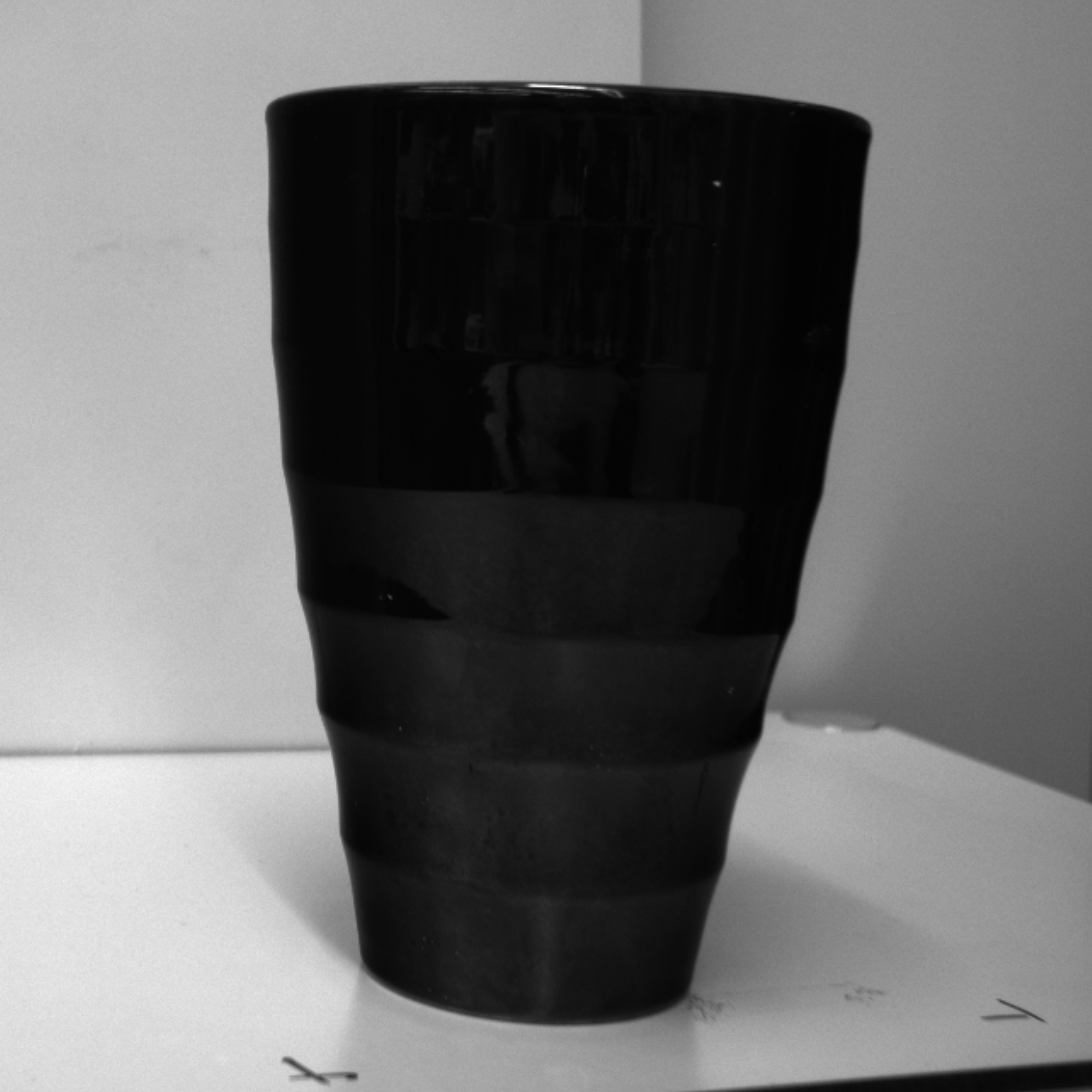}{1\linewidth}{}{none}
		&\mae{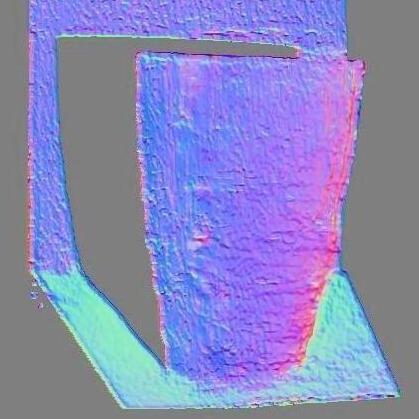}{1\linewidth}{19.97}{white}
		&\mae{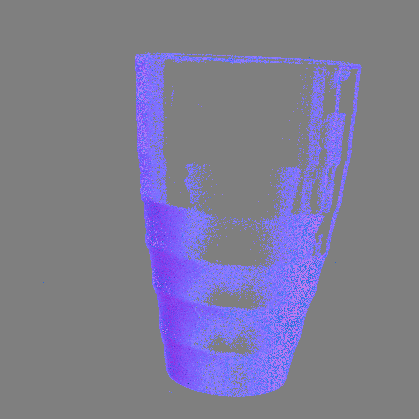}{1\linewidth}{27.65}{white}
		&\mae{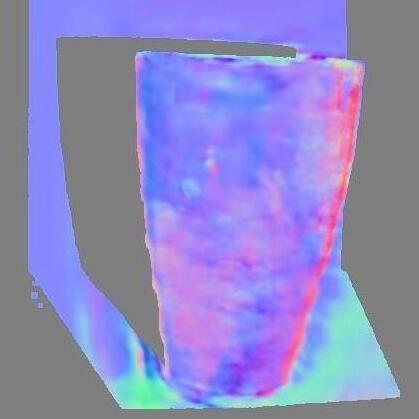}{1\linewidth}{26.60}{white}
		&\mae{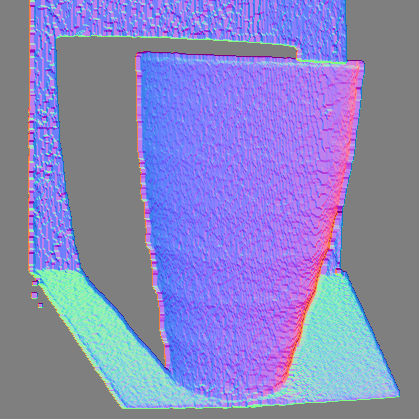}{1\linewidth}{}{none}
		\\
		\mae{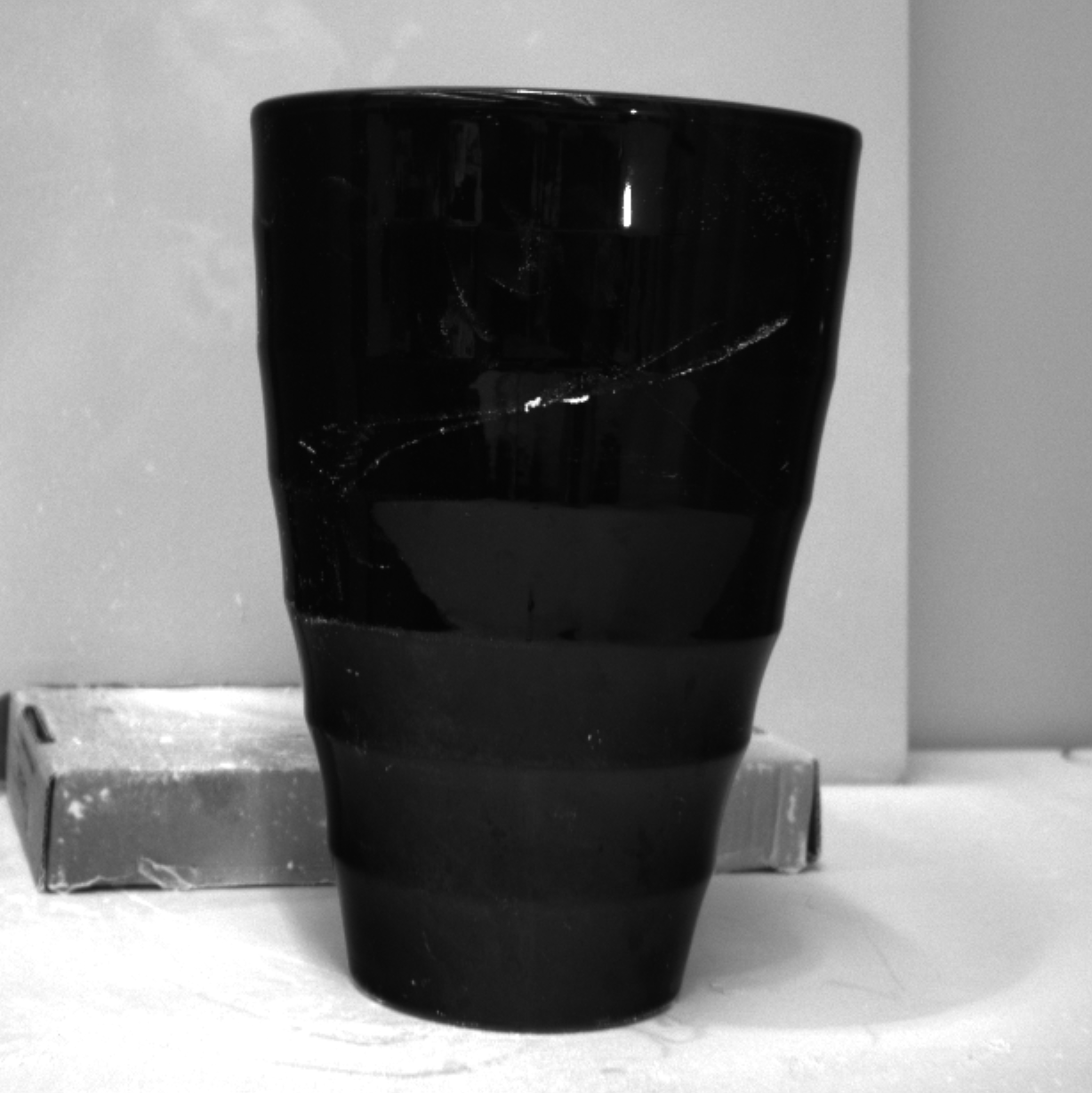}{1\linewidth}{}{none}
		&\mae{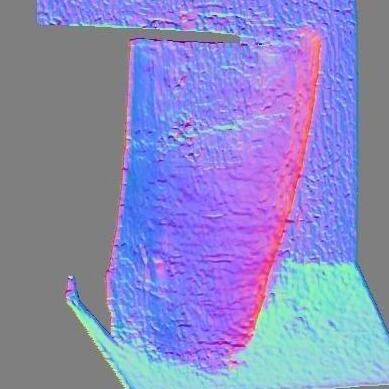}{1\linewidth}{24.54}{white}
		&\mae{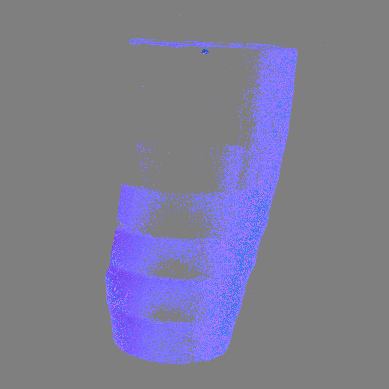}{1\linewidth}{33.14}{white}
		&\mae{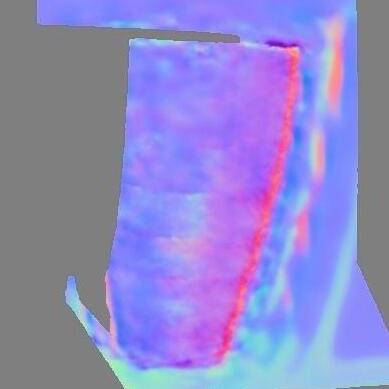}{1\linewidth}{25.71}{white}
		&\mae{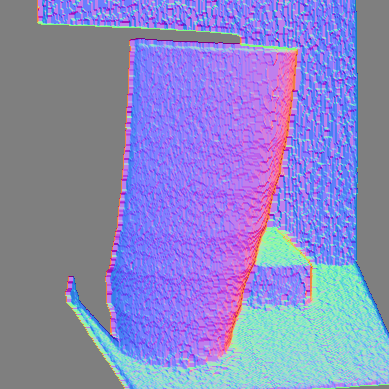}{1\linewidth}{}{none}
		\\
		\mae{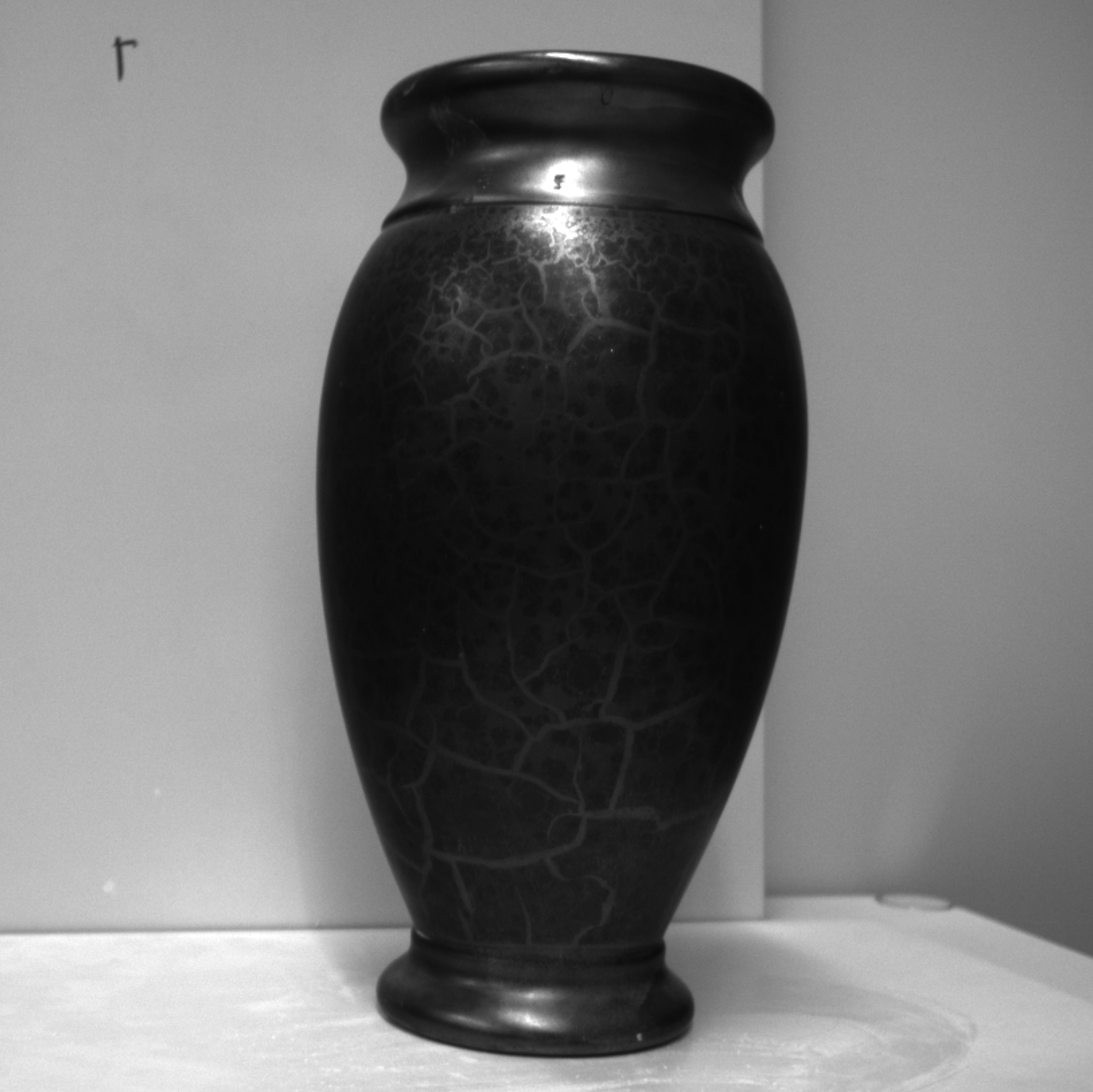}{1\linewidth}{}{none}
		&\mae{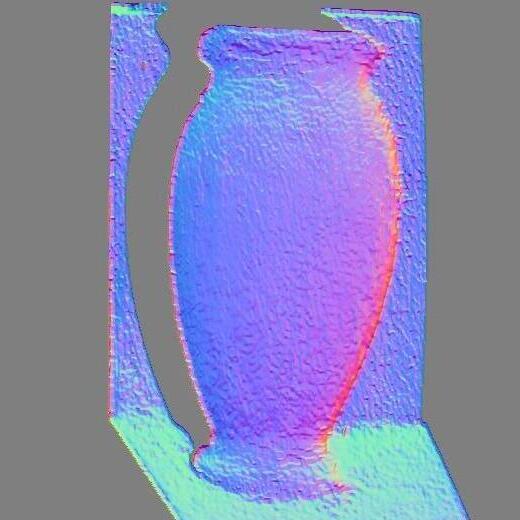}{1\linewidth}{18.06}{white}
		&\mae{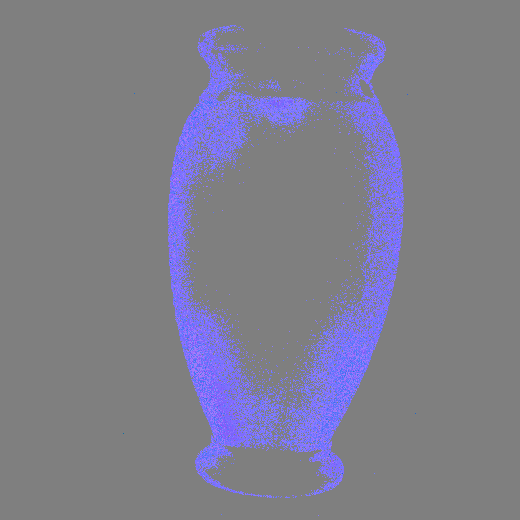}{1\linewidth}{35.38}{white}
		&\mae{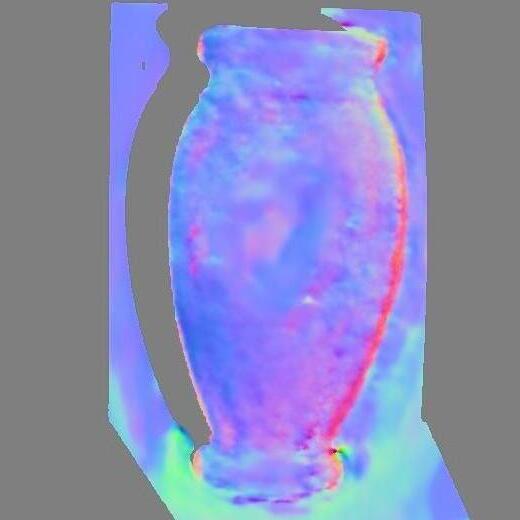}{1\linewidth}{25.48}{white}
		&\mae{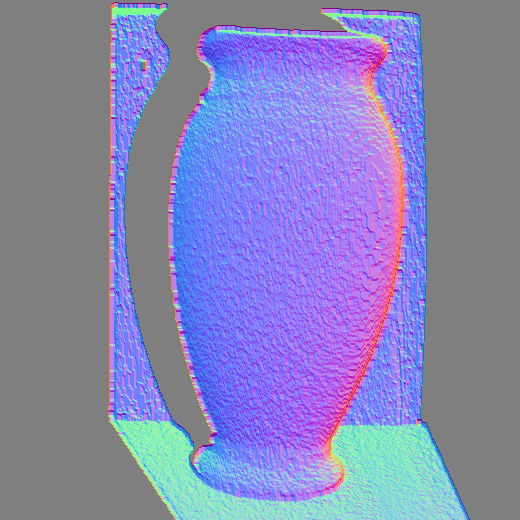}{1\linewidth}{}{none}
		\\
	\end{tabularx}
    \vspace{-1ex}
	\caption{Comparison of best image-baseline (Ba \etal) against ours event-based methods on the ESfP-Real dataset}
	\label{fig:qual_learn_real}
	\vspace{-4ex}
\end{figure}

%% file: floats/fig_illum_example.tex
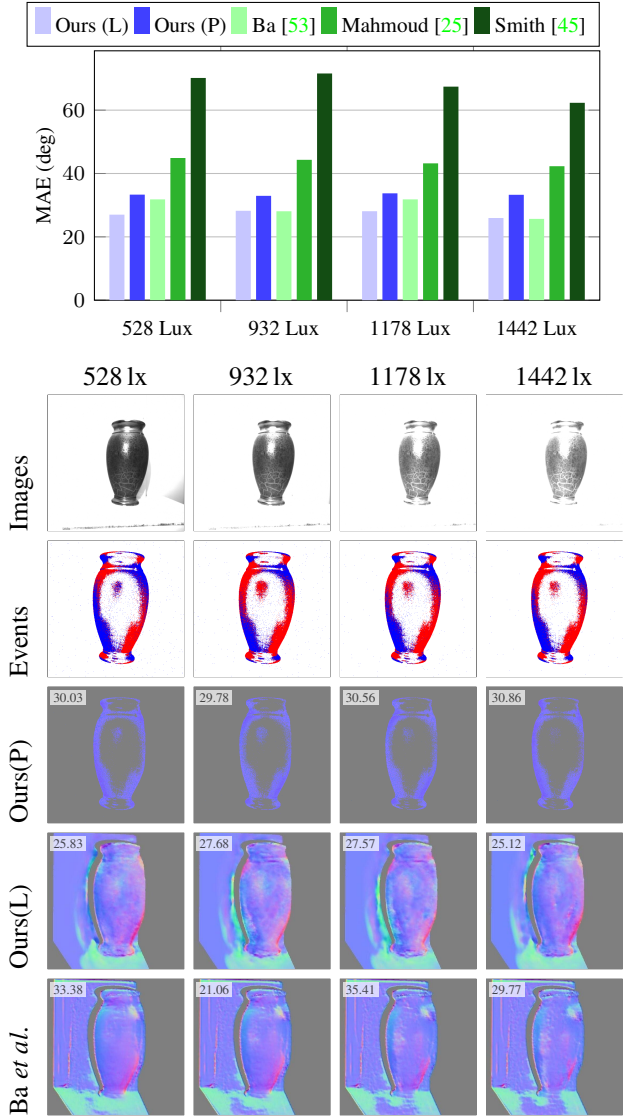
\begin{figure}[!t]
    \centering
    \colorlet{crystal}{blue!75}
\colorlet{lightgreen}{green!75}
\pgfplotstableread{
2	27.0070   31.8297   33.3289   44.8930   70.1609
3	28.2251   28.0758   32.9482   44.3170   71.5724
4	28.0912   31.8297   33.7380   43.2027   67.4197
5	25.9474   25.6721   33.2767   42.2995   62.3154
}\dataset
\tikzstyle{every node}=[font=\footnotesize]
\begin{tikzpicture}
\begin{axis}[ybar,
        width=8.3cm,
        height = 4.9cm,
        bar width = 2mm,
        ymin=0,
        ylabel shift = -4pt,
        ylabel={ MAE (deg)},
        xtick=data,
        xticklabels = {
            528 Lux,
            932 Lux,
            1178 Lux,
            1442 Lux,
        },
        ymajorgrids,
        major x tick style = {opacity=0},
        minor x tick num = 1,
        minor tick length=1ex,
         enlarge x limits={abs=0.5},
         legend columns = 5,
         legend style = {at={(axis cs:5.5,80)}, anchor=south east}, 
         legend image code/.code={
        \draw [#1,draw=none] (0cm,-0.1cm) rectangle (0.2cm,0.25cm); },
        ]
\addplot[draw=none,fill=crystal!30] table[x index=0,y index=1] \dataset; %
\addplot[draw=none,fill=crystal] table[x index=0,y index=3] \dataset; %
\addplot[draw=none,fill=lightgreen!50] table[x index=0,y index=2] \dataset; %
\addplot[draw=none,fill=lightgreen!70!black] table[x index=0,y index=4] \dataset; %
\addplot[draw=none,fill=lightgreen!30!black] table[x index=0,y index=5] \dataset; %
\legend{Ours (L), Ours (P), Ba \cite{Ba20ECCV},  Mahmoud \cite{Mahmoud12ICIP}, Smith \cite{Smith19PAMI}}
\end{axis}
 \end{tikzpicture}
 
 \vspace*{6pt}
    
    \setlength{\tabcolsep}{2pt}
	\begin{tabularx}{1\linewidth}{lCCCCCC}
		 & \unit[528]{lx}  & \unit[932]{lx}& \unit[1178]{lx}  & \unit[1442]{lx}
		\\
        \rotatebox{90}{\makecell{Images}}
		&\mae{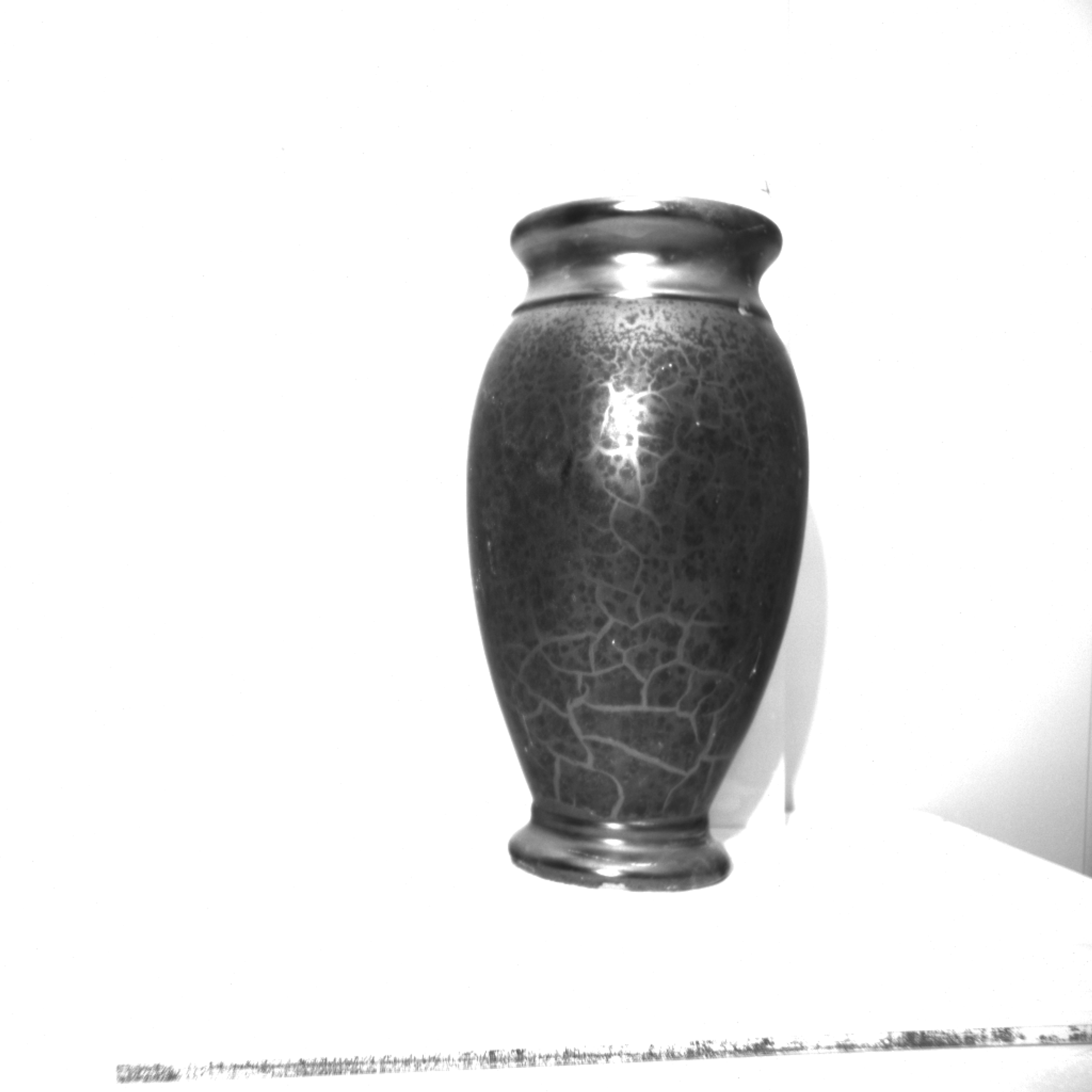}{1\linewidth}{}{none}
		&\mae{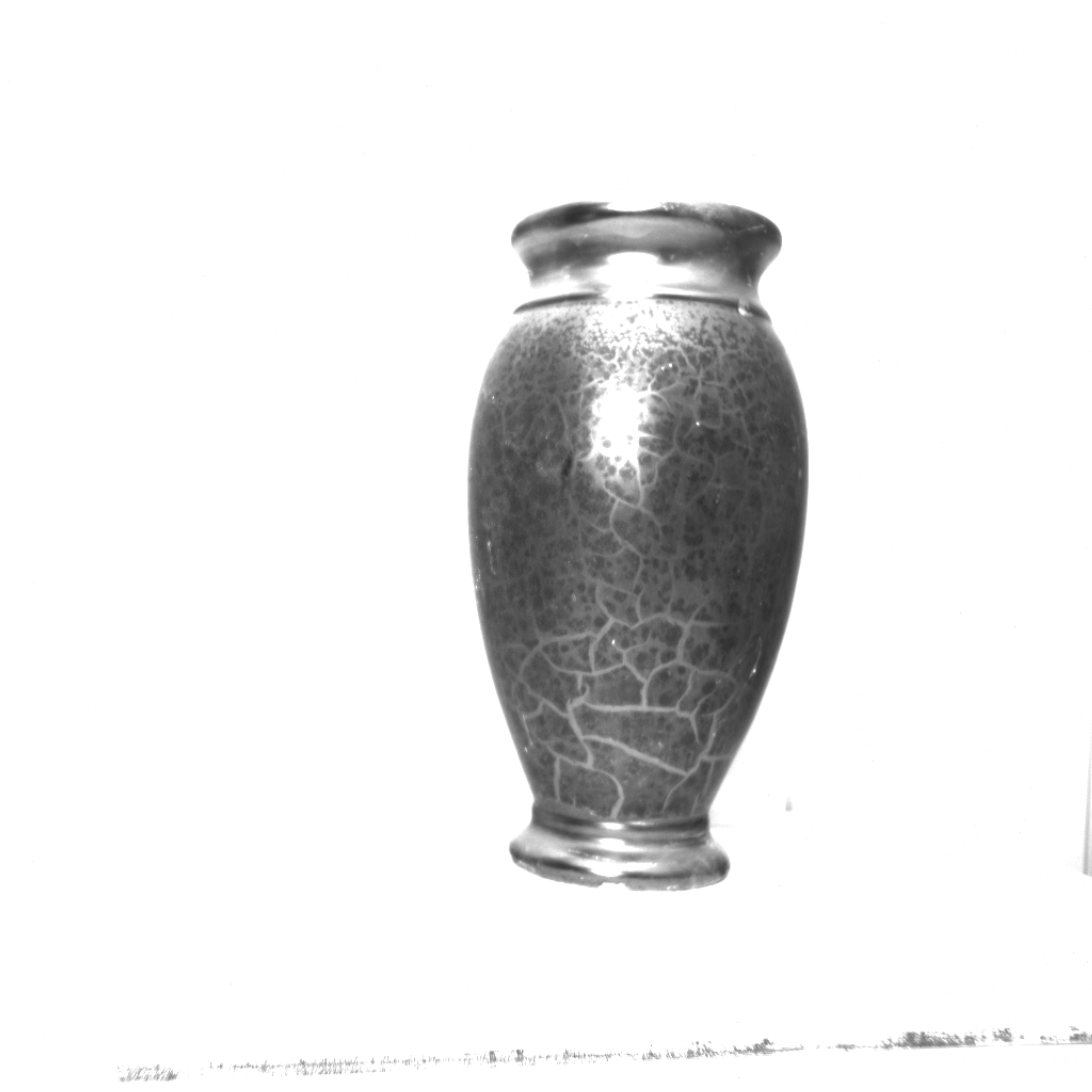}{1\linewidth}{}{none}
		&\mae{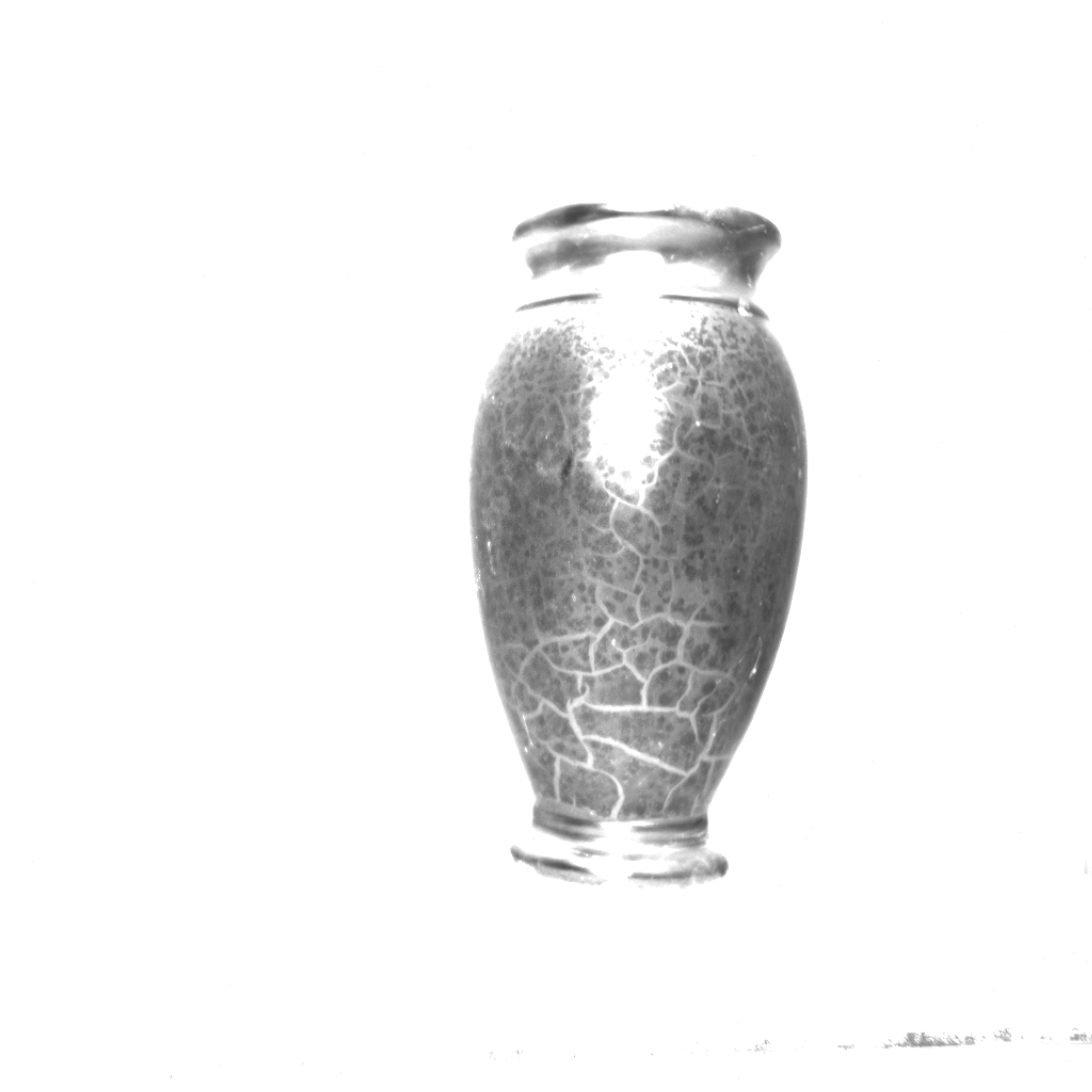}{1\linewidth}{}{none}
		&\mae{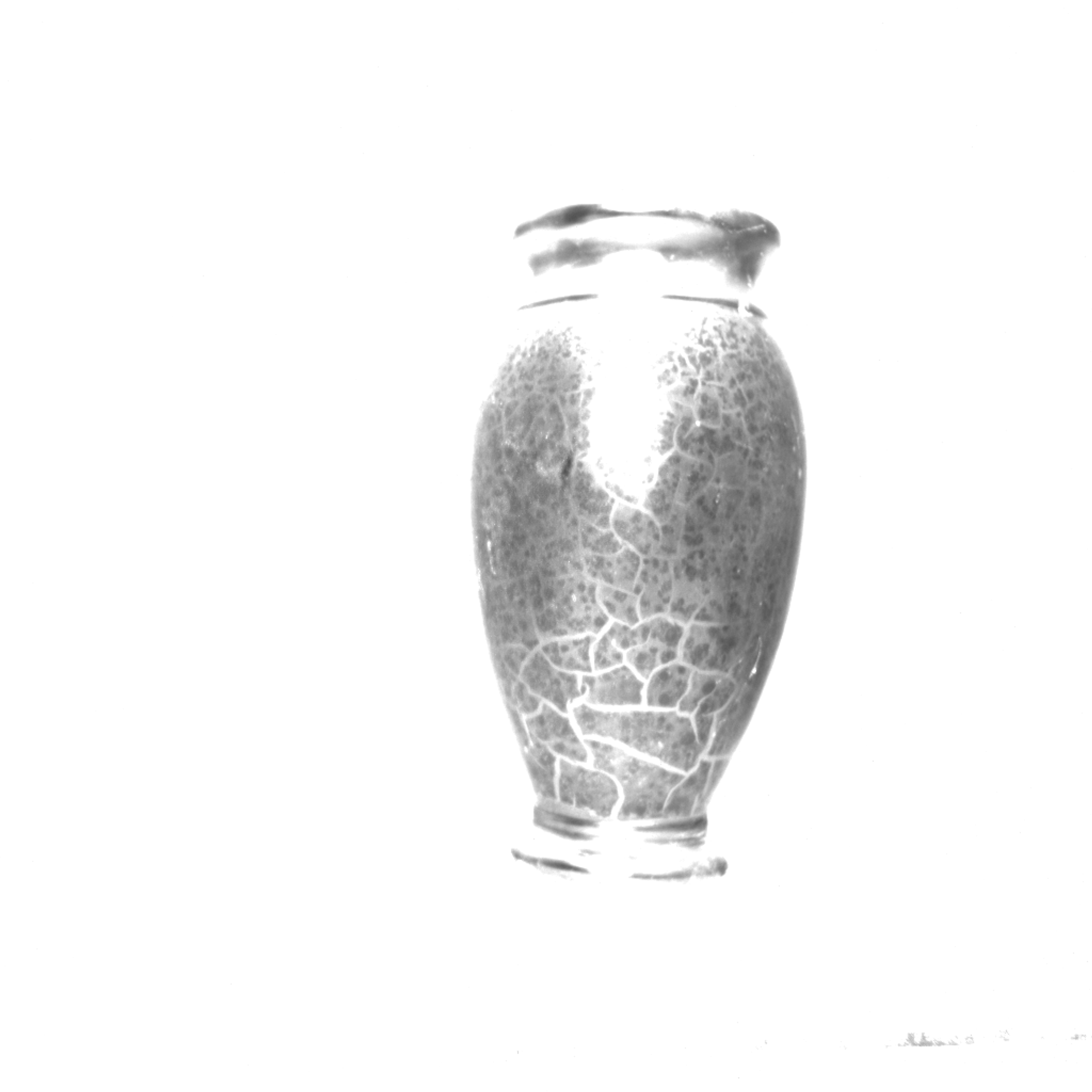}{1\linewidth}{}{none}
		\\
        \rotatebox{90}{\makecell{Events}}
		&\mae{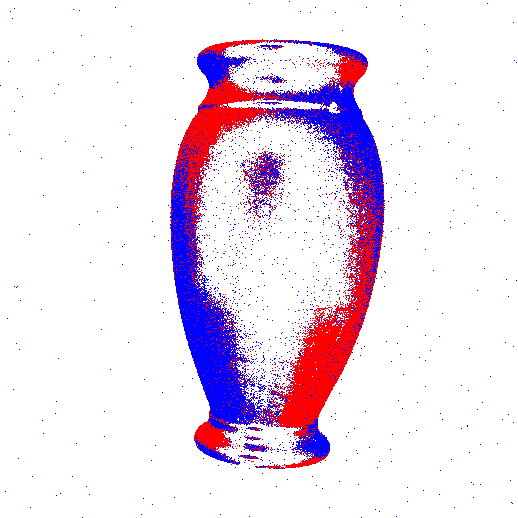}{1\linewidth}{}{none}
		&\mae{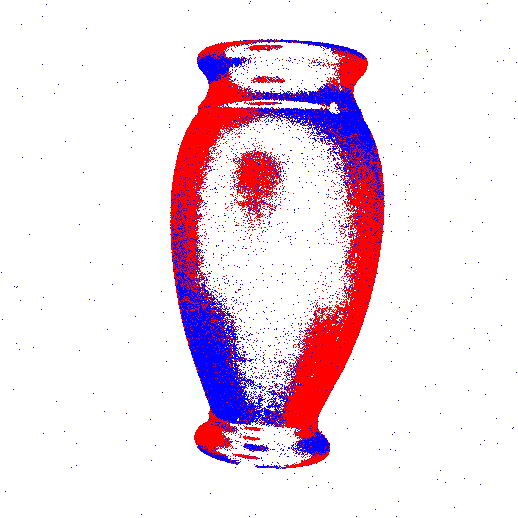}{1\linewidth}{}{none}
		&\mae{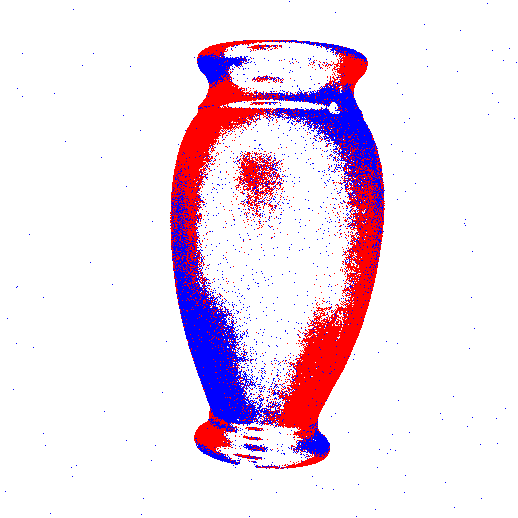}{1\linewidth}{}{none}
		&\mae{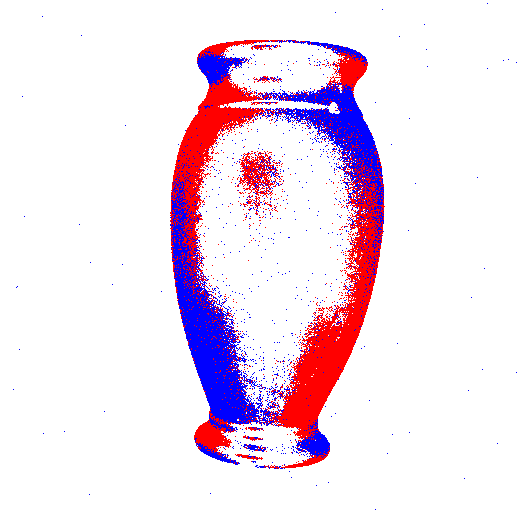}{1\linewidth}{}{none}
		\\
		\rotatebox{90}{\makecell{Ours(P)}}
		&\mae{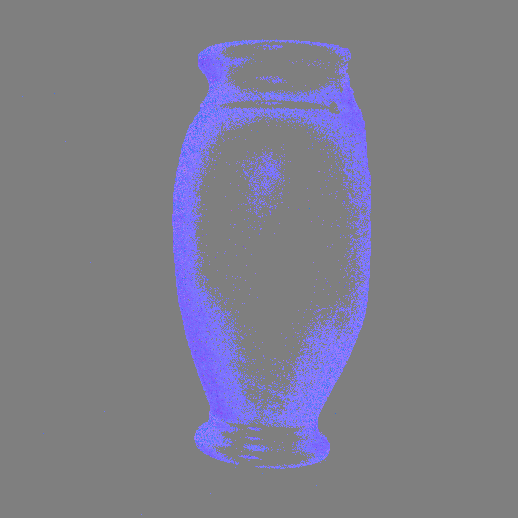}{1\linewidth}{30.03}{white}
		&\mae{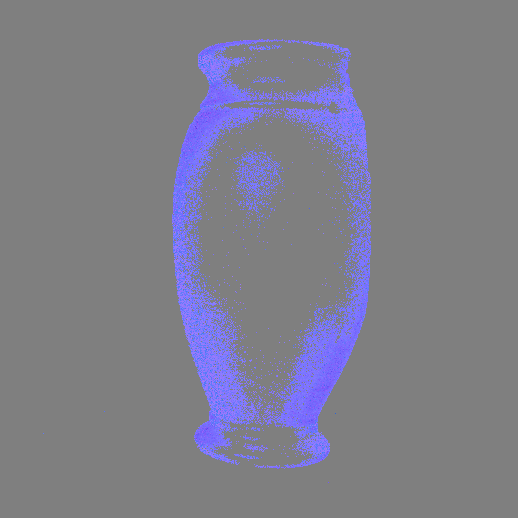}{1\linewidth}{29.78}{white}
		&\mae{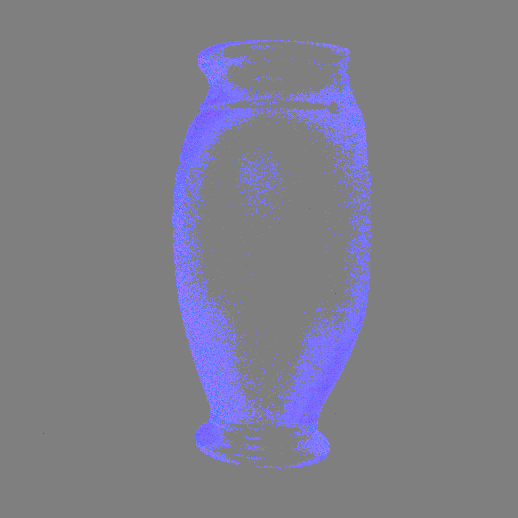}{1\linewidth}{30.56}{white}
		&\mae{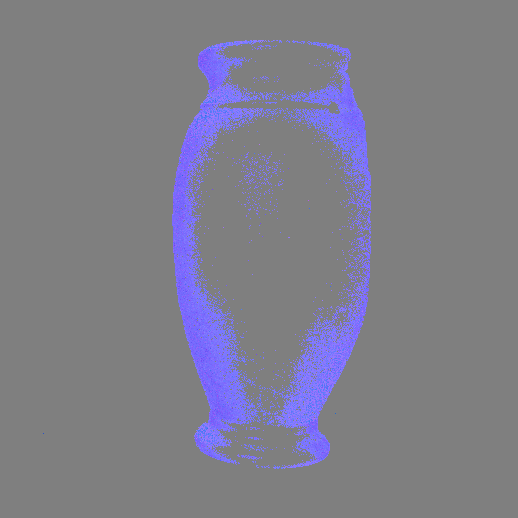}{1\linewidth}{30.86}{white}
		\\
		\rotatebox{90}{\makecell{Ours(L)}}
		&\mae{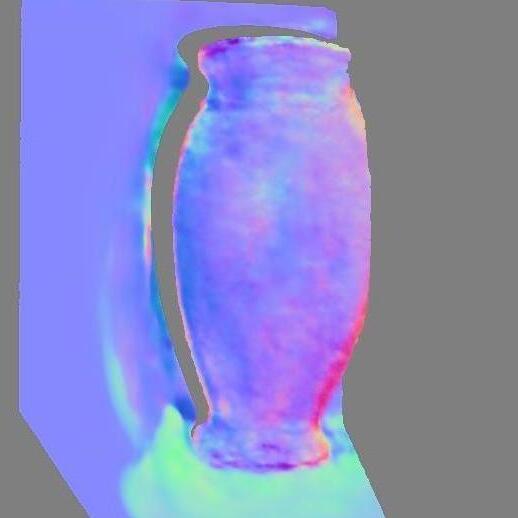}{1\linewidth}{25.83}{white}
		&\mae{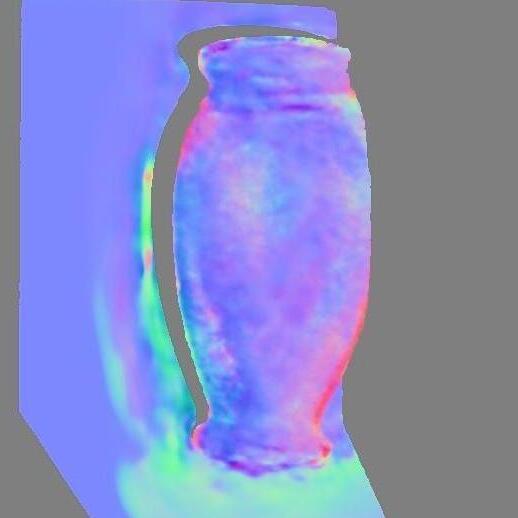}{1\linewidth}{27.68}{white}
		&\mae{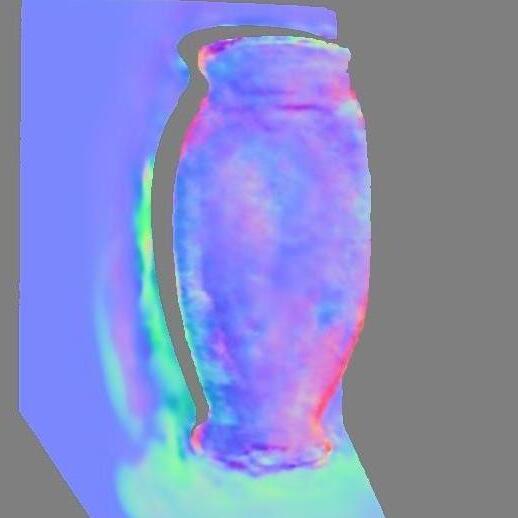}{1\linewidth}{27.57}{white}
		&\mae{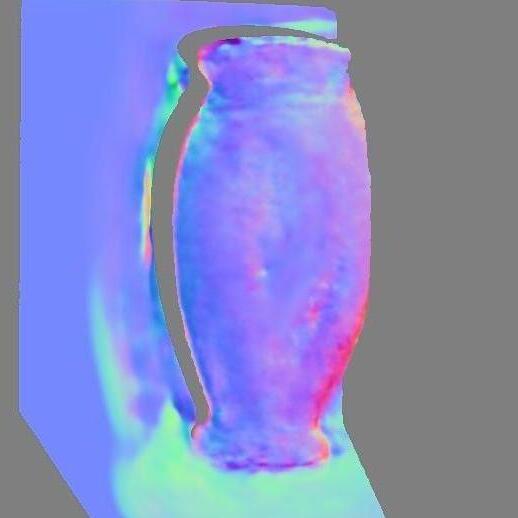}{1\linewidth}{25.12}{white}
		\\
		\rotatebox{90}{\makecell{Ba \etal}}
		&\mae{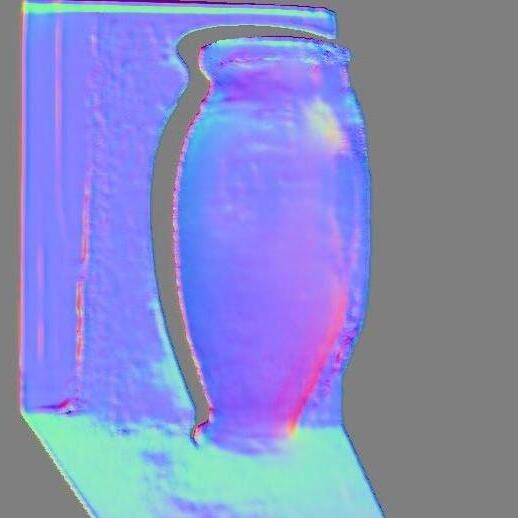}{1\linewidth}{33.38}{white}
		&\mae{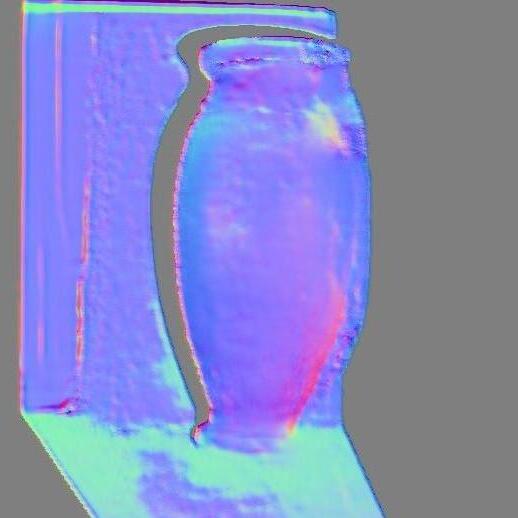}{1\linewidth}{21.06}{white}
		&\mae{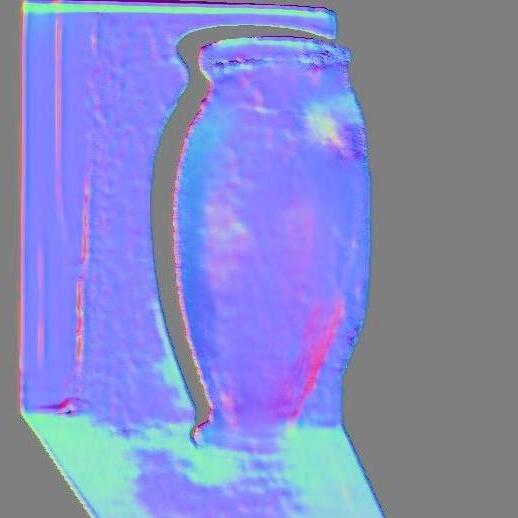}{1\linewidth}{35.41}{white}
		&\mae{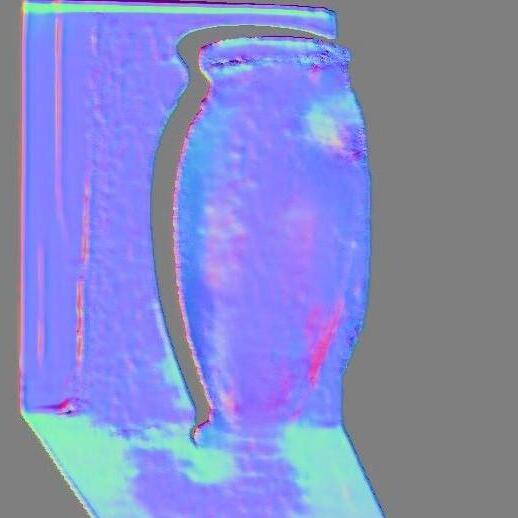}{1\linewidth}{29.77}{white}
		\\
	\end{tabularx}
    \vspace{-1ex}
	\caption{Varying the ambient illumination has a drastic effect on the image-based baselines (bar plot: green colors), whereas both our methods (bar plot: blue colors), perform consistently.
	}
	\label{fig:illum_real}
	\vspace{-2ex}
\end{figure}

%% file: floats/tab_speed_effect.tex
\begin{table}[t]
    \centering
    \begin{adjustbox}{max width=\linewidth}
    \setlength{\tabcolsep}{4pt}
    {\small
    \begin{tabular}{|l|c|c rrr|}
        \toprule
        Speed (RPM) & Task & Angular Error $\downarrow$  &  \multicolumn{3}{c|}{Accuracy $\uparrow$} \\
         & & Mean & \aeone & \aetwo & \aethree \\
        \midrule
        57.5   & Physics-based  & 44.29 & 0.01026 & 0.1294 & 0.1998 \\ 
        171.25 & Physics-based & 44.13 & 0.01028 & 0.1297 & 0.2000 \\ 
        308.75 & Physics-based & \underline{44.02} & \underline{0.01043} & \underline{0.1302} & \underline{0.2008} \\ \bottomrule
        57.5  & Learning-based & 29.47 & 0.147 & 0.441 & 0.613 \\ 
        171.25 & Learning-based & \textbf{27.10} &\textbf{ 0.177 }& \textbf{0.498} &\textbf{ 0.667} \\ 
        308.75 & Learning-based & 27.47 & 0.172 & 0.491 & 0.656 \\ \bottomrule
    \end{tabular}}
    \end{adjustbox}
    \caption{Ablation study on the effect of using different angular velocities for the rotation of the polarizing lens on the proposed ESfP-Real dataset.}
    \label{tab:realdataset_speed}
\end{table}

%% file: floats/fig_outdoor.tex
\begin{figure}[t]
	\centering
    \setlength{\tabcolsep}{2pt}
	\begin{tabularx}{1\linewidth}{CCCC}
		Scene & Events &  Ours L
		\\
		\mae{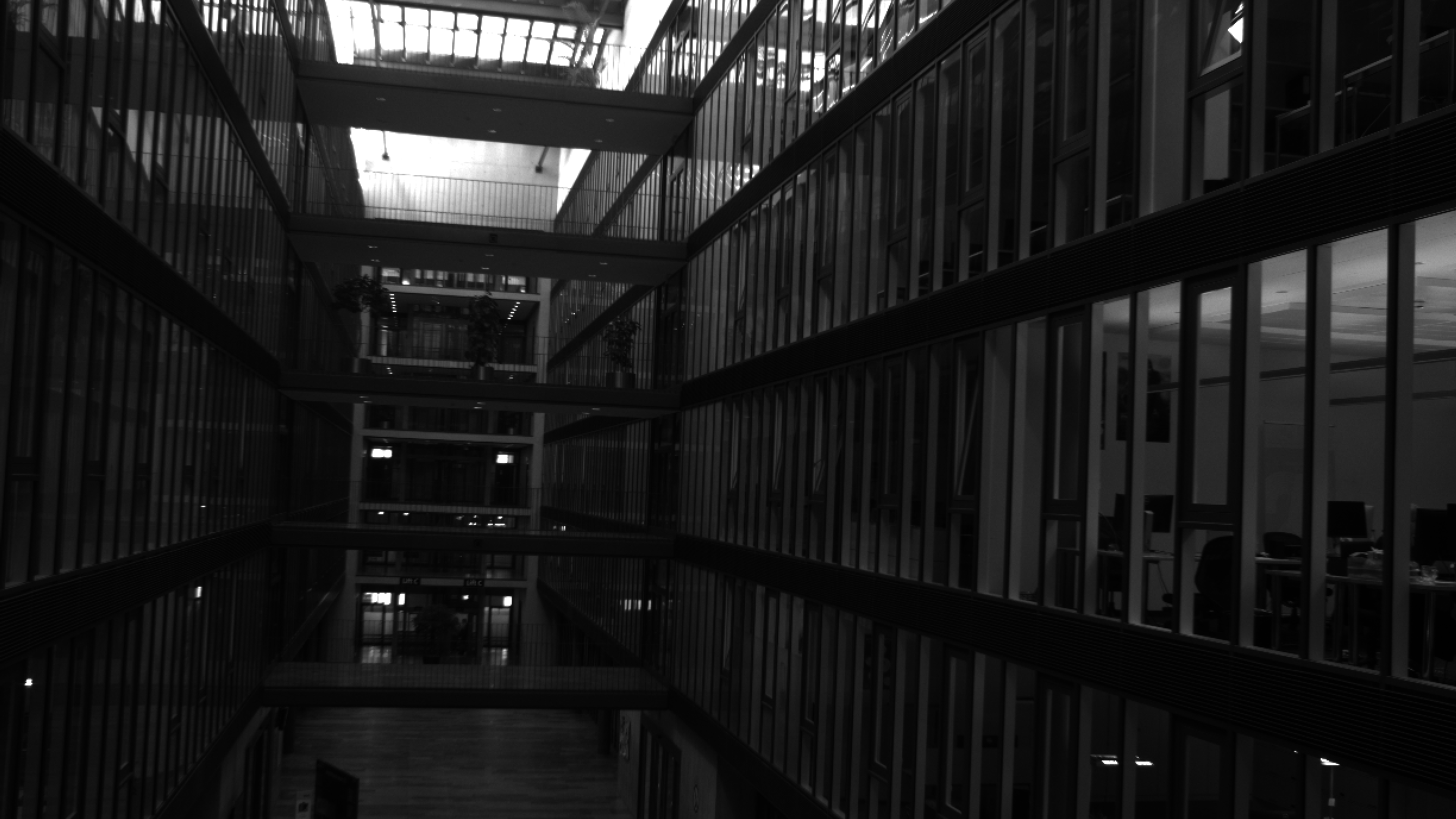}{1\linewidth}{}{none}
		&\mae{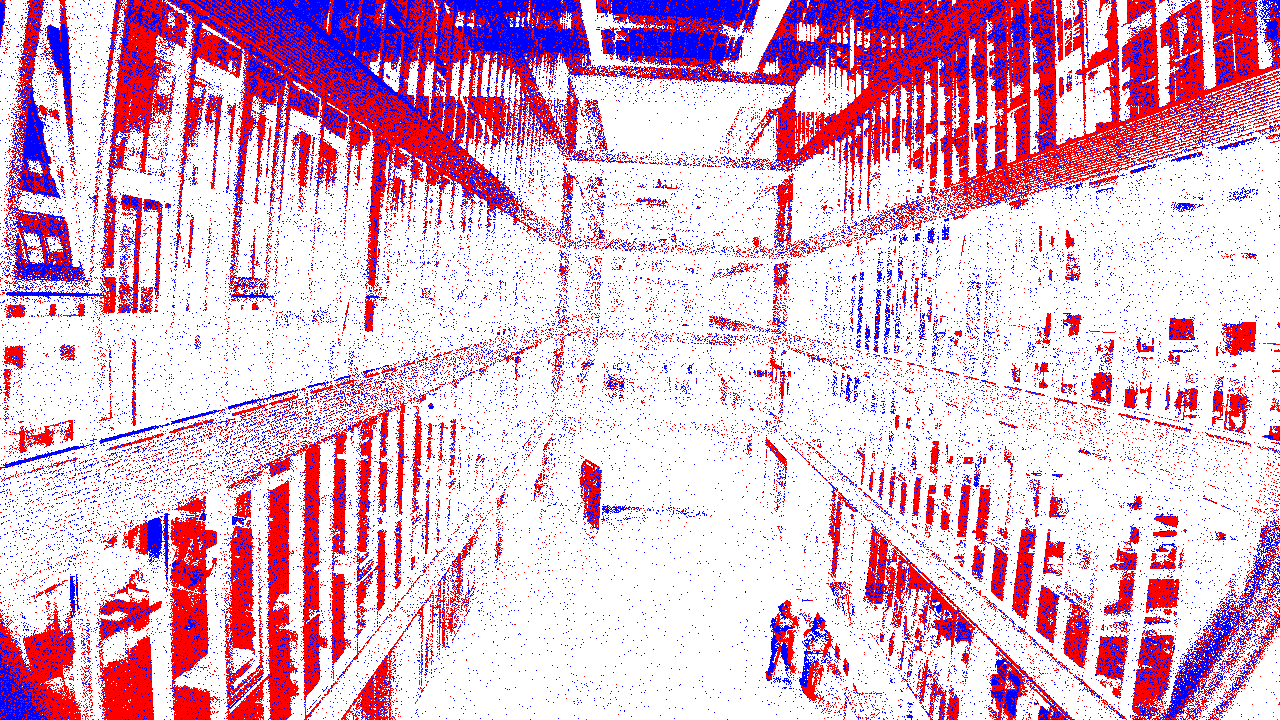}{1\linewidth}{}{none}
		&\mae{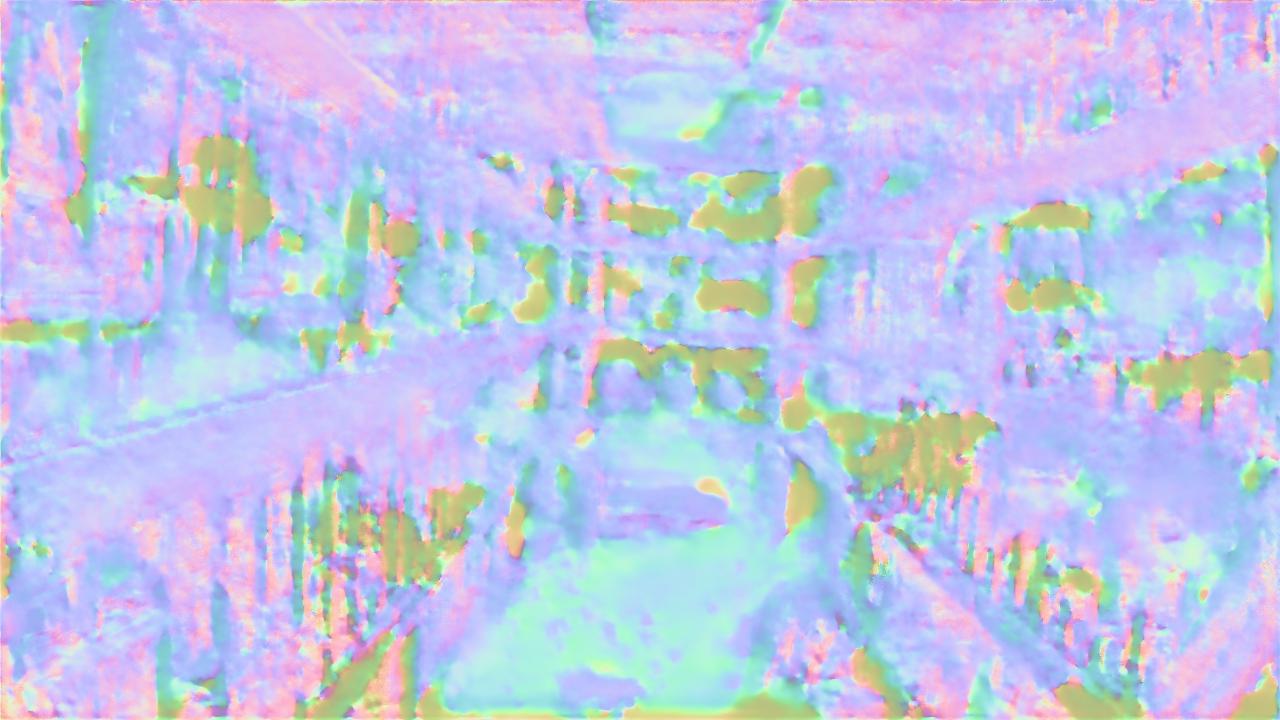}{1\linewidth}{}{none}
		\\
		\mae{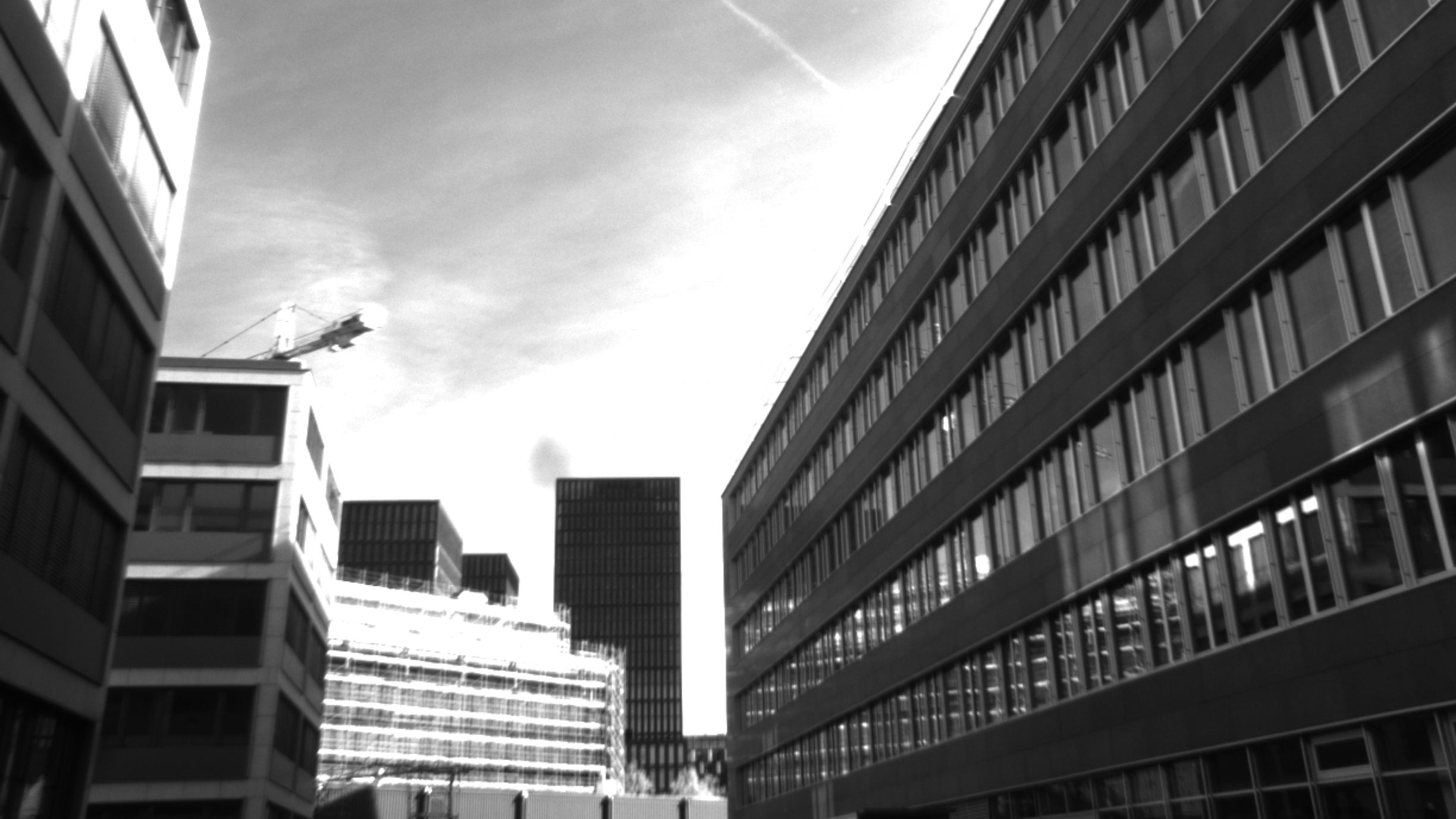}{1\linewidth}{}{none}
		&\mae{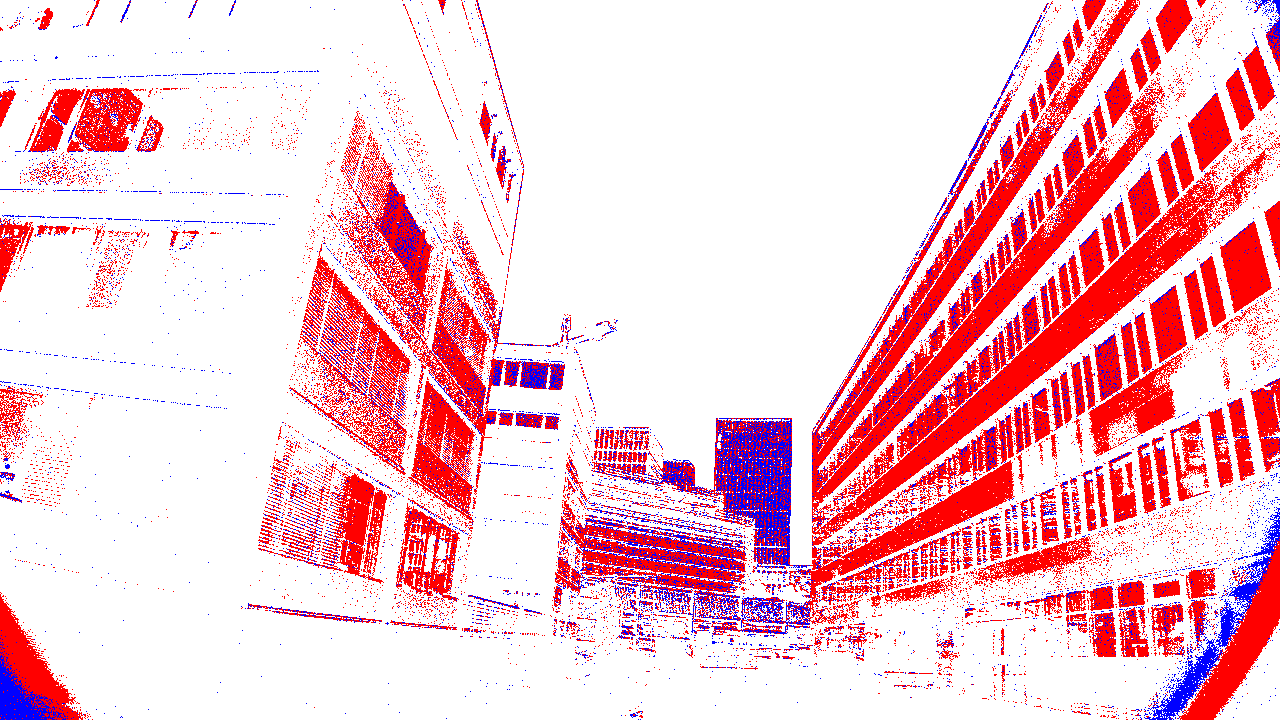}{1\linewidth}{}{none}
		&\mae{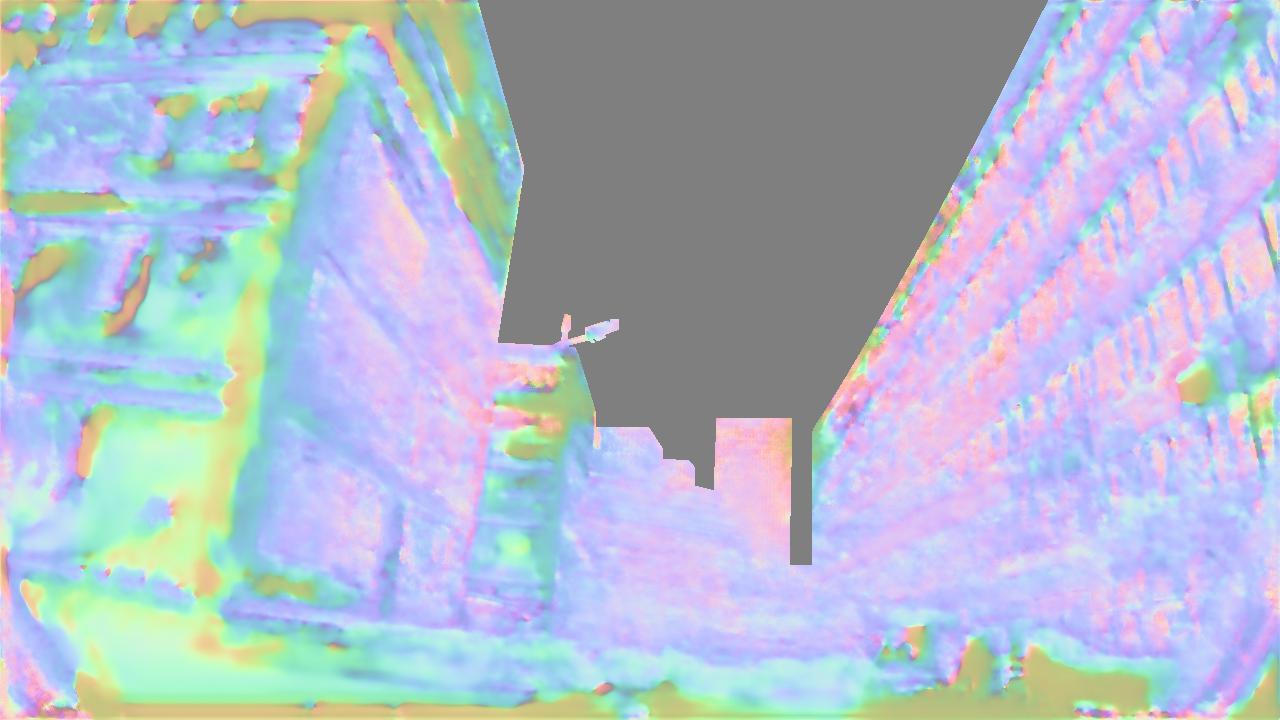}{1\linewidth}{}{none}
		\\
		\mae{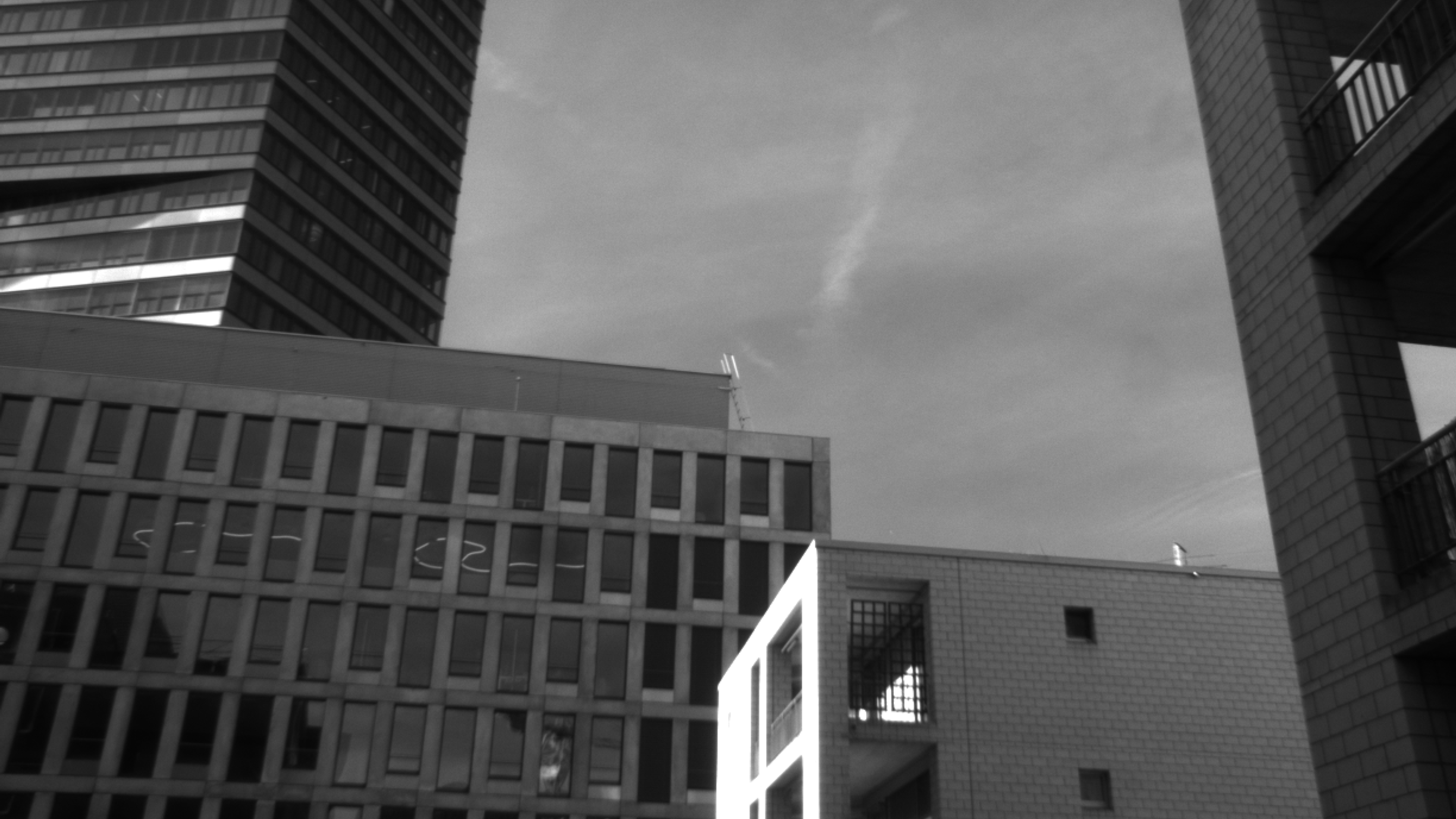}{1\linewidth}{}{none}
		&\mae{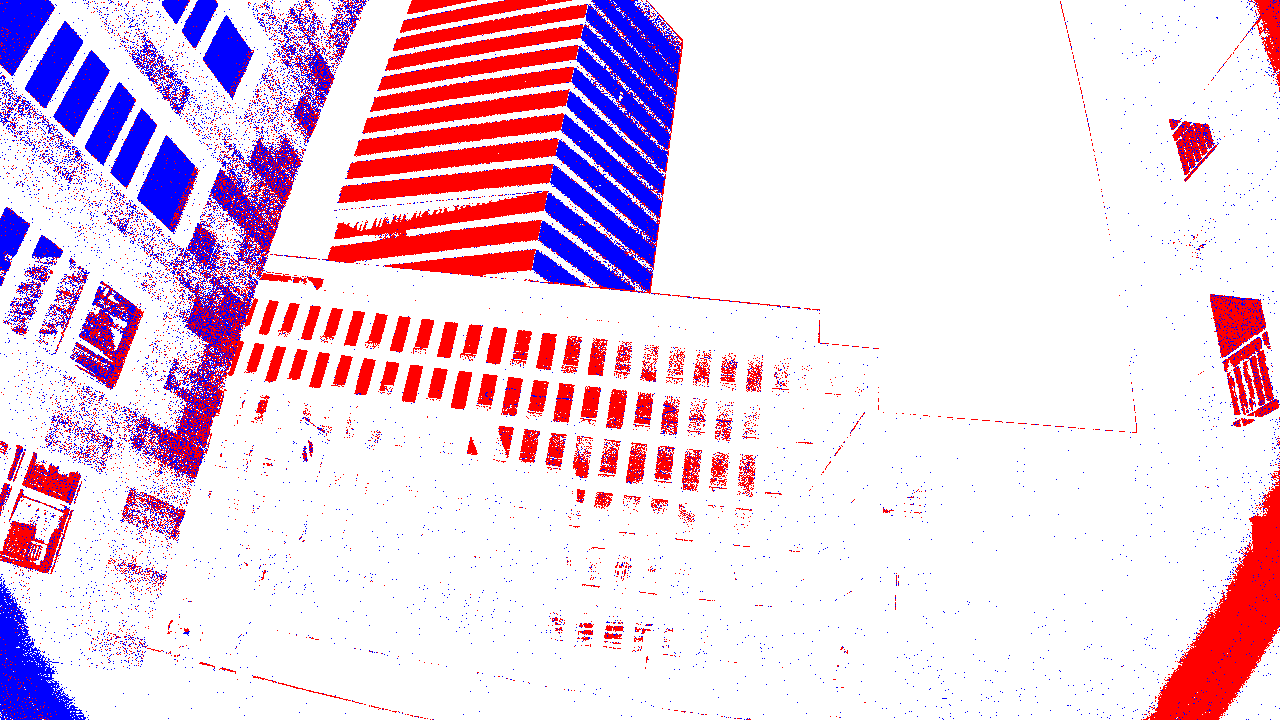}{1\linewidth}{}{none}
		&\mae{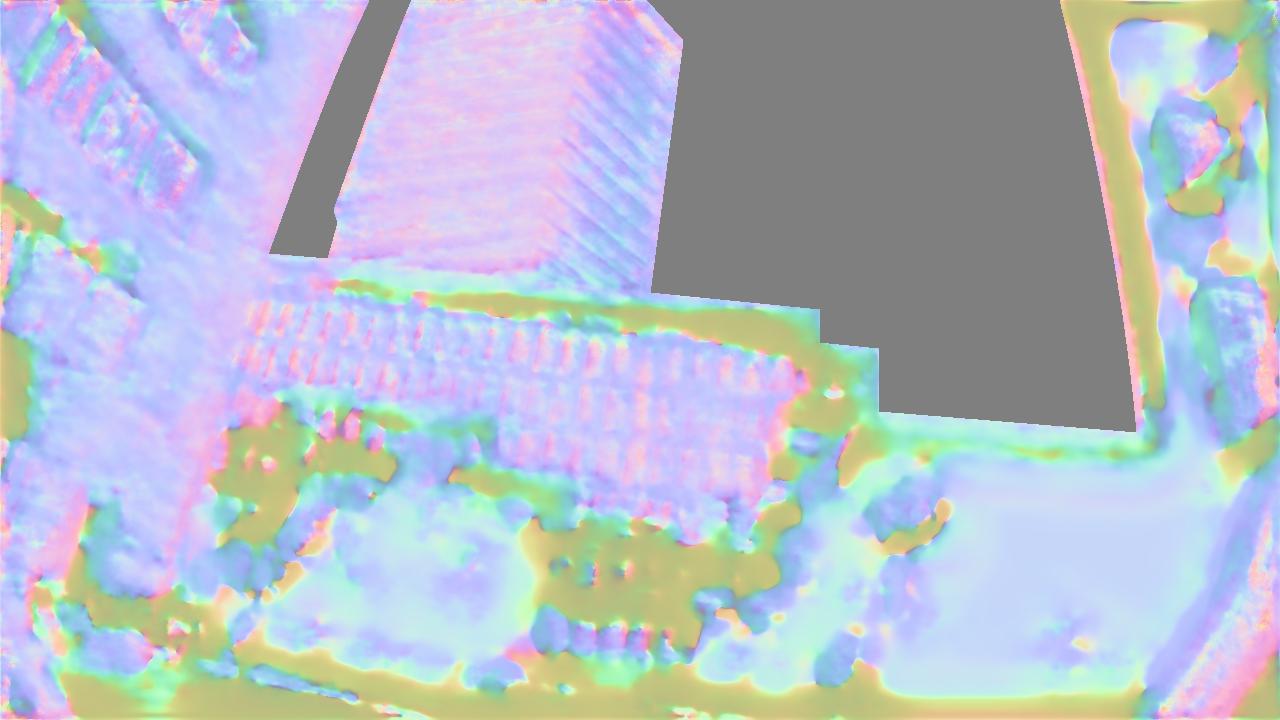}{1\linewidth}{}{none}
		\\
		\mae{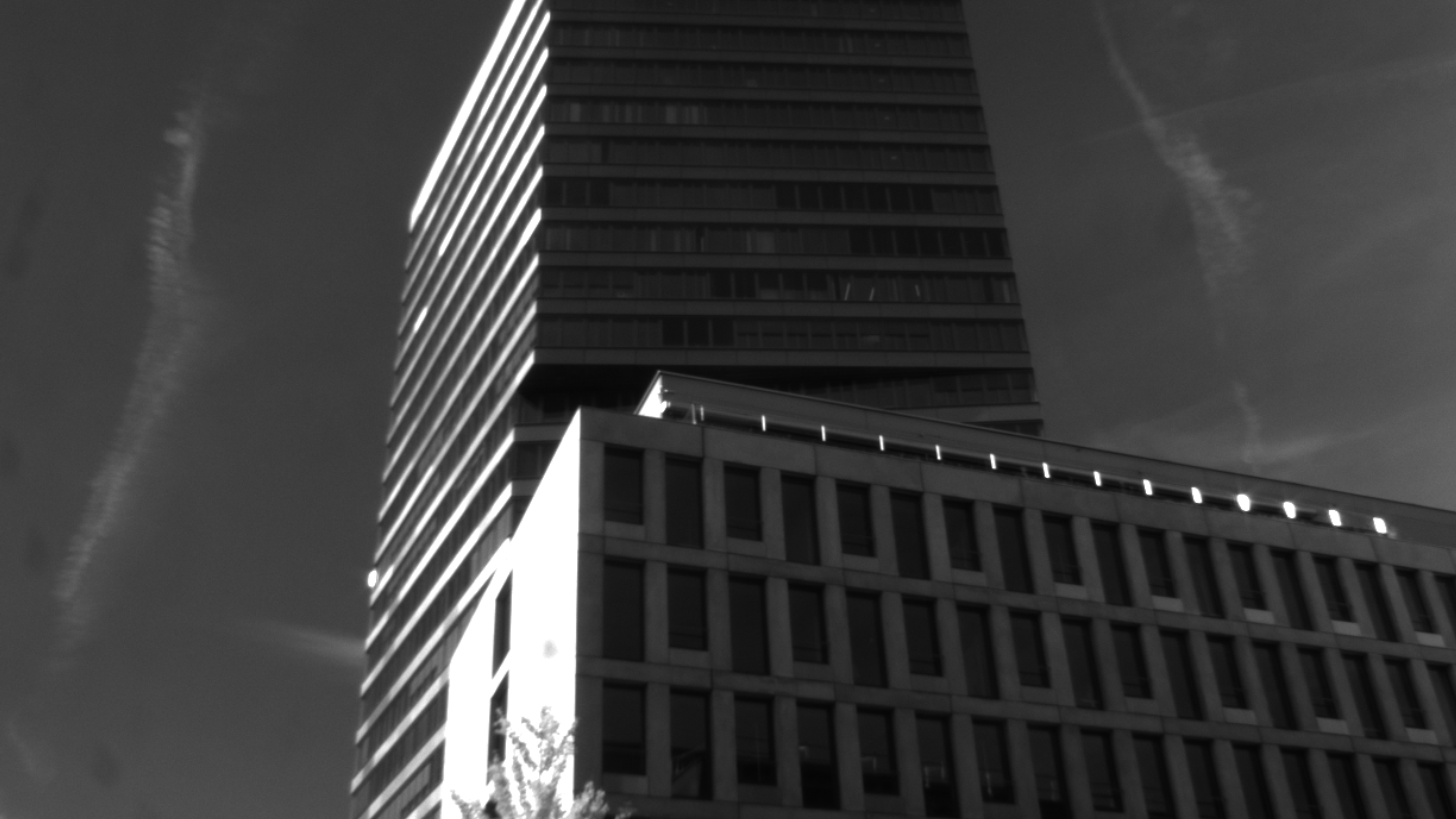}{1\linewidth}{}{none}
		&\mae{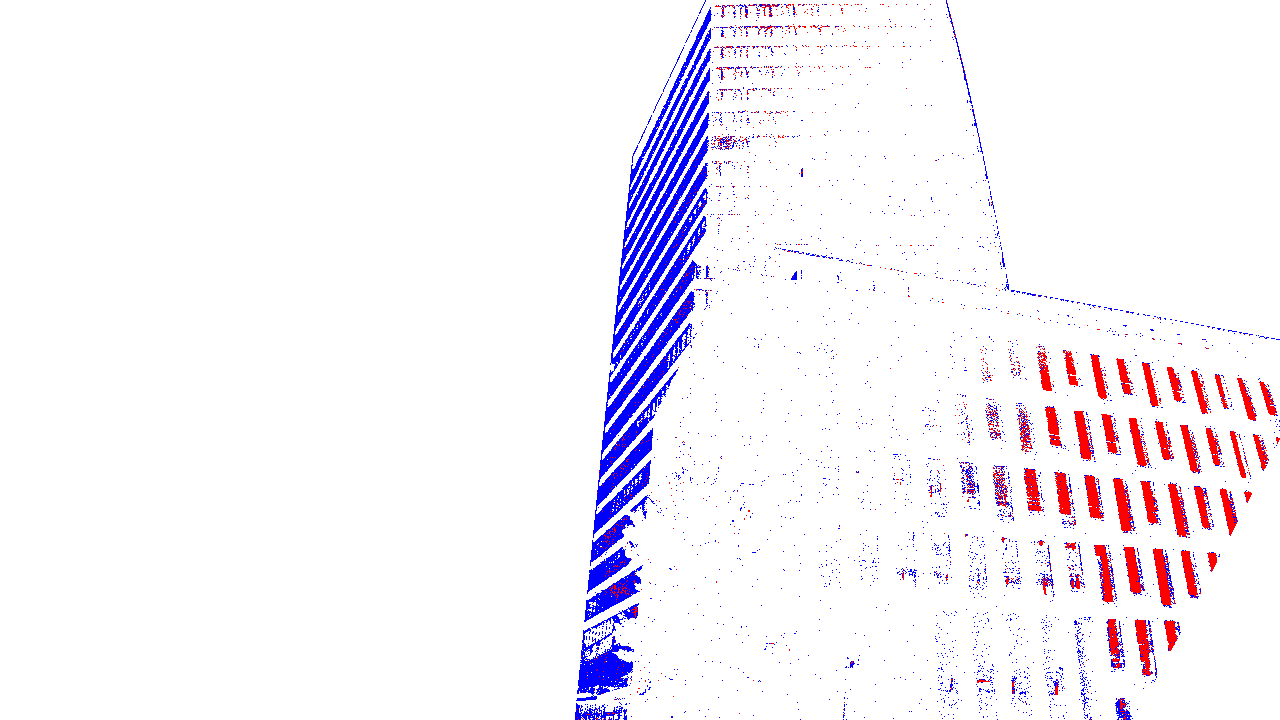}{1\linewidth}{}{none}
		&\mae{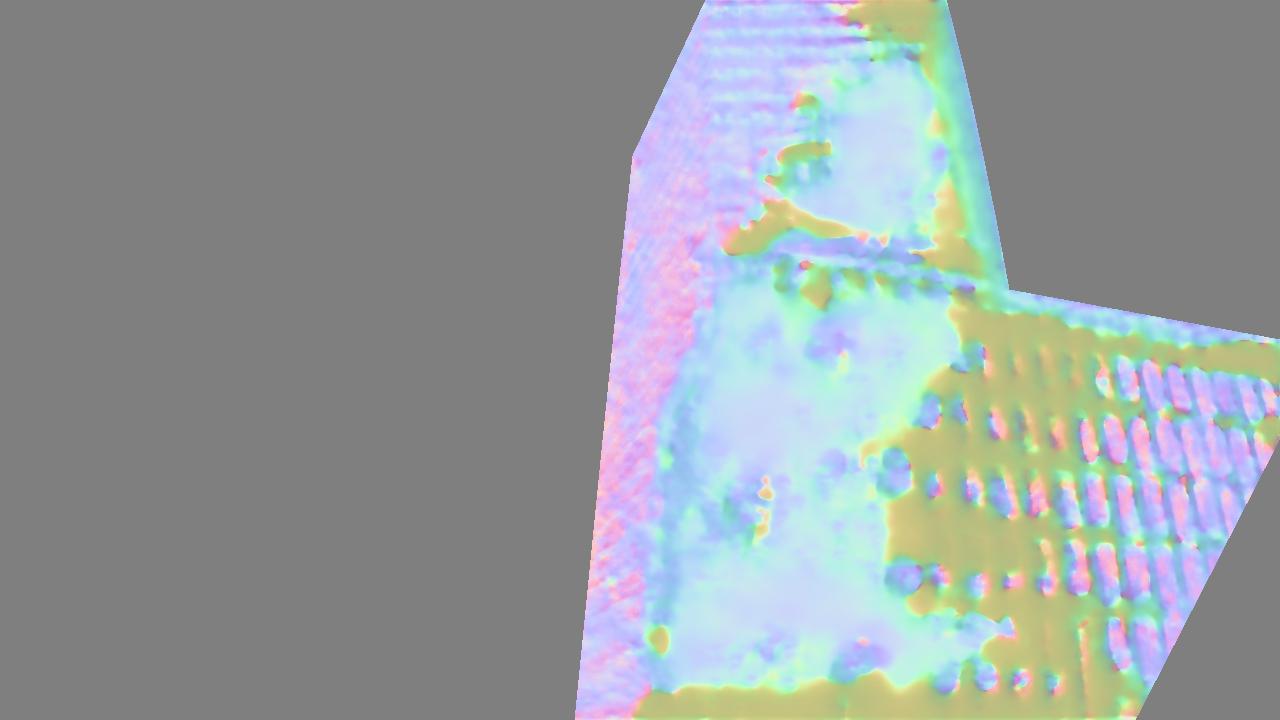}{1\linewidth}{}{none}
		\\

	\end{tabularx}
    \vspace{-1ex}
	\caption{Performance of our method on outdoor scenes}
	\label{fig:qual_outdoor}
	\vspace{-3ex}
\end{figure}

%% file: sections/05_conclusion.tex
\vspace*{-3pt}
\section{Conclusions}
\vspace*{-2pt}
We introduce a novel method for surface normal estimation using a rotating linear polarizer and an event camera.
To the best of our knowledge, we are the first to propose a principled way of estimating the surface normals from polarization events using a physics-based solution.
Our method takes advantage of the rich temporal event information to reconstruct relative event intensities at multiple polarizer angles.
We also propose a learning-based method to overcome the non-idealities of the physical sensor which improves the performance by $52 \%$ in terms of MAE.
Despite reduced performance in the presence of event camera non-idealities, we believe our derivations could pave the way to future hybrid solutions to increased robustness and interpretability.

%% file: sections/06_acknoledgement.tex
\section{Acknowledgement}
This research was supported by SONY R\&D Center Europe and the National Centre of Competence in Research (NCCR) Robotics (grant agreement No. 51NF40-185543) through the Swiss National Science Foundation (SNSF).

%% file: sections/supplementary_material.tex
\section{Surface Normal from Events}
In this section, we describe the details for surface normal estimation from polarizer images.
We then extend this knowledge to estimate surface normals from events.
\subsection{Basics of Shape-from-Polarization (SfP)}
Intensity change at $\phi_{pol}$ can be expressed as:
\begin{equation}
    I(\phi_\text{pol}) = \frac{I_\text{max} + I_\text{min}}{2} + \frac{I_\text{max} - I_\text{min}}{2}  \cdot \cos (2(\phi_\text{pol} - \phi)),
\end{equation}
where $I_{min}$ and $I_{max}$ represent the minimum and maximum magnitude seen through the polarizer respectively \cite{Kadambi15ICCV, Morel05SPIE}.
This equation can be expressed in terms of the magnitude of the light $I_{un}$ and the proportion of polarized component $\rho$ (also known as degree of polarizer) as follows:
\begin{equation}
    I = I_{max}+I_{min}
\end{equation}
\begin{equation}
    \rho = \frac{I_\text{max} - I_\text{min}}{I_\text{max} + I_\text{min}}
\end{equation}
Lastly, $\phi$ is the angle of the linearly polarized component which corresponds to the phase shift of the sinosoid. \cite{Morel05SPIE}.
Estimating these three parameters forms the crux of shape-from-polarization techniques \cite{Wolff97IVC}.
These quantities can be estimated from images captured at $4$ different polarization angles as follows:
\begin{equation}
    I_{un} = \frac{I[0] + I[\sfrac{\pi}{4}] + I[\sfrac{\pi}{2}] + I[\sfrac{3\pi}{4}]}{2} \\
\end{equation}
 
 \begin{equation}
    \rho = \frac{ \sqrt{(I[0] - I[\sfrac{\pi}{2}])^2 + (I[\sfrac{\pi}{4}] - I[\sfrac{3\pi}{4}])^2} } 
    {I_{un}} \\
\end{equation} 

\begin{equation}
    \phi = \frac{1}{2} \cdot \arctan \frac{I[\sfrac{\pi}{4}] - I[\sfrac{3\pi}{4}]}{(I[0] - I[\sfrac{\pi}{2}]}
\end{equation}
To estimate these quantities, minimum $3$observations of the intensity are required. 
However, increasing the observations, improves the accuracy of surface normals.
To use $12$ polarization angles the above quantities can be derived as follows:
\begin{equation}
    I_{un} = \sum_{i=0}^{i=\pi} I[i]
\end{equation}
 
 \begin{align}
    Q1 = (I[0] - I[\sfrac{\pi}{2}])\\
    Q2 = (I[\sfrac{\pi}{12}] - I[\sfrac{7\pi}{12}])\\
    Q3 = (I[\sfrac{\pi}{6}] - I[\sfrac{3\pi}{2}])\\
    U1 = (I[\sfrac{\pi}{4}] - I[\sfrac{3\pi}{4}])\\
    U2 = (I[\sfrac{\pi}{3}] - I[\sfrac{5\pi}{6}])\\
    U3 = (I[\sfrac{5\pi}{12}] - I[\sfrac{11\pi}{12}])\\
 \end{align}
\begin{equation}
    \rho = \frac{ \sqrt{Q1^2 + U1^2 + Q2^2+U2^2 + Q3^2 + U3^2} } 
    {3*I_{un}}
\end{equation}
\begin{align}
    \phi = 1.5 * ( \arctan(U1/Q1) \\
         + \arctan(U2/Q2)- \pi/6 \\
         + \arctan(U3/Q3)- \pi/3)
\end{align}

Estimating the surface normals from $\rho$ and $\phi$ is a matter of estimating the zenith angle $\theta$ and azimuth angle $\alpha$ as shown in the equations below:
\begin{equation}
    \rho^{diffuse} = \frac{(n - \frac{1}{n})^2 \sin ^2 \theta}{ 2 + 2 n^2 - (n +\frac{1}{n})^2 \sin^2 \theta + 4 \cos \theta \sqrt{n^2 - sin^2\theta} }
\end{equation}

\begin{equation}
    \rho^{spec} = \frac{2 n \tan \theta}{\tan^2 \theta \sin^2 \theta + n^{*2}}
\end{equation}
where $n$ denotes the refractive index and $\theta$ is the zenith angle.
Depending on the type of reflection (diffuse or specular), the $\rho$ is computed differently.
Similarly depending the type of reflection, the azimuth angle $\alpha$ is $\phi$ if diffuse reflection dominates otherwise it is $\phi-\pi/2$:

\subsection{SfP from Events}
When estimating surface normals from events, we reconstruct event intensities ($I_e$) as explained in the main paper.
Using the above equations, we estimate $\rho$ and $\theta$ by first estimating event intensities at $12$ polarizer angles,
The use of event intensities enables us to use the traditional SfP algorithms to estimation surface normals.
Depending on the type of polarization (specular or diffuse), this can result in multiple solutions.
We observed using the specular solution results in the lowest angular error.
We also used the Smith \etal \cite{Smith19PAMI} baseline with our event intensities.
However, this results in a lower performance as shown in \Tab \ref{tab:ev_repr}.

\section{Event Representation}
When learning surface normals from events, the event representation have a significant effect on the performance of the network.
In this section, we describe the performance of $3$ kinds of input representations namely: event intensities ($I_e$), voxel grid \cite{Zhu18eccvw}, CVGR representation and CVGR-I representation on the ESfP-synthetic dataset.
The event intensity representation concatenates $I_e$ at polarizer angles of $15, 60, 105, 150$ as input the network.
(note, we cannot use the intensity at $0$ angle, since it will always be zero for all pixels).
The CVGR representation builds on top of voxel grid representation as follows:
\begin{equation}
    E(x, y, b) = \sum_{i=0}^{i=b} C \cdot V(x, y, i) = \sum_{i=0}^{i=b} C \left(\sum_{\substack{e_k \in \mathcal{E}_i :\\ x_k=x, y_k=y }}{p_k}\right).
\end{equation}
Lastly, the CVGR-I representation combines a single image with events and is expressed as follows:
\begin{equation}
    E(x, y, b) = I[0] + \sum_{i=0}^{i=b} C \cdot V(x, y, i)
\end{equation}
As can be seen from \Tab \ref{tab:ev_repr}, the best performing representation is CVGR-I.
The main reason for improvement is because the image gives more context to the network to estimate surface normals in the areas where the event information is insufficient.
Qualitative results on real dataset are shown in \Fig \ref{fig:input_repr}.
As can be seen, the events are triggered prominently on the edge of the vase and are missing from the front-parallel surface of the vase.
The network using only events has a difficult time to estimate normals on these front-parallel surfaces. On the other hand, using CVGR-I representation, the network performs better resulting in a lower MAE score.
Additionally, we also evaluate the effect of number of bins on the performance of the network.
For the same representation, increasing the number of bins from $4$ to $8$ improves the performance by 6\% in terms of angular error.
Higher number of bins preserves the temporal information of events better.
However, further increasing the bins to $12$ results in a a decrease in performance.
This is because not all bins add new information due to limitation of contrast threshold.

\input{floats/tab_event_representation}
\input{floats/fig_img_offset}
\subsection{Dynamic scenes}
\input{floats/fig_motion}
An advantage of using event camera is the high temporal resolution as compared to the frame-based sensor.
To highlight this, we record a dynamic scene.
The scene consists of a rectangular block which is rotating about it's diagonal axis with a drill.
The speed of the drill could be adjusted to three increasing levels.
As seen in \Fig \ref{fig:obj_speed}, the images corresponding to three different speeds are shown in the first row.
Increasing the speed introduces motion blur for the standard camera, which is also reflected in the surface normals (second row, high speed).
In contrast, event-based SfP methods (last row) are better than the image-based counterpart, as can be seen by the sharpness of the edge of the rectangular block.
This is primarily due the high rotation speeds of the polarizer enabled by the high temporal resolution of the camera.

\section{Dataset}
\paragraph{ESfP Synthetic Dataset} In this section, we provide details on the ESfP-Synthetic dataset which we use for evaluation.
The dataset was generated using publicly available meshes \cite{Downs22arxiv} which consists of  over $1000$ 3D scanned common household objects. 
These meshes were textured using the $25$ textures available in this dataset \cite{Baek20siggraph}.
These textures provide polarimetric BRDF of real-world materials which provide accurate polarimetric state information when used with physically-based simulation such as Mitsuba.

\section{Limitations}
Our real-world dataset only considers specular objects such as reflective metallic surfaces.
We specifically chose specular objects as they are the most challenging to obtain surface normals for. 
The geometry of objects with diffuse reflection can be captured easily by methods such as structured light (SL).
Additionally, the intensity changes of diffuse materials when observed with a rotating polarizer is low compared to specular objects, which the real event camera cannot capture due to a high contrast threshold.
Therefore, capturing events for diffuse materials was not possible with the current version of event cameras.

\input{floats/fig_mitsuba_dataset}
\section{Effect of speed}
Conducting experiments at different rotation speeds of the polarizer, we observed a slight increase in the performance for our method and linked this to the decreasing relevance of nonidealities in the event-camera pixel circuits. In this section, we provide more details on why an increase in rotational speed improves the performance. Our analysis is based on additional experiments, general considerations on event-camera circuitry \cite{Lichtsteiner08ssc} and the technical details of the Prophesee Gen 4 event camera \cite{Finateu20isscc}. The additional data we recorded cover rotation speeds between \unit[53]{RPM}  up to \unit[1,734]{RPM} and two illumination conditions, \unit[200]{lux} and \unit[800]{lux}.
In this section, we will use the number of events triggered on the object per revolution of the polarizer as a proxy-measure for the quality of the resulting surface normals.
\begin{figure}[t!]
    \centering
    \input{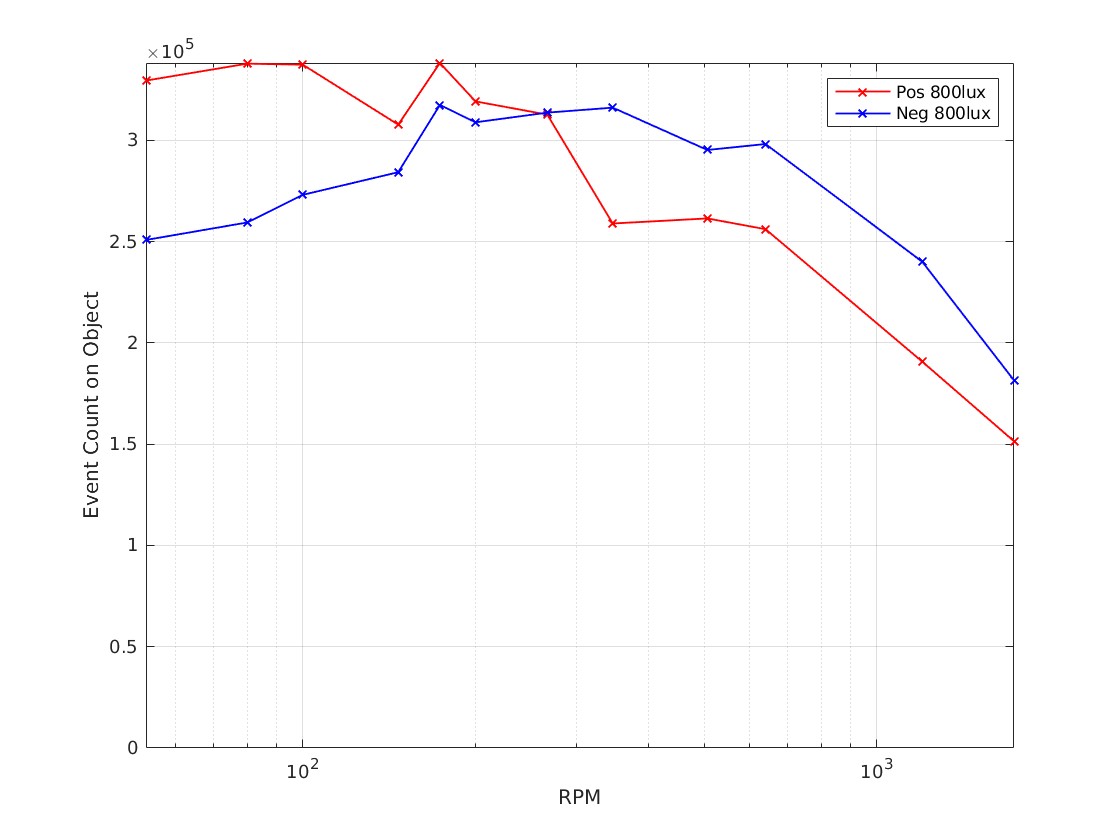}
    \vspace*{-8pt}
    \caption{At low speeds more positive than negative events are triggered. Towards higher rotation speeds this trend reverses but the overall number of events per filter rotation decreases visibly.}
    \label{fig:suppl_number_of_events}
\end{figure}

\paragraph{Ideal Event Camera}
A key observation stated in the main paper is that the illumination intensity cancels out in an idealized event camera model. For a given surface on the object degree of polarization $\rho$ and polarization angle $\phi$,  the event camera observes a sinusoidal intensity profile of the form
\begin{equation}
I(t) = I_\text{un} (1 + \rho \cos(2\omega t - \phi)) 
\label{eq:suppl_intensity}
\end{equation} 

The ideal event-sensor triggers and event when the temporal contrast $T$ (logarithmic intensity change) exceeds some threshold $C$ \cite{Lichtsteiner08ssc}. Combining this with (\ref{eq:suppl_intensity}) yields an expression which, for an object with illumination independent polarization characteristics, only depends on the rotational speed.
\begin{equation}
    T = \frac{\mathrm{d}\left(\ln I(t)\right)}{\mathrm{d}t} = \frac{-2\omega\rho\sin(2\omega t - \phi)}{1 + \rho \cos(2\omega t - \phi)}
    \label{eq:supply_tconpol}
\end{equation}
Based on (\ref{eq:supply_tconpol}), we can see that the number of events triggered per unit time linearly depends on the rotational speed. By considering only the number of events per rotation, this dependency is also cancelled out and \emph{an ideal event camera} would not show any dependency on the illumination condition or rotational speed. 

\begin{figure}[t!]
    \centering
    \input{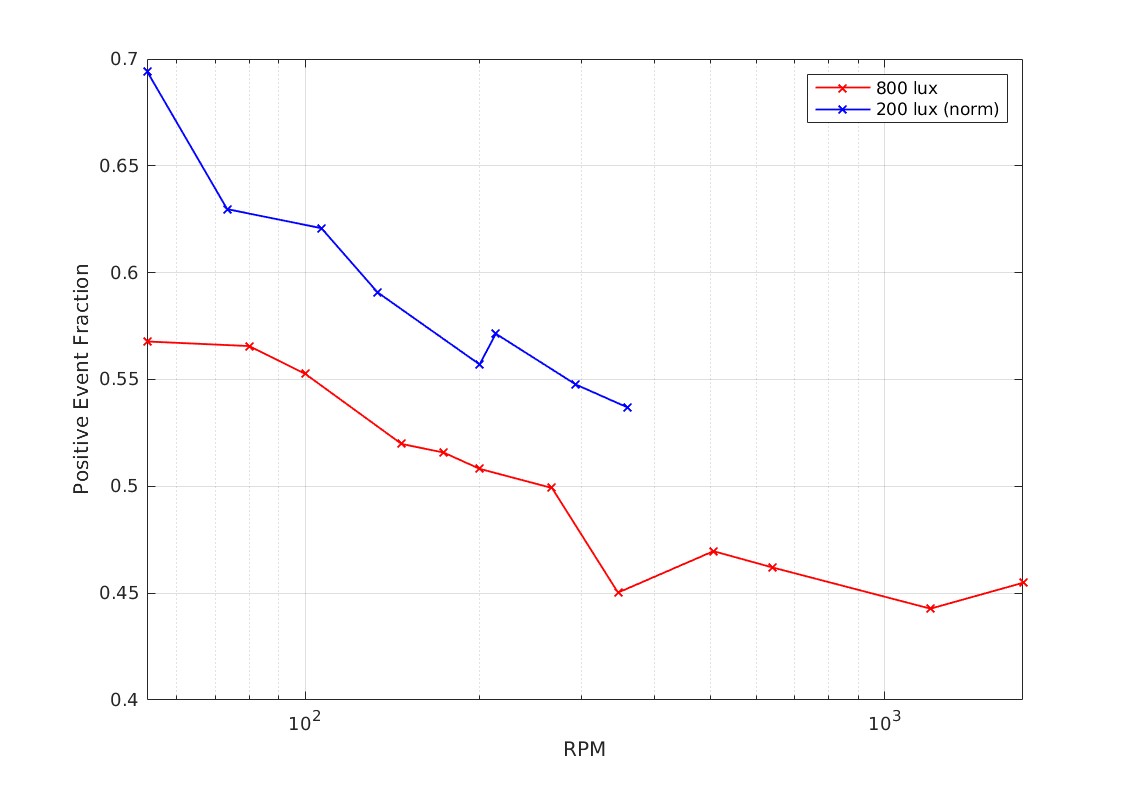}
    \vspace*{-8pt}
    \caption{At low rotational speeds and low light conditions the number of positive events drastically exceeds 0.5 due to the background rate \cite{Lichtsteiner06isscc}}
    \label{fig:ratio}
\end{figure}

\paragraph{Real Event Camera}
In practice however, we observe that illumination and rotation speed have an effect on the quality of the surface normal estimation. To better understand this, we look at the number of events triggered per rotation of the polarizer and observe that
\begin{enumerate}[itemsep=-3pt, topsep=-3pt]
    \item\label{item:number_of_events} at low rotation speeds, more positive events are triggered (Fig.~\ref{fig:suppl_number_of_events}, Fig.~\ref{fig:ratio}),
    \item\label{item:ratio} the difference in fraction of positive and negative events at low RPM becomes more pronounced at lower illumination conditions (Fig.~\ref{fig:ratio}), 
    \item the number of events per rotation decreases at high rotation speeds  (Fig.~\ref{fig:suppl_number_of_events_illum}), and
    \item at low illumnation conditions, less events are triggered at a set rotation speed (Fig.~\ref{fig:suppl_number_of_events_illum}). \\[-4pt]
\end{enumerate}

While the ideal event camera model fails to explain those observations, a more realistic model takes the non-idealities of the circuitry into account. In \cite{Lichtsteiner06isscc} the leakage of the reset-transistor is described as a major source of non-ideality as it leads to the spurious positive events, thus increasing the fraction of positive events. 
Because we consider the number of events per rotation, slower rotation speeds correspond to a longer accumulation times and the BG (background rate)  rate corrupts such low-speed measurements stronger as shown in Fig.~\ref{fig:suppl_number_of_events}. 
This so-called \emph{BG rate} (background rate) is illumination dependent \cite{Finateu20isscc}.
Together with the increase in BG rate at lower light levels\cite{Finateu20isscc} (for bright scenes) this explains observations \ref{item:number_of_events} and \ref{item:ratio}.

\begin{figure}[t!]
    \centering
    \input{floats/TotalEventCount}
    \vspace*{-8pt}
    \caption{The total number of events (given in thousands) per rotation of the polarizer decreases at higher speeds and at lower illumination levels.}
    \label{fig:suppl_number_of_events_illum}
\end{figure}
\begin{figure}[ht!]
    \centering
    \input{floats/EventFrequency}
    \vspace*{-8pt}
    \caption{The frequency at which events are triggered on the object shows a saturation effect due to pixel dead time\cite{Lichtsteiner06isscc}. This explains the decrease in event count at high polarzier speeds.}
    \label{fig:suppl_event_frequency}
\end{figure}
At high rotation speeds the BG rate has negligible influence. However, a second non-ideality becomes visible: after triggering an event the pixel needs a certain \emph{dead time} until it can trigger the next event. This is typically done to avoid bus-saturation by a small group of pixels \cite{Lichtsteiner06isscc}. This effect is clearly visible when looking at the highest event frequency of pixels on the object (Fig.~\ref{fig:suppl_event_frequency}). For an ideal camera the event-frequency would depend linearly on the rotation speed as derived in (\ref{eq:supply_tconpol}). However, the data clearly shows a saturation effect because pixels can only be triggered with a limited frequency, around \unit[1]{kHz} at \unit[800]{lux} illumination. In accordance with literature, this maximum frequency decreases with decreasing illumination \cite{Lichtsteiner06isscc}, explaining the remaining two observations.

 In contrast to the ideal event camera model, the output of a real event-camera is sensitive to illumination and rotational speed of the polarizer. Low speeds increase the BG rate noise significantly and only medium polarizer speeds lead to a more even distribution of positive and negative events. If the speed is increased greatly, the pixel dead time may start to degrade the result again. This is in accordance with the results shown in the main paper.
\begin{figure}
    \centering
    \begin{tikzpicture}[]
    \def\r{0.08cm}
    \tikzstyle{every node}=[font=\footnotesize]
    \begin{axis}[%
    width=6.656cm,
    height=4.0cm,
    at={(0cm,0cm)},
    scale only axis,
    xmin=15,
    xmax=85,
    xlabel={Effective Framerate [fps]},
    ymin=0,
    ymax=90,
    ylabel shift=-4pt,
    ylabel={MAE [degrees]},
    axis background/.style={fill=white},
    xmajorgrids,
    ymajorgrids,
    xtick={0,10,...,60},
    xticklabels = {0,10,...,60}
    ]
    \draw [fill=red, draw=none] (axis cs:22,80.9) circle (\r) node [right=1mm, fill = white, text opacity = 1.0, fill opacity = 0.7] {Mahmoud \emph{et al.} [24]};%
    \draw [fill=red, draw=none] (axis cs:22,67.6) circle (\r) node [right=1mm, fill = white, text opacity = 1.0, fill opacity = 0.7] {Smith \emph{et al.} [43]};%
    \draw [fill=red, draw=none] (axis cs:22,24.5) circle (\r) node [right=1mm, fill = white, text opacity = 1.0, fill opacity = 0.7] {Ba \emph{et al.}};%
    \draw [fill=blue, draw=none] (axis cs:58,58.2) circle (\r) node [inner sep = 1pt, right=1mm, fill = white, text opacity = 1.0, fill opacity = 0.7] {Ours (P)};
    \draw [fill=blue, draw=none] (axis cs:58,27.9) circle (\r) node [inner sep = 1pt, right=1mm, fill = white, text opacity = 1.0, fill opacity = 0.7] {Ours (L)};
    \draw [fill=yellow, draw = none, draw opacity = 1, fill opacity = 0.5] (axis cs:60,0) rectangle (axis cs: 85,20);
    \node [anchor = center] at (axis cs:72.5,10) {Ideal};
    \end{axis}
    \end{tikzpicture}
    \tikzstyle{every node}=[font=\normalsize]
    \caption{Acquisition time versus Mean Angular Error (MAE).}
    \label{fig:fmae}
\end{figure}

\section{Advantage of event camera}

\Fig \ref{fig:fmae} illustrates the advantage of using an event-based SfP (blue) against frame-based SfP methods (red) using as metrics the framerate and
the Mean Angular Error (MAE). 
Image-based approaches focus on maximizing the performance; however, they are restricted by the camera’s framerate to 22 fps while reducing the effective resolution from 4MP to 1MP (DoFP approach). 
On the other hand, our event-based approach is 3 times faster and pushes the SfP methods toward higher framerates, without sacrificing the resolution. 
This enables the capture of surface normals of high-speed motion. Unlike high-framerate cameras, event cameras present a fundamentally new approach to 
visual information processing. 
While a high framerate camera would capture redundant information resulting in data bus saturation, an event camera only triggers events when there is contrast change, resulting in lower bandwidth.

%% file: floats/tab_event_representation.tex
\begin{table}[!t]
    \centering
    \begin{adjustbox}{max width=\linewidth}
    \setlength{\tabcolsep}{4pt}
    {\small
    \begin{tabular}{|l|c|c rrr|}
        \toprule
         Method & Dimension & Angular Error $\downarrow$  &  \multicolumn{3}{c|}{Accuracy $\uparrow$} \\
           & & Mean & \aeone & \aetwo & \aethree \\
        \midrule
        Events (P) \cite{Smith19PAMI} & $ 12 \times H \times W$ & 69.722 & 0.028 & 0.067 & 0.098\\
        Events (P, specular) & $ 12 \times H \times W$ & 58.196 & 0.007 & 0.046 & 0.095 \\ \midrule
        Event intensities & $ 4 \times H \times W$ & 39.316 & 0.147 & 0.321 & 0.402\\
        VoxelGrid \cite{Zhu18eccvw} - 8Bins & $ 8 \times H \times W$ & 34.232 & 0.230 & 0.465 & 0.556 \\
        CVGR - 4Bins & $ 4 \times H \times W$ & 34.053 & 0.220 & 0.494 & 0.579 \\
        CVGR - 8Bins & $ 8 \times H \times W$& 32.010 & 0.248 & 0.515 & 0.594 \\
        CVGR - 12Bins & $ 12 \times H \times W$ & 34.655 & 0.227 & 0.510 & 0.596 \\
        CVGR-I & $ 8 \times H \times W$ & \textbf{27.953} & \textbf{0.263} & \textbf{0.527} & \textbf{0.655} \\
        \bottomrule
    \end{tabular}}
    \end{adjustbox}
    \caption{Comparison of event representations: The first two rows correspond to physics-based baseline.
    Rest of the rows correspond to learning-based approaches with different event representations.
    }
    \label{tab:ev_repr}
\end{table}

%% file: floats/fig_img_offset.tex
\begin{figure}[t]
	\centering
    \setlength{\tabcolsep}{2pt}
	\begin{tabularx}{1\linewidth}{CCCCC}
		Scene & \cite{Ba20ECCV} & Events & E (CVGR) & E\&I (CVGR-I)
		\\
		\mae{images/real_data/vase3_26_10/img_0000.png}{1\linewidth}{}{none}
		&\mae{images/real_data/vase3_26_10/img_unet.jpg}{1\linewidth}{18.06}{white}
		&\mae{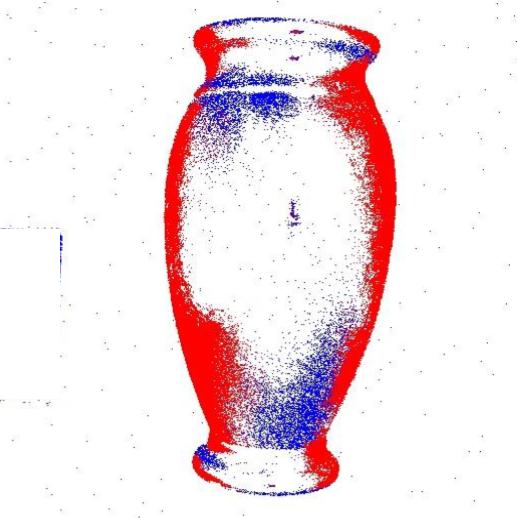}{1\linewidth}{}{none}
		&\mae{images/real_data/vase3_26_10/ev_unet.jpg}{1\linewidth}{25.48}{white}
		&\mae{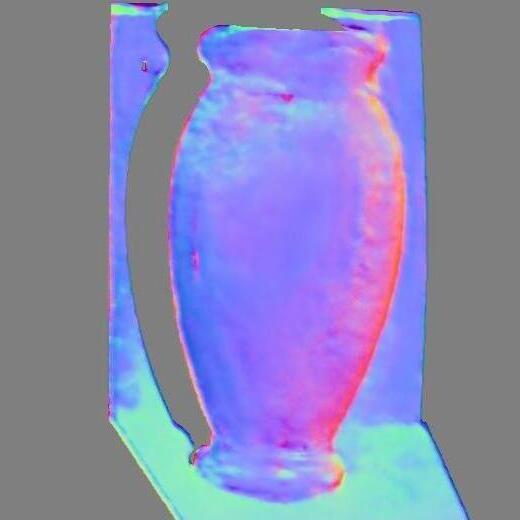}{1\linewidth}{19.64}{white}

	\end{tabularx}
    \vspace{-5ex}
	\caption{Qualitative comparison of event representation}
	\label{fig:input_repr}
    \vspace{-1ex}
\end{figure}

%% file: floats/fig_motion.tex
\begin{figure}[t]
	\centering
    \setlength{\tabcolsep}{2pt}
	\begin{tabularx}{1\linewidth}{CCC}
		Low Speed & Med Speed & High Speed
		\\
		\mae{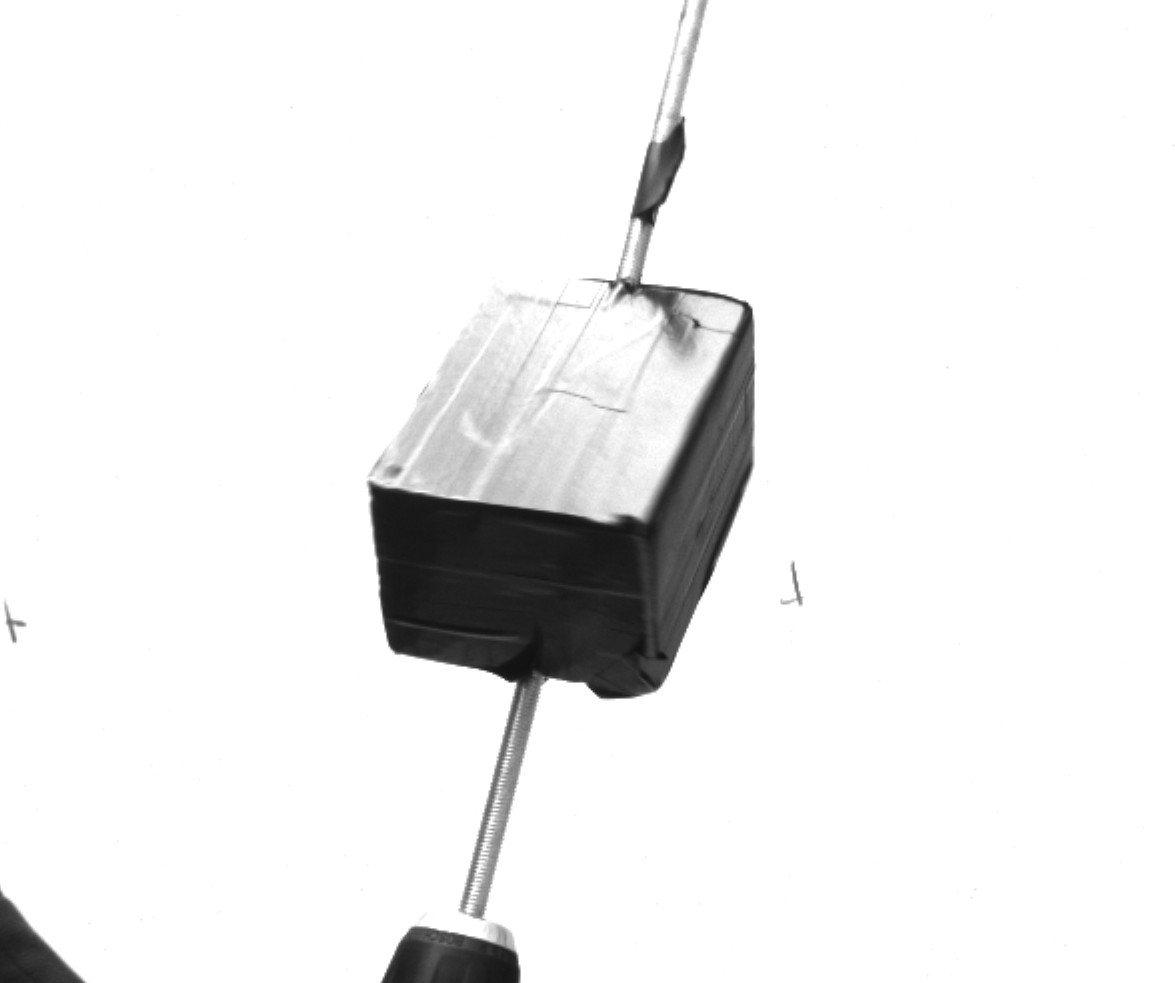}{1\linewidth}{}{none}
		&\mae{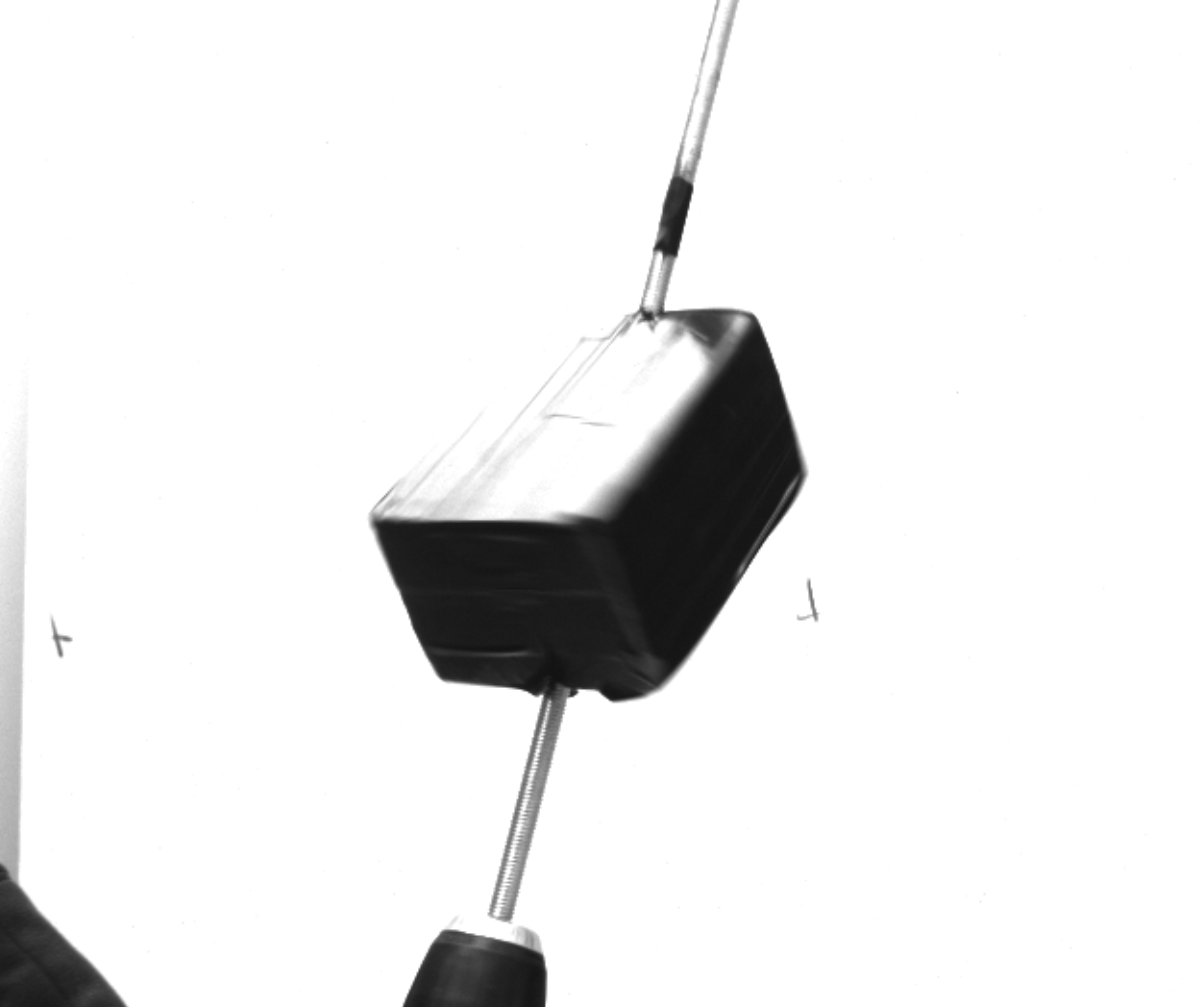}{1\linewidth}{}{none}
		&\mae{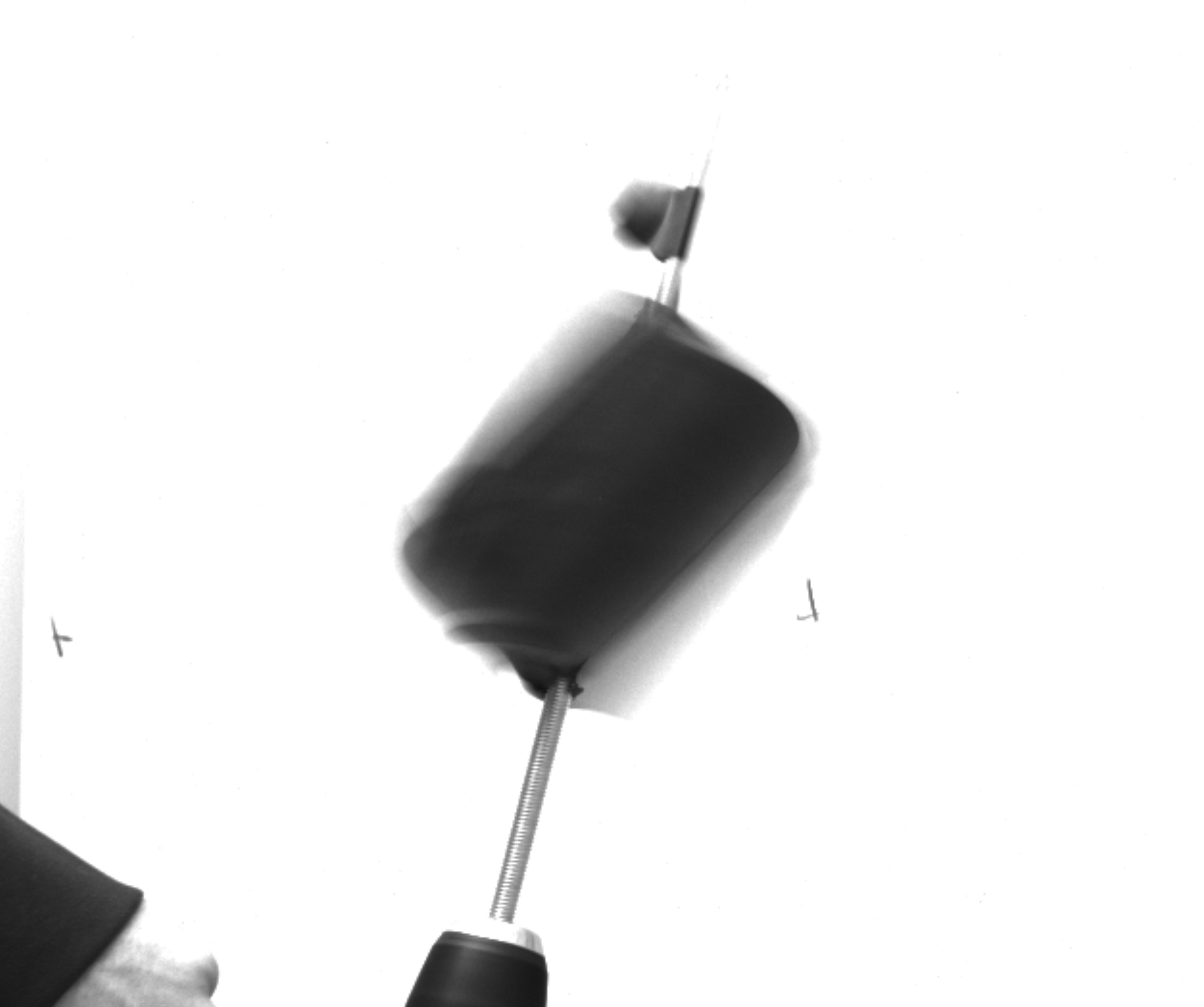}{1\linewidth}{}{none}
		\\
		\mae{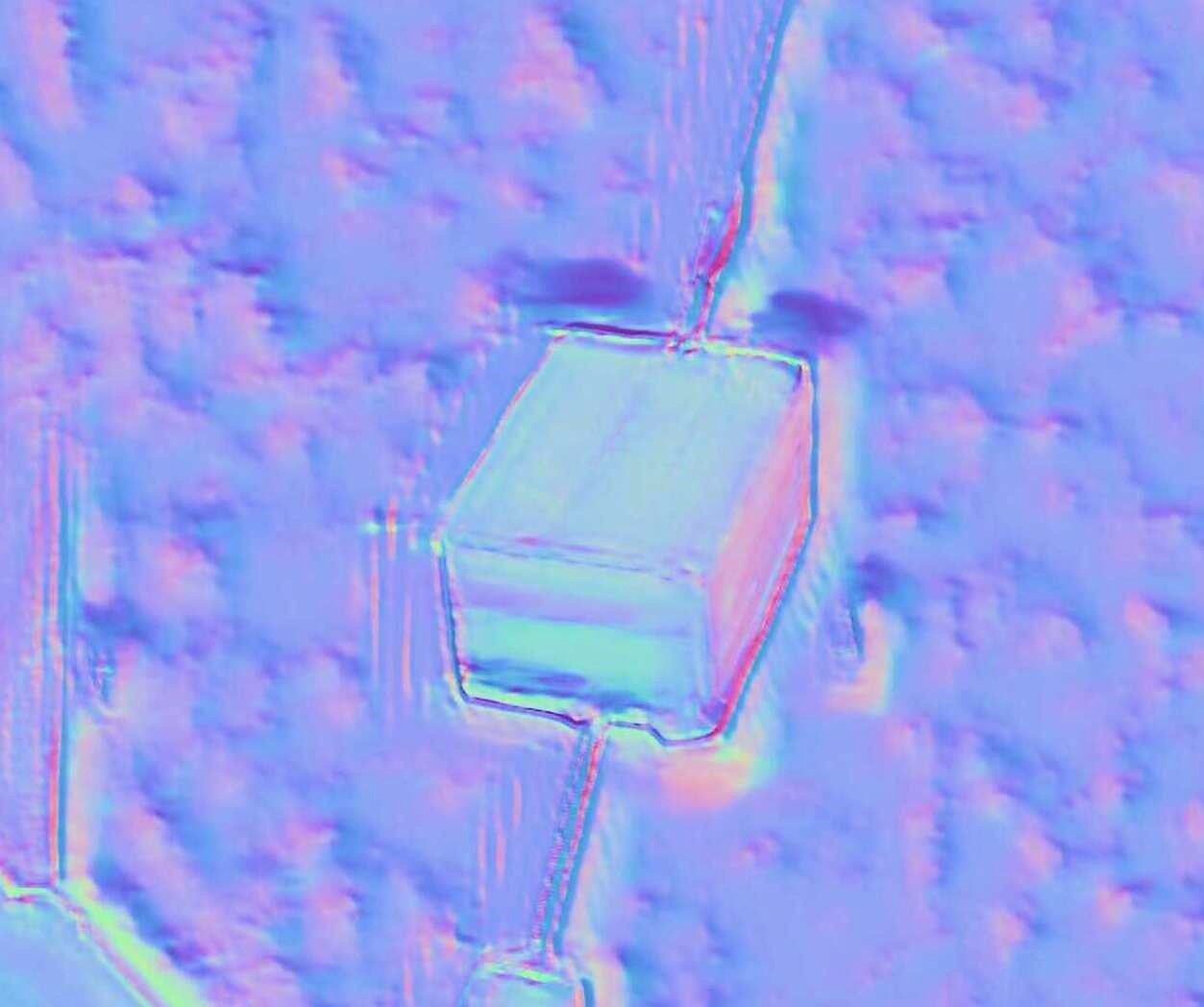}{1\linewidth}{}{none}
		&\mae{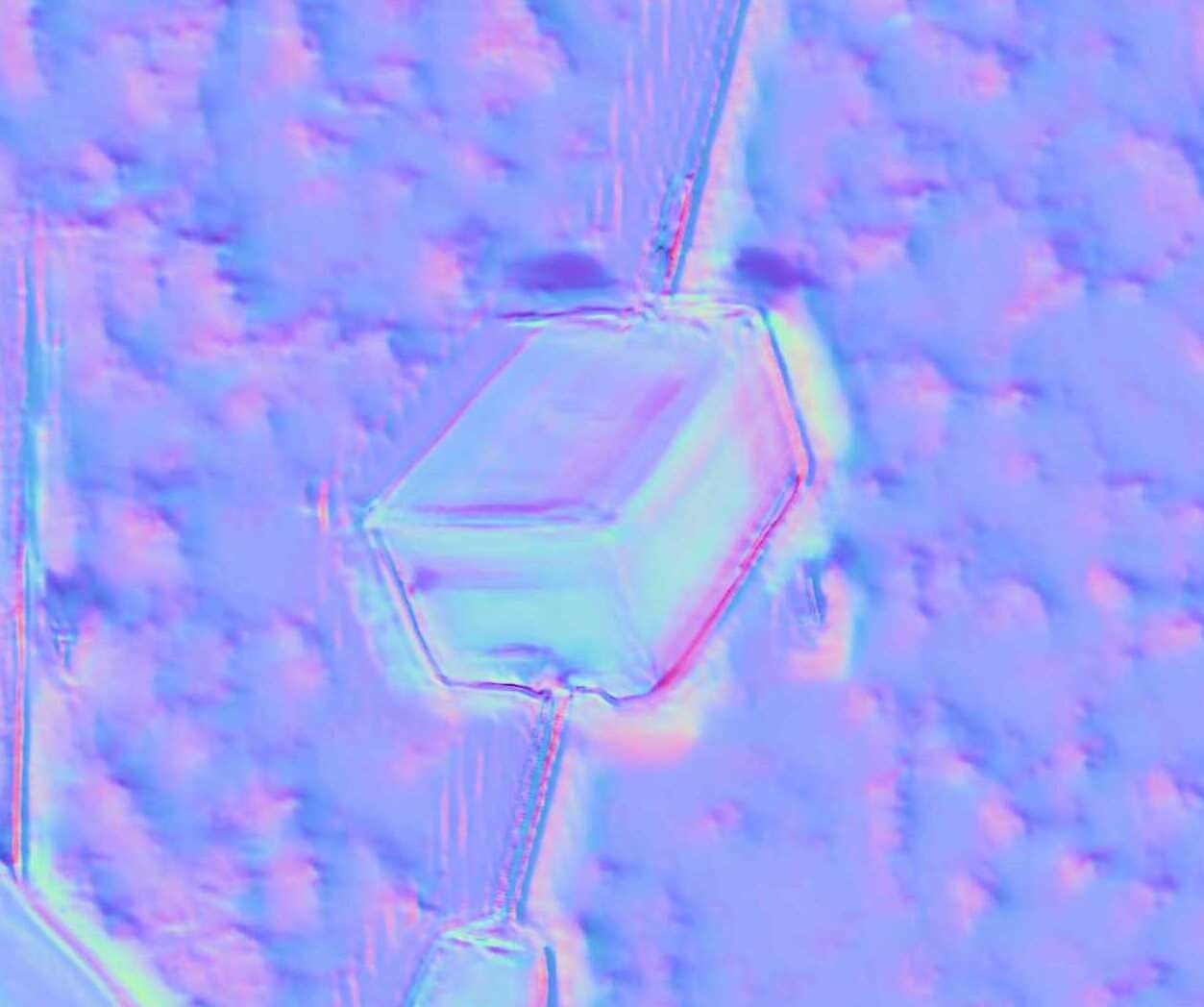}{1\linewidth}{}{none}
		&\mae{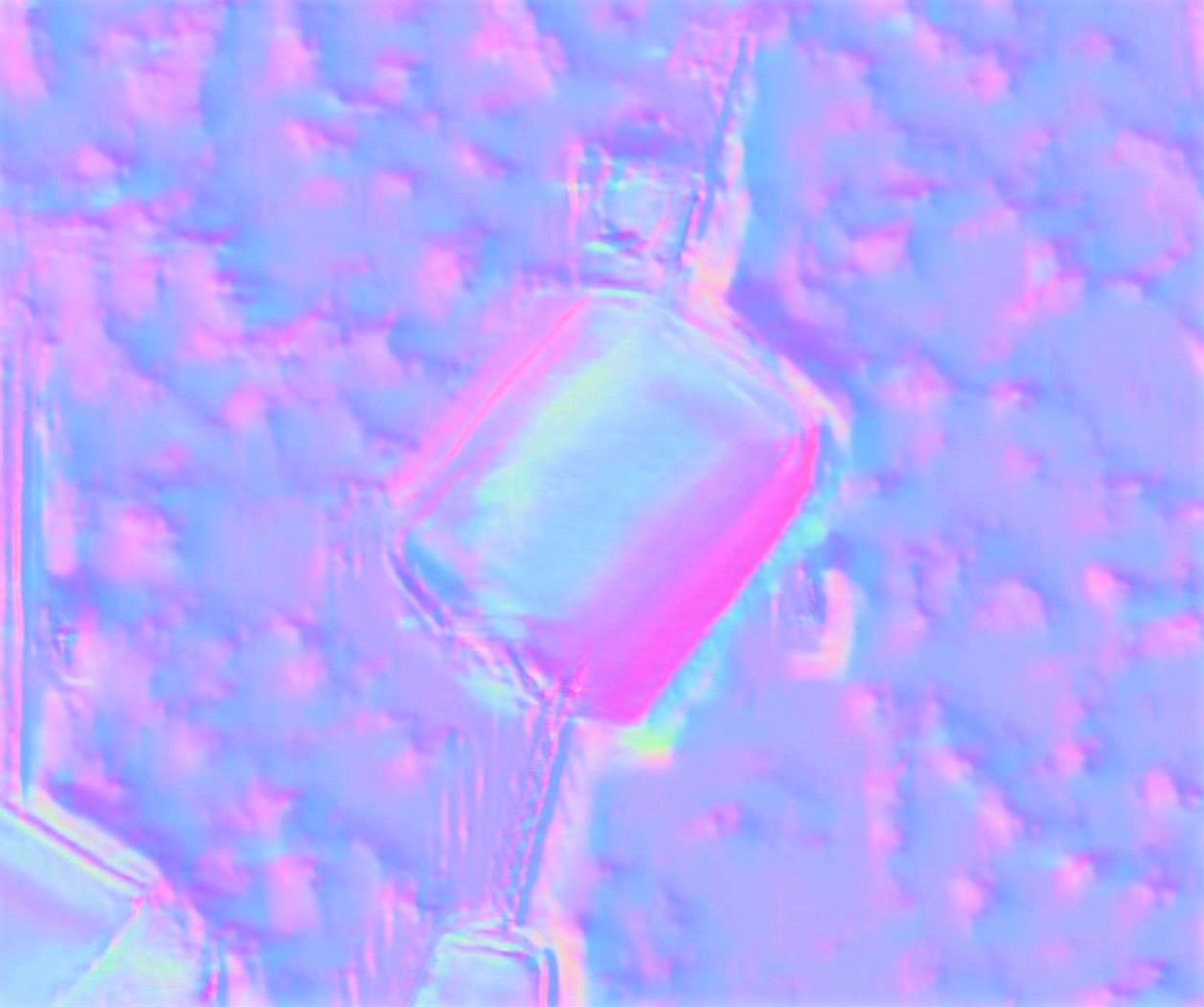}{1\linewidth}{}{none}
		\\
		\mae{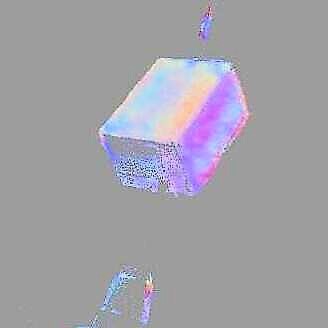}{1\linewidth}{}{none}
		&\mae{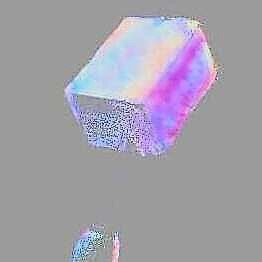}{1\linewidth}{}{none}
		&\mae{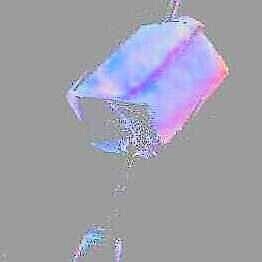}{1\linewidth}{}{none}
		\\

	\end{tabularx}
	\caption{Comparison on dynamic scenes: The object is rotating with increasing speeds from left to right.
	The top row shows the images captured by the camera. The second row shows the surface normals estimated by image-based SfP baseline Ba \etal \cite{Ba20ECCV} and last row shows the surface normals estimation by our learning-based SfP baseline.}
	\label{fig:obj_speed}
\end{figure}

%% file: floats/NumberOfEvents.tex
\tikzstyle{every node}=[font=\footnotesize]
\begin{tikzpicture}

\begin{axis}[%
width=6.656cm,
height=4.5cm,
at={(0cm,0cm)},
scale only axis,
xmode=log,
xmin=53.3337,
xmax=1733.3526,
xminorticks=true,
xlabel={Rotation Speed [RPM]},
ymin=70.0000,
ymax=170.0000,
ylabel={Event Count [kEv]},
axis background/.style={fill=white},
xmajorgrids,
xminorgrids,
ymajorgrids,
legend style={legend cell align=left, align=left, draw=white!15!black},
legend columns=1,
xlabel shift = -2pt,
ylabel shift = -2pt,
line join = round,
legend style={at={(0.5,0.02)},anchor=south},,
scale only axis=true
]
\addplot [color=red, mark=x, mark options={solid, red}]
  table[row sep=crcr]{%
53.3337	164.7320\\
80.0001	168.9175\\
100.0003	168.6675\\
146.6668	153.8665\\
173.3345	169.0775\\
200.0003	159.5920\\
266.6673	156.3375\\
346.6675	129.4610\\
506.6690	130.7125\\
640.0068	128.0480\\
1200.0120	95.3480\\
1733.3526	75.7150\\
};
\addlegendentry{Pos 800lux}

\addplot [color=blue, mark=x, mark options={solid, blue}]
  table[row sep=crcr]{%
53.3337	125.3970\\
80.0001	129.7075\\
100.0003	136.5315\\
146.6668	142.1230\\
173.3345	158.7060\\
200.0003	154.4430\\
266.6673	156.8245\\
346.6675	158.0605\\
506.6690	147.6090\\
640.0068	149.0770\\
1200.0120	120.0075\\
1733.3526	90.7465\\
};
\addlegendentry{Neg 800lux}

\end{axis}
\end{tikzpicture}%
\tikzstyle{every node}=[font=\normalsize]

%% file: floats/FractionPositive.tex
\tikzstyle{every node}=[font=\footnotesize]
\begin{tikzpicture}

\begin{axis}[%
width=6.656cm,
height=3.5cm,
at={(0cm,0cm)},
scale only axis,
xmode=log,
xmin=53.3334,
xmax=1733.3526,
xminorticks=true,
xlabel={Rotation Speed [RPM]},
ymin=0.4000,
ymax=0.7000,
ylabel={Positive Event Fraction},
axis background/.style={fill=white},
xmajorgrids,
xminorgrids,
ymajorgrids,
legend style={legend cell align=left, align=left, draw=white!15!black},
legend columns=1,
xlabel shift = -2pt,
ylabel shift = -2pt,
line join = round,
scale only axis=true
]
\addplot [color=black, mark=x, mark options={solid, black}]
  table[row sep=crcr]{%
53.3337	0.5678\\
80.0001	0.5657\\
100.0003	0.5526\\
146.6668	0.5198\\
173.3345	0.5158\\
200.0003	0.5082\\
266.6673	0.4992\\
346.6675	0.4503\\
506.6690	0.4696\\
640.0068	0.4621\\
1200.0120	0.4427\\
1733.3526	0.4548\\
};
\addlegendentry{800 lux}

\addplot [color=black, dashed, mark=x, mark options={solid, black}]
  table[row sep=crcr]{%
53.3334	0.6943\\
73.3338	0.6297\\
106.6668	0.6208\\
133.3342	0.5908\\
200.0003	0.5570\\
213.3337	0.5714\\
293.3369	0.5476\\
360.0014	0.5369\\
};
\addlegendentry{200 lux}

\end{axis}
\end{tikzpicture}%
\tikzstyle{every node}=[font=\normalsize]

%% file: floats/TotalEventCount.tex
\tikzstyle{every node}=[font=\footnotesize]
\begin{tikzpicture}

\begin{axis}[%
width=6.656cm,
height=3.5cm,
at={(0cm,0cm)},
scale only axis,
xmode=log,
xmin=53.3334,
xmax=1733.3526,
xminorticks=true,
xlabel={Rotation Speed [RPM]},
ymin=0.0000,
ylabel={Total Event Count [kEv]},
axis background/.style={fill=white},
xmajorgrids,
xminorgrids,
ymajorgrids,
legend style={legend cell align=left, align=left, draw=white!15!black},
legend columns=1,
xlabel shift = -2pt,
ylabel shift = -2pt,
line join = round,
legend style={at={(0.5,0.02)},anchor=south},,
scale only axis=true
]
\addplot [color=black, mark=x, mark options={solid, black}]
  table[row sep=crcr]{%
53.3337	290.1290\\
80.0001	298.6250\\
100.0003	305.1990\\
146.6668	295.9895\\
173.3345	327.7835\\
200.0003	314.0350\\
266.6673	313.1620\\
346.6675	287.5215\\
506.6690	278.3215\\
640.0068	277.1250\\
1200.0120	215.3555\\
1733.3526	166.4615\\
};
\addlegendentry{800 lux}

\addplot [color=black, dashed, mark=x, mark options={solid, black}]
  table[row sep=crcr]{%
53.3334	196.5350\\
73.3338	211.0050\\
106.6668	215.8335\\
133.3342	226.1175\\
200.0003	206.5505\\
213.3337	223.5540\\
293.3369	211.0895\\
360.0014	201.0950\\
};
\addlegendentry{200 lux}

\end{axis}
\end{tikzpicture}%
\tikzstyle{every node}=[font=\normalsize]

%% file: floats/EventFrequency.tex
\tikzstyle{every node}=[font=\footnotesize]
\begin{tikzpicture}

\begin{axis}[%
width=6.656cm,
height=3.5cm,
at={(0cm,0cm)},
scale only axis,
xmin=0.0000,
xmax=1800.0000,
xlabel={Rotation Speed [RPM]},
ymin=0.0000,
ymax=800.0000,
ylabel={Maximal Event Frequency},
axis background/.style={fill=white},
xmajorgrids,
ymajorgrids,
legend style={legend cell align=left, align=left, draw=white!15!black},
legend columns=1,
xlabel shift = -2pt,
ylabel shift = -2pt,
line join = round,
legend style={at={(0.5,0.02)},anchor=south},,
scale only axis=true
]
\addplot [color=black, mark=x, mark options={solid, black}]
  table[row sep=crcr]{%
53.3337	113.2154\\
80.0001	146.9543\\
100.0003	196.4977\\
146.6668	267.9483\\
173.3345	319.5123\\
200.0003	336.6947\\
266.6673	387.8445\\
346.6675	422.7336\\
506.6690	499.0279\\
640.0068	566.7528\\
1200.0120	695.5220\\
1733.3526	757.0827\\
};
\addlegendentry{800 lux}

\addplot [color=black, dashed, mark=x, mark options={solid, black}]
  table[row sep=crcr]{%
53.3334	98.7920\\
73.3338	126.4873\\
106.6668	168.0716\\
133.3342	198.1856\\
200.0003	242.6355\\
213.3337	263.5450\\
293.3369	303.8340\\
360.0014	336.0562\\
};
\addlegendentry{200 lux}

\end{axis}
\end{tikzpicture}%
\tikzstyle{every node}=[font=\normalsize]